%% file: main.tex
\documentclass[nohyperref]{article}

\usepackage{microtype}
\usepackage{graphicx}
\usepackage{subfigure}
\usepackage{booktabs} 

\usepackage[pagebackref=true,colorlinks=true,citecolor=blue]{hyperref}

\usepackage[accepted]{icml2022}

\usepackage{amsmath}
\usepackage{amssymb}
\usepackage{mathtools}
\usepackage{amsthm}

\usepackage{multirow}
\usepackage{booktabs}
\usepackage{caption}
\usepackage{wrapfig}

\input{math_commands.tex}

\usepackage[export]{adjustbox}

\usepackage[capitalize,noabbrev]{cleveref}

\renewcommand*{\backref}[1]{}
\renewcommand*{\backrefalt}[4]{%
  \ifcase #1 %
No citations.
  \or
(p. #2).%
  \else
(pp. #2).%
  \fi%
}

\usepackage{url}
\usepackage[autostyle]{csquotes}
\usepackage{epigraph}
\usepackage{graphicx}
\usepackage{pifont}
\usepackage[]{authblk}

\newcommand{\cmark}{\ding{51}}%
\newcommand{\xmark}{\ding{55}}%

\def\approxprop{%
  \def\p{%
    \setbox0=\vbox{\hbox{$\propto$}}%
    \ht0=0.6ex \box0 }%
  \def\s{%
    \vbox{\hbox{$\sim$}}%
  }%
  \mathrel{\raisebox{0.7ex}{%
      \mbox{$\underset{\s}{\p}$}%
    }}%
}

\theoremstyle{plain}
\newtheorem{theorem}{Theorem}[section]
\newtheorem{proposition}{Proposition}

\theoremstyle{definition}

\newtheorem{assumption}[theorem]{Assumption}
\theoremstyle{remark}
\newtheorem{remark}[theorem]{Remark}

\usepackage[textsize=tiny]{todonotes}

\icmltitlerunning{Fishr: Invariant Gradient Variances for Out-of-Distribution Generalization}

\begin{document}

\twocolumn[
    \icmltitle{Fishr: Invariant Gradient Variances for Out-of-Distribution Generalization}




    \begin{icmlauthorlist}
        \icmlauthor{Alexandre Ramé}{yyy}
        \icmlauthor{Corentin Dancette}{yyy}
        \icmlauthor{Matthieu Cord}{yyy,comp}
    \end{icmlauthorlist}

    \icmlaffiliation{yyy}{Sorbonne Université, CNRS, LIP6, Paris, France}
    \icmlaffiliation{comp}{Valeo.ai}

    \icmlcorrespondingauthor{Alexandre Ramé}{alexandre.rame@sorbonne-universite.fr}

    \icmlkeywords{Machine Learning, ICML, Out-of-Distribution Generalization,Invariance}
    \vskip 0.3in
]



\printAffiliationsAndNotice{}  

\begin{abstract}
    \input{sections/00_abstract.tex}
\end{abstract}
\input{sections/01_introduction.tex}
\input{sections/02_relatedwork.tex}
\input{sections/03_model.tex}

\input{sections/05_experiment.tex}

\input{sections/06_conclusion.tex}

\subsubsection*{Acknowledgments}
This work was granted access to the HPC resources of IDRIS under the allocation A0100612449 made by GENCI.
We acknowledge the financial support by the ANR agency in the chair VISA-DEEP (ANR-20-CHIA-0022-01).

\newpage


\bibliography{main}
\bibliographystyle{icml2022}

\newpage
\appendix
\onecolumn
\input{sections/07_appendix.tex}

\end{document}

%% file: math_commands.tex
\usepackage{amsmath,amsfonts,bm}

\def\eqref#1{equation~\ref{#1}}

\def\1{\bm{1}}

\def\vg{{\bm{g}}}

\def\vv{{\bm{v}}}

\def\vx{{\bm{x}}}
\def\vy{{\bm{y}}}

\def\mC{{\bm{C}}}

\def\mF{{\bm{F}}}
\def\mG{{\bm{G}}}
\def\mH{{\bm{H}}}

\def\mK{{\bm{K}}}

\def\mM{{\bm{M}}}

\def\mZ{{\bm{Z}}}

\DeclareMathAlphabet{\mathsfit}{\encodingdefault}{\sfdefault}{m}{sl}
\SetMathAlphabet{\mathsfit}{bold}{\encodingdefault}{\sfdefault}{bx}{n}

\newcommand{\E}{\mathbb{E}}

\newcommand{\Var}{\mathrm{Var}}

\newcommand{\Cov}{\mathrm{Cov}}


%% file: sections/00_abstract.tex
Learning robust models that generalize well under changes in the data distribution is critical for real-world applications.
To this end, there has been a growing surge of interest to learn simultaneously from multiple training domains --- while enforcing different types of invariance across those domains.
Yet, all existing approaches fail to show systematic benefits under controlled evaluation protocols.
In this paper, we introduce a new regularization --- named Fishr --- that
enforces domain invariance in the space of the gradients of the loss: specifically, the domain-level variances of gradients are matched across training domains.
Our approach is based on the close relations between the gradient covariance, the Fisher Information and the Hessian of the loss: in particular, we show that Fishr eventually aligns the domain-level loss landscapes locally around the final weights.
Extensive experiments demonstrate the effectiveness of Fishr for out-of-distribution generalization.
Notably, Fishr improves the state of the art on the DomainBed benchmark and performs consistently better than Empirical Risk Minimization.
Our code is available at \url{https://github.com/alexrame/fishr}.

%% file: sections/01_introduction.tex
\section{Introduction}


The success of deep neural networks in supervised learning \citep{krizhevsky2012imagenet} relies on the crucial assumption that the train and test data distributions are identical.
In particular, the tendency of networks to rely on simple features \citep{valle-perez2018deep,geirhos2020shortcut} is generally a desirable behavior reflecting Occam’s razor.
However, in case of distribution shift, this simplicity bias deteriorates performance when more complex features are needed \citep{tenenbaum2018building,NEURIPS2020_6cfe0e61}.
%
For example, in the recent fight against Covid-19, most of the deep learning methods developed to detect coronavirus from chest scans were shown useless for clinical use \citep{degrave2021ai,roberts2021common}: indeed, networks exploited simple bias in the training datasets such as patients' age or body position rather than `truly' analyzing medical pathologies.


\input{pictures/model/fig_intro.tex}

To better generalize under distribution shifts, most works \citep{blanchard2011generalizing,muandet2013domain} assume that the training data is divided into different training domains
in which there is a constant underlying causal mechanism \citep{peters2016causal}.
To remove the domain-dependent explanations, different \textbf{invariance criteria across those training domains} have been proposed. \citet{ganin2016domain,coral216aaai,sun2016deep} enforce similar feature distributions, others \citep{arjovsky2019invariant,krueger2020utofdistribution} force the classifier to be simultaneously optimal across all domains.
Yet, despite the popularity of this research topic,
none of these methods perform significantly better than the classical Empirical Risk Minimization (ERM) when applied with controlled model selection and restricted hyperparameter search \citep{gulrajani2021in,ye2021odbench}.
These failures motivate the need for new ideas.


To foster the emergence of a shared mechanism with consistent generalization properties, our intuition is that learning should progress consistently and similarly across domains.
Besides, the learning procedure of deep neural networks is dictated by the distribution of the gradients with respect to the network weights \citep{yin2018gradient,sankararaman2020impact} --- usually backpropagated in the network during gradient descent.
Additionally, individual gradients are expressive representations of the input \citep{fort2019stiffness,charpiat2019input}.
Thus, we seek distributional invariance across domains in the gradient space: \textbf{domain-level gradients should be similar}, not only in average direction, but most importantly in statistics such as variance and disagreements.

In this paper, we propose the Fishr regularization for out-of-distribution generalization in classification for computer vision --- summarized in Fig.\ \ref{fig:figintro}. We \textbf{match the domain-level gradient variances}, \textit{i.e.}, the second moment of the gradient distributions.
In contrast, previous gradient-based works such as Fish \citep{shi2021gradient} only match the domain-level gradients means, \textit{i.e.}, the first moment.

Our strategy is also motivated by the close relations between the gradient variance, the Fisher Information \citep{fisher1922mathematical} and the Hessian.
This explains the name of our work, Fishr, using gradients as in Fish and related to the Fisher Matrix.
Notably, we will study how \textbf{Fishr forces the model to have similar domain-level Hessians} and promotes consistent explanations --- by generalizing the inconsistency formalism introduced in \citet{parascandolo2021learning}.

To reduce the computational cost, we justify an approximation that tackles the gradients only in the classifier, easily implemented with BackPACK \citep{dangel2020backpack}.

We summarize our contributions as follows:
\begin{itemize}%
    \item We introduce Fishr, a scalable regularization that brings closer the domain-level gradient variances.
    \item We theoretically justify that Fishr matches domain-level risks and Hessians, and consequently, reduces inconsistencies across domains.
\end{itemize}%
Empirically, we first validate that Fishr tackles distribution shifts on the synthetic Colored MNIST \citep{arjovsky2019invariant}.
Then, we show that Fishr performs best on the DomainBed benchmark \citep{gulrajani2021in} when compared with state-of-the-art counterparts.
Critically, Fishr is the only method to perform systematically better than ERM on all real datasets --- PACS, VLCS, OfficeHome, TerraIncognita and DomainNet.

%% file: pictures/model/fig_intro.tex
\begin{figure}[t]
    \includegraphics[width=1.0\columnwidth]{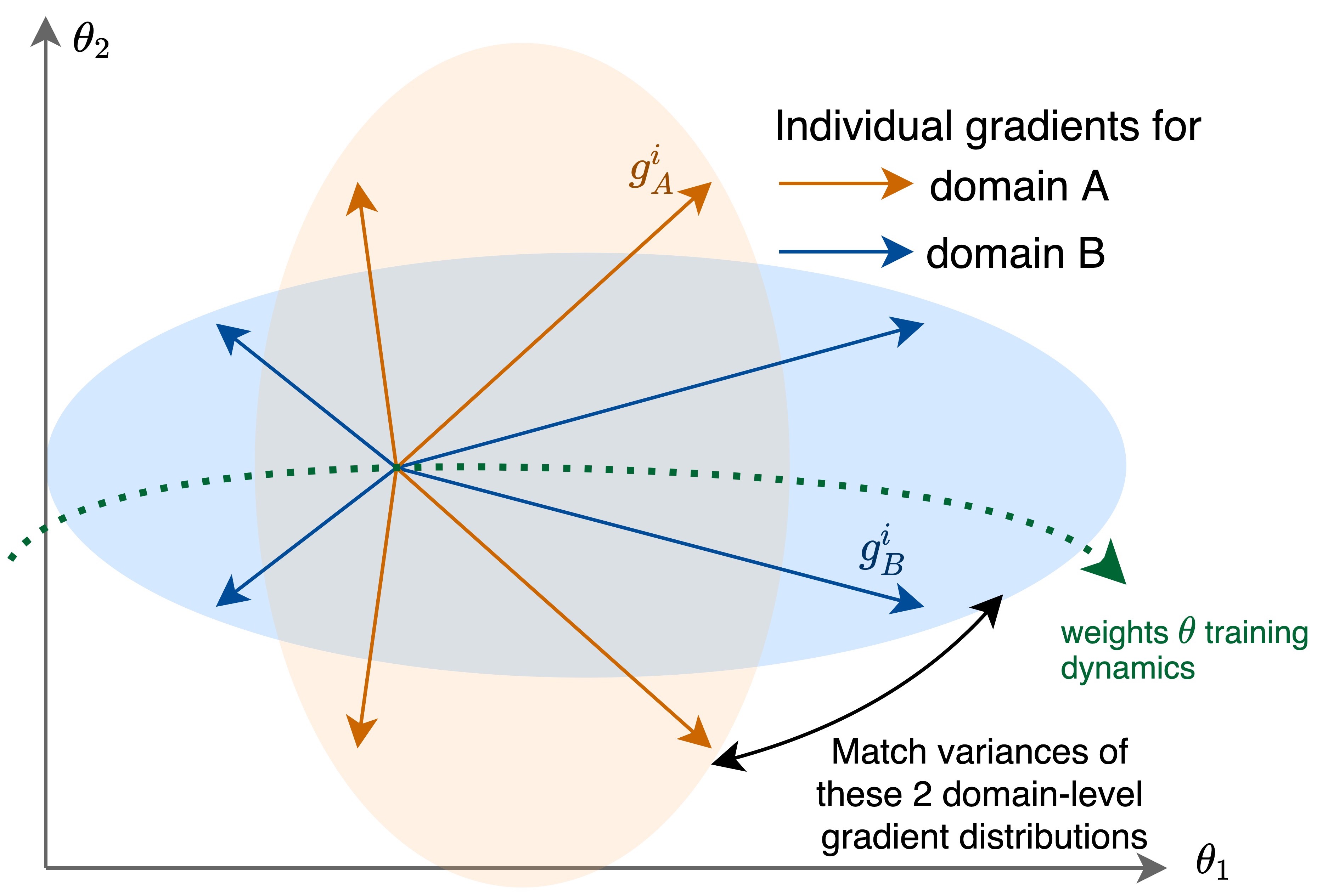}%
    \caption{
    \textbf{Fishr principle}.
    Fishr considers the individual (per-sample) gradients of the loss in the network weights $\theta$.
    Specifically, Fishr matches the domain-level gradient variances of the distributions across the two training domains: $A$ ($\{\vg_A^i\}_{i=1}^{n_A}$ in orange) and $B$ ($\{\vg_B^i\}_{i=1}^{n_B}$ in blue).
    We will show how this regularization during the learning of $\theta$ improves the out-of-distribution generalization properties by aligning the domain-level loss landscapes at convergence.}
    \label{fig:figintro}
\end{figure}

%% file: sections/02_relatedwork.tex
\section{Context and Related Work}
\label{relatedwork}
We first describe our task and provide the notations used along our paper. Then we remind some important related works to understand how our Fishr stands in a rich literature.

\paragraph{Problem definition and notations.}
We study out-of-distribution (OOD) generalization for classification.
Our model is a deep neural network (DNN) $f_{\theta}$ (parametrized by $\theta$) made of a deep features extractor $\Phi_{\phi}$ on which we plug a dense linear classifier $w_{\omega}$
: $f_{\theta}=w_{\omega} \circ \Phi_{\phi}$ and $\theta=(\phi, \omega)$.
In training, we have access to different domains $\mathcal{E}$: for each domain $e\in\mathcal{E}$, the dataset $\mathcal{D}_{e}=\left\{\left(\vx_{e}^{i}, \vy_{e}^{i}\right)\right\}_{i=1}^{n_{e}}$ contains $n_{e}$ i.i.d. (input, labels) samples drawn from a domain-dependent probability distribution. Combined together, the datasets $\left\{\mathcal{D}_{e}\right\}_{e\in\mathcal{E}}$ are of size $n=\sum_{e\in\mathcal{E}} n_e$. Our goal is to learn weights $\theta$ so that $f_{\theta}$ predicts well on a new test domain, unseen in training.
As described in \citet{koh2020wilds} and \citet{ye2021odbench}, most common distribution shifts are \textbf{diversity shifts} --- where the training and test distributions comprise data from related but distinct domains, for instance pictures and drawings of the same objects --- or \textbf{correlation shifts} --- where the distribution of the covariates at test time differs from the one during training.
To generalize well despite these distribution shifts, $f_{\theta}$ should ideally capture an invariant mechanism across training domains.
Following standard notations, $\left\|\mM\right\|_{F}^{2}$ denotes the Frobenius norm of matrix $\mM$; $\left\|\vv\right\|_{2}^{2}$ denotes the euclidean norm of vector $\vv$; $\bm{1}$ is a column vector with all elements equal to $1$.

The standard \textbf{Empirical Risk Minimization} (ERM) \citep{vapnik1999overview} framework simply minimizes the average empirical risk over all training domains, \textit{i.e.}, $\frac{1}{|\mathcal{E}|} \sum_{e\in\mathcal{E}}\mathcal{R}_{e}(\theta)$ where\linebreak
$\mathcal{R}_{e}(\theta)=\frac{1}{n_{e}}\sum_{i=1}^{n_{e}}\ell\left(f_{\theta}\left(\vx_{e}^{i}\right), \vy_{e}^i\right)$
and $\ell$ is the negative log-likelihood loss.
Many approaches try to exploit some external source of knowledge \citep{xie2021innout}, in particular the domain information.
As a side note, these partitions may be inferred if not provided \citep{creager2020nvironment}.
Some works explore data augmentations to mix samples from different domains \citep{wang2020heterogeneous,wu2020dual}, some re-weight the training samples to favor underrepresented groups \citep{Sagawa2020Distributionally,pmlr-v119-sagawa20a,zhang2021deep} and others include domain-dependent weights \citep{ding2017deep,mancini2018best}. Yet, most recent works promote invariance via a regularization criterion and only differ by the choice of the statistics to be matched across training domains.
They can be categorized into three groups: these methods enforce agreement either (1) in features (2) in predictors or (3) in gradients.

First, some approaches aim at extracting \textbf{domain-invariant features} and were extensively studied for unsupervised domain adaptation. The features are usually aligned with adversarial methods \citep{ganin2016domain,pmlr-v48-gong16,li2018domain,li2018domaincond} or with kernel methods \citep{muandet2013domain,long2014transfer}.
Yet, the simple covariance matching in CORAL \citep{coral216aaai,sun2016deep} performs best on various tasks for OOD generalization \citep{gulrajani2021in}. With $\mZ_{e}^{ij}$ the $j$-th dimension of the features extracted by $\Phi_{\phi}$ for the $i$-th example $\vx_{e}^i$ of domain $e \in \mathcal{E}=\left\{A,B\right\}$, CORAL minimizes $\left\|\Cov(\mZ_{A})-\Cov(\mZ_{B})\right\|_{F}^{2}$ where $\Cov(\mZ_{e})=\frac{1}{n_{e}-1}(\mZ_{e}^{\top}\mZ_{e}-\frac{1}{n_{e}}\left(\bm{1}^{\top}\mZ_{e}\right)^{\top}\left(\bm{1}^{\top}\mZ_{e}\right))$ is the feature covariance matrix.
CORAL is more powerful than mere feature matching
$\left\|\frac{1}{n_{A}} \bm{1}^{\top} \mZ_{A} - \frac{1}{n_{B}} \bm{1}^{\top} \mZ_{B}\right\|_{2}^{2}$ as in Deep Domain Confusion (DDC) \citep{tzeng2014deep}.
Yet, \citet{johansson2019support} and \citet{zhao2019learning} show that these approaches are insufficient to guarantee good generalization.

Motivated by arguments from causality \citep{pearl2009causality} and the idea that statistical dependencies are epiphenomena of an underlying structure, Invariant Risk Minimization (IRM) \citep{arjovsky2019invariant} explains that the \textbf{predictor should be invariant} \citep{peters2016causal,rojas2018invariant}, \textit{i.e.}, simultaneously optimal across all domains.
Yet, recent works point out pitfalls of IRM \citep{guo2021utofdistribution,pmlr-v130-kamath21a,ahuja2021invariance}, that does not provably work with non-linear data \citep{rosenfeld2021the} and could not improve over ERM when hyperparameter selection is restricted \citep{koh2020wilds,gulrajani2021in}.
Among many suggested improvements \citep{chang2020invariant,idnanilearning,teney2020nshuffling,ahmed2021systematic}, Risk Extrapolation (V-REx) \citep{krueger2020utofdistribution} argues that training risks from different domains should be similar and thus penalizes $|\mathcal{R}_{A} - \mathcal{R}_{B}|^2$ when $\mathcal{E}=\left\{A,B\right\}$. 

A third and most recent line of work promotes \textbf{agreements between gradients} with respect to network weights.
Gradient agreements help batches from different tasks to cooperate, and have been previously employed for multitasks \citep{du2018adapting,yu2020gradient}, continual \citep{NIPS2017_f8752278}, meta \citep{finn2017odelagnostic,zhang2020daptive} and reinforcement \citep{zhang2019learning} learning.
In OOD generalization, \citet{koyama2020out,parascandolo2021learning,shi2021gradient} try to find minimas in the loss landscape that are shared across domains.
Specifically, these works tackle the domain-level expected gradients:
\begin{equation}
\vg_{e}=\E_{(\vx_{e}, \vy_{e}) \sim \mathcal{D}_{e}} \nabla_{\theta} \ell\left(f_{\theta}(\vx_{e}), \vy_{e}\right).
\end{equation}
When $\mathcal{E}=\left\{A,B\right\}$, IGA \citep{koyama2020out} minimizes $||\vg_{A} - \vg_{B}||_{2}^2$; Fish \citep{shi2021gradient} increases $\vg_{A} \cdot \vg_{B}$; AND-mask \citep{parascandolo2021learning} and others \citep{mansilla2021omain,shahtalebi2021andmask} update weights only when $\vg_{A}$ and $\vg_{B}$ point to the same direction.

Along with the increased computation cost, the main limitation of previous gradient-based methods is the per-domain batch averaging of gradients: this removes more granular statistics, in particular the information from pairwise interactions between gradients from samples in a same domain.
In opposition, our new regularization for OOD generalization keeps extra information from individual gradients and matches across domains the domain-level gradient variances.
In a nutshell, Fishr is similar to the covariance-based CORAL \citep{coral216aaai,sun2016deep} but in the gradient space rather than in the feature space.

%% file: sections/03_model.tex
\input{sections/model/03_0_model.tex}

\input{sections/model/03_1_model.tex}

%% file: sections/model/03_0_model.tex
\section{Fishr}
\label{model:fishr}

\subsection{Gradient variance matching}

The \textbf{individual gradient} $\vg_{e}^{i}=\nabla_{\theta} \ell\left(f_{\theta}(\vx_{e}^i), \vy_{e}^{i}\right)$ is the first-order derivative for the $i$-th data example $\left(\vx_{e}^i, \vy_{e}^{i}\right)$ from domain $e\in\mathcal{E}$ with respect to the weights $\theta$.
Previous methods have matched the gradient means $\vg_{e}=\frac{1}{n_{e}} \sum_{i=1}^{n_{e}} \vg_{e}^{i}$ for each domain $e\in\mathcal{E}$.
These gradient means capture the average learning direction but can not capture gradient disagreements \cite{sankararaman2020impact,yin2018gradient}.
With $\mG_{e}=[\vg_{e}^{i}]_{i=1}^{n_{e}}$ of size $n_{e} \times |\theta|$, we compute the \textbf{domain-level gradient variance} vectors of size $|\theta|$:
\begin{equation}
      \vv_e = \Var(\mG_e) = \frac{1}{n_{e} - 1}\sum_{i=1}^{n_{e}} \left(\vg_{e}^{i} - \vg_{e}\right)^{2},
   \label{eq:variance}
\end{equation}
where the square indicates an element-wise product. To reduce the distribution shifts in the network $f_{\theta}$ across domains, we bring the domain-level gradient variances $\{\vv_e\}_{e\in\mathcal{E}}$ closer.
Hence, our Fishr regularization is:
\begin{equation}
   \mathcal{L}_{\text{Fishr}}(\theta)=\frac{1}{|\mathcal{E}|} \sum_{e\in\mathcal{E}} \left\|\vv_e-\vv \right\|_{2}^{2},
\label{eq:fishrreg}
\end{equation}
the square of the Euclidean distance between the gradient variance from the different domains $e\in\mathcal{E}$ and the mean gradient variance $\vv=\frac{1}{|\mathcal{E}|} \sum_{e\in\mathcal{E}} \vv_{e}$. Balanced with a hyperparameter coefficient $\lambda > 0$, this Fishr penalty complements the original ERM objective, \textit{i.e.}, the empirical training risks:
\begin{equation}
   \mathcal{L}(\theta)=\frac{1}{|\mathcal{E}|} \sum_{e\in\mathcal{E}} \mathcal{R}_{e}(\theta) + \lambda \mathcal{L}_{\text{Fishr}}(\theta).
   \label{eq:loss}
\end{equation}
\begin{remark}
   Gradients $\vg_{e}^{i}$ can be computed on all network weights $\theta$. Yet, to reduce the memory and training costs, they will often be computed only on a subset of $\theta$, \textit{e.g.}, only on classification weights $\omega$. This approximation is discussed in Section \ref{expe:domainbedimpledetails} and Appendix \ref{appendix:iga}.
\end{remark}

%% file: sections/model/03_1_model.tex
\subsection{Theoretical analysis}
\label{model:theo}

We theoretically motivate our Fishr regularization by leveraging the \textbf{domain inconsistency score} introduced in AND-mask \citep{parascandolo2021learning}.
We first derive a generalized upper bound for this score.
Then, we show that Fishr minimizes this upper bound by matching simultaneously \textbf{domain-level risks and Hessians}.

\input{sections/model/03_10t2_model.tex}

\input{sections/model/03_11_model.tex}

\input{sections/model/03_12_model.tex}
\paragraph{Conclusion.} Fishr efficiently matches (1) domain-level empirical risks and (2) domain-level Hessians across the training domains, using gradient variances as a proxy.
This will align domain-level loss landscapes, reduce domain inconsistencies and increase domain generalization.
In particular, the domain-level Hessian matching illustrates that Fishr is more than just a generalization of gradient-mean approaches such as Fish \cite{shi2021gradient}.

Finally, we refer the readers to Appendix \ref{appendix:ntk} where we leverage the Neural Tangent Kernel (NTK) \citep{NEURIPS2018_5a4be1fa} theory to further motivate the gradient variance matching during the optimization process --- and not only at convergence. In brief, as $\mF$ and the NTK matrices share the same non-zero eigenvalues, similar $\{\mC_e\}_{e\in\mathcal{E}}$ during training reduce the simplicity bias by preventing the learning of different domain-dependent shortcuts at different training speeds: this favors a shared mechanism that predicts the same thing for the same reasons across domains.

%% file: sections/model/03_10t2_model.tex
\subsubsection{Inconsistency formalism}
\label{model:inconsistency}
\input{pictures/model/fig_consistency.tex}

\citet{parascandolo2021learning} argues that \enquote{patchwork solutions sewing together different strategies} for different domains may not generalize well: good weights should be optimal on all domains and \enquote{hard to vary} \citep{hardtovary}.
They formalize this insight with an inconsistency score:
\begin{equation}
    \mathcal{I}^{\epsilon}\left(\theta^{*}\right)=\max_{(A, B)\in\mathcal{E}^2} \max_{\theta \in N_{A, \theta^{*}}^{\epsilon}}\left|\mathcal{R}_{B}(\theta)-\mathcal{R}_{A}(\theta^{*})\right|,
\end{equation}
where $\theta \in N_{A, \theta^{*}}^{\epsilon}$ if there exists a path in the weights space between $\theta$ and $\theta^{*}$ where the risk $\mathcal{R}_{A}$ remains in an ${\epsilon>0}$ interval around $\mathcal{R}_{A}(\theta^{*})$.
$\mathcal{I}$ increases with conflicting geometries in the loss landscapes around $\theta^{*}$ as in Fig.\ \ref{fig:fig_consistenty}: \textit{i.e.}, when another `close' solution $\theta$ is equivalent to the current solution $\theta^{*}$ in a domain $A$ but yields different risks in $B$.

For $e\in \mathcal{E}$, the second-order Taylor expansion of $\mathcal{R}_{e}$ around $\theta^{*}=0$ (with a change of variable) gives:
$$
    \mathcal{R}_{e}(\theta)=\mathcal{R}_{e}(\theta^{*}) + \theta^{\top} \nabla_{\theta} \mathcal{R}_{e}\left(\theta^{*}\right) + \frac{1}{2} \theta^{\top} H_{e} \theta + \mathcal{O}(\left\|\theta\right\|_{2}^2),
$$
where the Hessian $\mH_e = \nabla_{\theta}^{2} \mathcal{R}_{e}(\theta^{*})$ approximates the local curvature of the loss landscape.
Moreover, we assume simultaneous convergence, \textit{i.e.}, $\theta^{*}$ is a local minima across all domains: $\nabla_{\theta} \mathcal{R}_{e}(\theta^{*})=0$.
Thus, locally around $\theta^{*}$:
\begin{equation}
    \begin{split}
        &\max_{\theta \in N_{A, \theta^{*}}^{\epsilon}}\left|\mathcal{R}_{B}(\theta)-\mathcal{R}_{A}(\theta^{*})\right|\\
        &\approx \max_{\left|\mathcal{R}_{A}(\theta) - \mathcal{R}_{A}(\theta^{*})\right| \leq \epsilon} \left|\mathcal{R}_{B}(\theta) - \mathcal{R}_{A}(\theta^{*})\right|\\
        &\approx \max _{\frac{1}{2}\left|\theta^{\top} H_{A} \theta\right| \leq \epsilon} \left|\mathcal{R}_{B}(\theta^{*}) + \frac{1}{2} \theta^{\top} H_{B} \theta - \mathcal{R}_{A}(\theta^{*})\right|\\
        &\lessapprox \left|\mathcal{R}_{B}(\theta^{*}) - \mathcal{R}_{A}(\theta^{*})\right| + \max _{\frac{1}{2} \left|\theta^{\top} H_{A} \theta\right| \leq \epsilon} \frac{1}{2} \left|\theta^{\top} H_{B} \theta \right|,
    \end{split}
\end{equation}
where we deduced the last line from the triangle inequality.
Appendix \ref{appendix:limitincon} formally demonstrates following equality.
\vspace{1.em}
\begin{proposition}
Under the quadratic bowl Assumption \ref{assum:quadratic} with positive definite Hessians, for small $\epsilon$ (see Eq.\ \ref{eq:eps}):
\begin{equation}
\begin{split}
\mathcal{I}^{\epsilon}\left(\theta^{*}\right) = \max_{(A, B)\in\mathcal{E}^2} \scalebox{1.5}{$($}&\mathcal{R}_{B}(\theta^{*}) - \mathcal{R}_{A}(\theta^{*})\\
+ &\max _{\frac{1}{2}\theta^{\top} H_{A} \theta \leq \epsilon} \frac{1}{2} \theta^{\top} H_{B} \theta\scalebox{1.5}{$)$}.
\end{split}
\end{equation}
\label{prop:limitincon}
\end{proposition}
The Hessian being positive definite is a standard hypothesis, notably used in \citet{parascandolo2021learning}, that is empirically reasonable \citep{sagun2018empirical}: \enquote{in only very few steps \dots large negative eigenvalues disappear} \citep{pmlr-v97-ghorbani19b}.

\textbf{The first term} in the RHS of Proposition \ref{prop:limitincon} is the difference between domain-level risks, whose square is the criterion minimized in V-REx \citep{krueger2020utofdistribution}. We will prove and show that Fishr forces this term to be small in Section \ref{model:fishrmatchrisks}. In contrast, \citet{parascandolo2021learning} made the strong assumption: $\mathcal{R}_{A}(\theta^{*})= \mathcal{R}_{B}(\theta^{*}) = 0$.

While \citet{parascandolo2021learning} ignored this first term, we follow their diagonal approximation of the Hessians to analyze \textbf{the second term}.
In that case, $\mH_{e}=\operatorname{diag}\left(\lambda_{1}^{e}, \cdots, \lambda_{h}^{e}\right)$ with $\forall i\in \left\{1, \ldots, h\right\}, \lambda_{i}^{e} > 0$. Then:
\begin{equation}
    \begin{split}
        \max_{\frac{1}{2} \theta^{\top} H_{A} \theta \leq \epsilon} \frac{1}{2}  \theta^{\top} H_{B} \theta & = \max _{\|\tilde{\theta}\|_{2}^{2} \leq \epsilon} \sum_{i} \tilde{\theta}_{i}^{2} \lambda_{i}^{B}/\lambda_{i}^{A}\\
        & = \epsilon \times \max _{i} \lambda_{i}^{B}/\lambda_{i}^{A}.
    \end{split}
\end{equation}
This is large when exists $i$ such that $\lambda_i^A$ is small but $\lambda_i^B$ is large: indeed, a small weight perturbation in the direction of the associated eigenvector would change the loss slightly in the domain $A$ but drastically in domain $B$.
Thus, this second term decreases when $\mH_A$ and $\mH_B$ have similar eigenvalues.
This result holds when Hessians are co-diagonalizable.
In conclusion, this explains why forcing $\mH_A=\mH_B$ reduces inconsistencies in the loss landscape and thus improves generalization.
AND-mask matches Hessians by zeroing out gradients with inconsistent directions across domains; however, this masking strategy introduces dead zones \citep{shahtalebi2021andmask} in weights where the model could get stuck, ignores gradient magnitudes and empirically performs poorly with real datasets from DomainBed.
As shown in Section \ref{model:fishrmatchhessians}, Fishr proposes a new method to align domain-level Hessians leveraging the close relations between the gradient variance, the Fisher Information and the Hessian.

%% file: pictures/model/fig_consistency.tex
\begin{figure}[ht]
    \includegraphics[width=1.0\columnwidth]{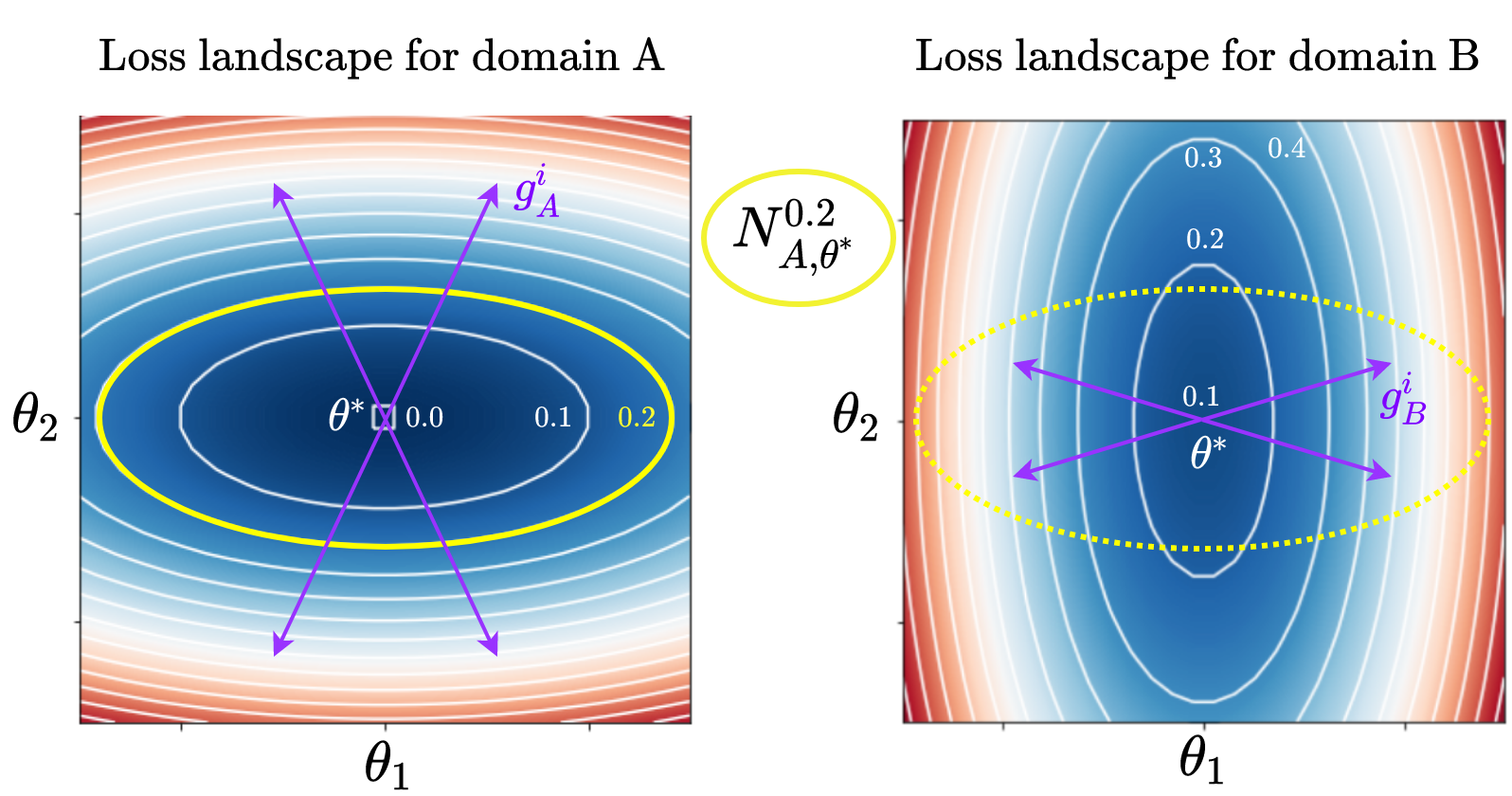}%
    \caption{
    \textbf{Loss landscapes around inconsistent weights} $\theta^{*}$ at convergence.
    $N_{A, \theta^{*}}^{0.2}$ contains weights $\theta$ for which $\mathcal{R}_{A}(\theta)$ is low ($\leq 0.2$) but $\mathcal{R}_{B}(\theta)$ is high  ($\geq 0.9$).
    This inconsistency is due to conflicting domain-level loss landscapes, specifically gaps between domain-level risks and curvatures at $\theta^{*}$. This is visible in the disagreements across the variances of gradients $\{\vg_A^i\}_{i=1}^{n_A}$ and $\{\vg_B^i\}_{i=1}^{n_B}$.}
    \label{fig:fig_consistenty}
\end{figure}

%% file: sections/model/03_11_model.tex
\subsubsection{Fishr matches the domain-level risks}
\label{model:fishrmatchrisks}

Gradients take into account the label $Y$, which appears as an argument for the loss $\ell$. Hence, gradient-based approaches are `label-aware' by design. In contrast, feature-based methods were shown to fail in case of label shifts, because they do not consider $Y$ \citep{johansson2019support,zhao2019learning}.

The fact that the label and the loss appear in the formula of the gradients has another important consequence: matching gradient distributions also matches training risks, as motivated in V-REx \citep{krueger2020utofdistribution}.
We confirm this insight in Table \ref{table:hessianermfishr}: matching gradient variances with Fishr induces $|\mathcal{R}_{A} - \mathcal{R}_{B}|^2 \to 0$ when $\mathcal{E}=\{A, B\}$.

\textit{Intuitively}, gradient amplitudes are directly weighted by the loss values: multiplying the loss by a constant will also multiply the gradients by the same constant. Thus roughly, if the domain-level empirical training risks are different, then the domain-level gradient norms should also differ.

\textit{Theoretically}, we prove in Appendix \ref{appendix:lineartheory} that Fishr regularization component with reference to the classification bias is exactly the difference between domain-level mean squared errors. We recover the objective from V-REx \citep{krueger2020utofdistribution}, with a different loss (squared error instead of negative log likelihood). More generally, we show in this Appendix that Fishr in the classifier $w_{\omega}$ acts as a feature-adaptive version of V-REx: the components in Fishr adaptively force the risks to be similar across domains.

%% file: sections/model/03_12_model.tex
\subsubsection{Fishr matches the domain-level Hessians}
\label{model:fishrmatchhessians}

The Hessian matrix $\mH=\sum_{i=1}^{n} \nabla_{\theta}^{2} \ell\left(f_{\theta}(\vx^i), \vy^{i}\right)$ is of key importance in deep learning. Yet, $\mH$ cannot be computed efficiently in general. Recent methods \citep{izmailov2018averaging,parascandolo2021learning,foret2021sharpnessaware} tackled the Hessian indirectly by modifying the learning procedure. In contrast, we use the fact that the diagonal of $\mH$ is approximated by the gradient variance $\Var(\mG)$; this is confirmed in Table \ref{table:simhessiangradvar}.
This result is derived below from $3$ individual and standard approximation steps.

\begin{table}[h]%
    \centering%
    \caption{%
        \textbf{Cosine similarity between Hessian diagonals and gradient variances} $\cos\left(\operatorname{Diag}\left(\mH_{e}\right), \Var(\mG_{e})\right)$, for an ERM at convergence on Colored MNIST with the two training domains $e\in\{90\%,80\%\}$.}%
    \resizebox{0.8\linewidth}{!}{%
        \begin{tabular}{c c c}%
            \toprule%
            & $e=90\%$ & $e=80\%$ \\ 
            \midrule
            On classifier weights $w$           & 0.9999980  & 0.9999905  \\
            On all network weights $\theta$     & 0.9971040  & 0.9962264  \\
            \bottomrule%
        \end{tabular}%
    }%
    \label{table:simhessiangradvar}%
\end{table}

\paragraph{The Hessian and the `true' Fisher Information Matrix (FIM).} The `true' FIM  $\mF=\sum_{i=1}^{n} \E_{\hat{\vy} \sim P_{\theta}(\cdot|\vx^i)} \left[ \nabla_{\theta} \log p_{\theta}(\hat{\vy}|\vx^i) \nabla_{\theta} \log p_{\theta}(\hat{\vy}|\vx^i)^{\top}\right]$ \citep{fisher1922mathematical,raofim1945} approximates the Hessian $\mH$ with theoretically probably bounded errors under mild assumptions \citep{schraudolph2002fast}.

\paragraph{The `true' FIM and the `empirical' FIM.}
Yet, $\mF$ remains costly as it demands one backpropagation per class. That's why most empirical works (\textit{e.g.}, in compression \citep{frantar2021fficient,pmlr-v139-liu21ab} and optimization \citep{dangel2021vivit}) approximate  the  `true' FIM $\mF$ with
the `empirical' FIM $\tilde{\mF}=\mG_{e}^{\top} \mG_{e}=\sum_{i=1}^{n}\nabla_{\theta} \log p_{\theta}(\vy^i|\vx^i)\nabla_{\theta}\log p_{\theta}(\vy^i|\vx^i)^{\top}$ \citep{martens2014new} where $p_{\theta}(\cdot|\vx)$ is the density predicted by $f_{\theta}$ on input $\vx$.
While $\mF$ uses the model distribution $P_{\theta}(\cdot|X)$, $\tilde{\mF}$ uses the data distribution $P(Y|X)$.
Despite this key difference, $\tilde{\mF}$ and $\mF$ were shown to share the same structure and to be similar up to a scalar factor \citep{interplayinfomatrix2020}. They also have analogous properties: $\operatorname{Tr}(\tilde{\mF}) \approx \operatorname{Tr}(\mF)$.
This was discussed in \citet{li2020hessian} and further highlighted even at early stages of training (before overfitting) in the Fig.\ 1 and the Appendix S3 of \citet{NEURIPS2020_d1ff1ec8}. 

\paragraph{The `empirical' FIM and the gradient covariance.}
Critically, $\tilde{\mF}$ is nothing else than the unnormalized uncentered covariance matrix when $\ell$ is the negative log-likelihood. Thus, the gradient covariance matrix $\mC=\frac{1}{n - 1} \left(\mG^{\top} \mG -\frac{1}{n}\left(\bm{1}^{\top} \mG\right)^{\top}\left(\bm{1}^{\top} \mG\right)\right)$ of size $|\theta| \times |\theta|$ and $\tilde{\mF}$ are equivalent (up to the multiplicative constant $n$) at any first-order stationary point: $\mC \approxprop \tilde{\mF}$. Overall, this suggests that $\mC$ and $\mH$ are closely related \cite{jastrzebski2018three};.
\begin{table}[h]
    \centering
    \caption{
        \textbf{Invariance analysis} at convergence on Colored MNIST across the two training domains $\mathcal{E}=\{90\%,80\%\}$.
        Compared to ERM, Fishr matches the gradient variance ($\operatorname{Diag}(\mC_{90\%})\approx\operatorname{Diag}(\mC_{80\%})$) in all network weights $\theta$.
        Most importantly, this enforces invariance in domain-level risks ($\mathcal{R}_{90\%}\approx\mathcal{R}_{80\%}$) and in domain-level Hessians ($\operatorname{Diag}(\mH_{90\%})\approx\operatorname{Diag}(\mH_{80\%})$).
        The gradient variance, computable efficiently with a unique backpropagation, serves as a proxy for the Hessian.
        Details and more experiments in Section \ref{expe:irmcmnist} (notably Fig.\ \ref{fig:fishrhessianacc}) and in Appendix \ref{appendix:hessianmatching}.
    }
    \resizebox{\linewidth}{!}{%
        \begin{tabular}{c c c}
            \toprule
                                                                                           & ERM                  & Fishr                \\
            \midrule
            $\left\|\Var(\mG_{90\%})-\Var(\mG_{80\%})\right\|_{F}^{2}$                     & $1.6$                & $4.1 \times 10^{-5}$ \\
            $ |\mathcal{R}_{90\%} - \mathcal{R}_{80\%}|^2$
            & $1.0 \times 10^{-2}$ & $3.8 \times 10^{-6}$ \\
            $\left\|\operatorname{Diag}\left(\mH_{90\%}-\mH_{80\%}\right)\right\|_{F}^{2}$ & $2.9 \times 10^{-1}$ & $2.7 \times 10^{-4}$ \\
            \bottomrule%
        \end{tabular}%
    }%
    \label{table:hessianermfishr}%
\end{table}
\paragraph{Consequences for Fishr.}
Critically, Fishr considers the gradient variance $\Var(\mG)$, \textit{i.e.}, the diagonal components of $\mC$.
In our multi-domain framework, we define the domain-level matrices with the subscript $e$.
Table \ref{table:hessianermfishr} empirically confirms that matching $\{\operatorname{Diag}(\mC_e)\}_{e\in\mathcal{E}}$  --- i.e., $\{\Var(\mG_e)\}_{e\in\mathcal{E}}$  --- with Fishr forces the domain-level Hessians $\{\operatorname{Diag}(\mH_e)\}_{e\in\mathcal{E}}$ to be aligned at convergence (on the diagonal for computational reasons).
Tackling the second moment of the first-order derivatives enables to regularize the second-order derivatives.
Moreover, Appendix \ref{appendix:centered} shows that matching the diagonals of $\{\mC_e\}_{e\in\mathcal{E}}$ or $\{\tilde{\mF}_e\}_{e\in\mathcal{E}}$ --- \textit{i.e.}, centering or not the variances --- perform similarly.

\paragraph{}
\begin{remark}
\textbf{Limitation of our approximation.}
We acknowledge that approximating the `true' FIM $\mF$ by the `empirical' FIM $\tilde{\mF}$ is not fully justified theoretically \citep{martens2014new,NEURIPS2019_46a558d9}. Indeed, this approximation is valid only under strong assumptions, in particular $\chi^2$ convergence of predictions $P_{\theta}(\cdot|X)$ towards labels $P(Y|X)$ --- as detailed in Proposition 1 from \citet{interplayinfomatrix2020}.
In this paper, we trade off theoretical guarantees for efficiency.
\end{remark}

\paragraph{}
\begin{remark}
\textbf{Diagonal approximation.}
The empirical similarities between $\mC$ and $\mH$ motivate using \textbf{gradient variance rather than gradient covariance}, which scales down the number of targeted components from $|\theta|^2$ to $|\theta|$.
Indeed, diagonally approximating the Hessian is common: \textit{e.g.}, for OOD generalization \cite{parascandolo2021learning}, optimization \cite{series/lncs/LeCunBOM12,kingma2014adam}, continual learning \cite{kirkpatrick2017overcoming} and pruning \cite{e9ee2143e19d49cf9cbe8861950b6b2a,Theis2018a}.
This is based on the empirical evidence \cite{becker1988improving} that Hessians are diagonally dominant at the end of training.
Our diagonal approximation is also motivated by the critical importance of $\operatorname{Tr}(\mC)$ \cite{jastrzebski2021catastrophic,faghri2020study} to analyze the generalization properties of DNNs.
We confirm empirically in Appendix \ref{appendix:offdiag} that considering the off-diagonal parts of $\mC$ performs no better that just matching the diagonals.
\end{remark}

%% file: sections/05_experiment.tex
\section{Experiments}%
\label{expe}
We prove Fishr effectiveness on Colored MNIST \citep{arjovsky2019invariant} and then on the DomainBed benchmark \citep{gulrajani2021in}.
To facilitate reproducibility, the code is available at \url{https://github.com/alexrame/fishr}.
Moreover, we show in Appendix \ref{appendix:linearexpe} that Fishr is effective in the linear setting.

\input{sections/experiments/05_experiment_cmnist.tex}
\input{sections/experiments/05_experiment_domainbed.tex}

%% file: sections/experiments/05_experiment_cmnist.tex
\subsection{Proof of concept on Colored MNIST}
\label{expe:irmcmnist}
\input{tables/irm/cmnist_new.tex}
The task in Colored MNIST \citep{arjovsky2019invariant} is to predict whether the digit is below or above 5.
Moreover, the labels are flipped with 25\% probability (except in Appendix \ref{appendix:cmnistclean}).
Critically, the digits' colors spuriously correlate with the labels: the correlation strength varies across the two training domains $\mathcal{E}=\{90\%, 80\%\}$.
To test whether the model has learned to ignore the color, this correlation is reversed at test time.
In brief, a biased model that only considers the color would have 10\% test accuracy whereas an oracle model that perfectly predicts the shape would have 75\%.
As previously done in V-REx \citep{krueger2020utofdistribution}, we \textbf{strictly} follow the IRM implementation and just replace the IRM penalty by our Fishr penalty.
This means that we use the exact same MLP and hyperparameters, notably the same \textbf{two-stage scheduling}  selected in IRM for the regularization strength $\lambda$, that is low until epoch 190 and then jumps to a large value, which was optimized via a grid-search for IRM. More experimental details are provided in Appendix \ref{appendix:cmnistdetails}.

Table \ref{table:cmnist_25} reports the accuracy averaged over 10 runs with standard deviation.
Fishr$_{\theta}$ (\textit{i.e.}, applying Fishr on all weights $\theta$) obtains the best trade-off between train and test accuracies; notably in test, it reaches 71.2\%, or 70.2\% when digits are grayscale.
Moreover, computing the gradients only in the classifier $w_{\omega}$ performs almost as well (69.5\% in test for Fishr$_{\omega}$) while reducing drastically the computational cost.
Finally, Fishr$_{\phi}$ only in the features extractor $\phi$ works best in test, though it has lower train accuracy.
This last experiment shows that we can reduce domain shifts without explicitly forcing the predictors to be simultaneously optimal.
These results highlight the effectiveness of gradient variance matching --- even with standard hyperparameters --- at different layers of the network.

\input{pictures/model/cmnist_dynamics_fishr_hessian_acc.tex}

The main advantage of this synthetic dataset is the possibility of empirically validating some theoretical insights.
For example, the training dynamics in Fig.\ \ref{fig:fishrhessianacc} show that the domain-level empirical risks get closer once the Fishr$_{\theta}$ gradient variance matching loss is activated after step 190 ($|\mathcal{R}_{90\%} - \mathcal{R}_{80\%}| \to 0$), even though predicting accurately on the domain 90\% is easier than on the domain 80\%.
This confirms insights from Section \ref{model:fishrmatchrisks}.
Similarly, we observe that Fishr matches Hessians across the two training domains.
This is confirmed by further experiments in Appendix \ref{appendix:cmnistexpeadd}, and validates insights from Section \ref{model:fishrmatchhessians}.
Overall, Fishr regularization reduces train accuracy, but sharply increases test accuracy.
Yet, the main drawback of Colored MNIST is its insufficiency to ensure generalization for real-world datasets.
Overall, it should be considered as a proof-of-concept.

%% file: tables/irm/cmnist_new.tex
\begin{table}[b]
    \caption{\textbf{Colored MNIST} results. All methods use hyperparameters optimized for IRM.}
    \resizebox{1.0\linewidth}{!}{%
        \centering
        \begin{tabular}{c  c c c}
            \toprule
            Method           & Train acc.     & Test acc.      & Gray test acc. \\
            \midrule
            ERM              & 86.4 $\pm$ 0.2 & 14.0 $\pm$ 0.7 & 71.0 $\pm$ 0.7 \\
            IRM              & 71.0 $\pm$ 0.5 & 65.6 $\pm$ 1.8 & 66.1 $\pm$ 0.2 \\
            V-REx            & 71.7 $\pm$ 1.5 & 67.2 $\pm$ 1.5 & 68.6 $\pm$ 2.2 \\
            \midrule
            Fishr$_{\theta}$ & 69.6 $\pm$ 0.9 & 71.2 $\pm$ 1.1 & 70.2 $\pm$ 0.7 \\
            Fishr$_{\omega}$ & 71.0 $\pm$ 0.9 & 69.5 $\pm$ 1.0 & 70.2 $\pm$ 1.1 \\
            Fishr$_{\phi}$   & 65.6 $\pm$ 1.3 & 73.8 $\pm$ 1.0 & 70.0 $\pm$ 0.9 \\
            \bottomrule
        \end{tabular}%
    }
    \label{table:cmnist_25}
\end{table}

%% file: pictures/model/cmnist_dynamics_fishr_hessian_acc.tex
\begin{figure}[t]
    \centering
    \includegraphics[width=1.0\linewidth]{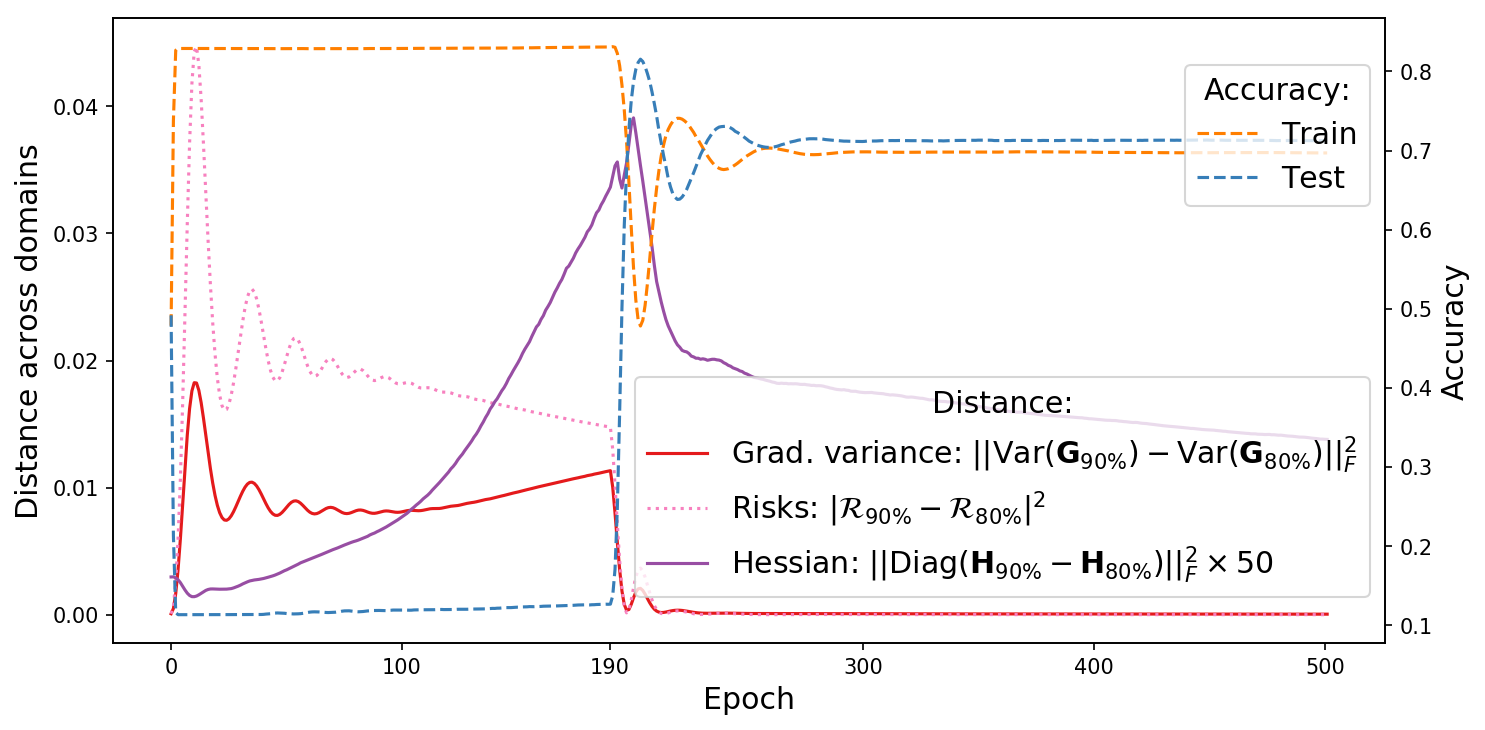}%
    \caption{\textbf{Colored MNIST dynamics}. At epoch 190, $\lambda$ strongly steps up: then, the Fishr$_{\theta}$ regularization matches the domain-level gradient variances (red) across domains $\mathcal{E}=\{90\%, 80\%\}$, and consequently, the training empirical risks (dotted pink) and Hessians (purple). This reduces train accuracy (orange) but increases test accuracy (blue) as the network learns to predict the digit's shape.
    As shown in \linebreak Fig.\ \ref{fig:ermhessianacc}, training dynamics are different for ERM.
    }
    \label{fig:fishrhessianacc}%
\end{figure}

%% file: sections/experiments/05_experiment_domainbed.tex
\subsection{DomainBed benchmark}
\subsubsection{Datasets and procedure}
\label{expe:domainbed}%
We conduct extensive experiments on \textbf{the DomainBed benchmark} \citep{gulrajani2021in}.
In addition to the synthetic Colored MNIST \citep{arjovsky2019invariant} and Rotated MNIST \citep{ghifary2015domain}, the multi-domain image classification datasets are the real VLCS \citep{fang2013unbiased}, PACS \citep{li2017deeper}, OfficeHome \citep{venkateswara2017deep}, TerraIncognita \citep{beery2018recognition} and DomainNet \citep{peng2019moment}.
To limit access to test domain, the framework enforces that all methods are trained with only 20 different configurations of hyperparameters and for the same number of steps. Results are averaged over three trials.
This experimental setup is further described in Appendix \ref{appendix:domainbeddetails}.
By imposing the datasets, the training procedure and controlling the hyperparameter search, DomainBed is arguably the fairer open-source benchmark to rigorously compare the different strategies for OOD generalization.
\input{pictures/pseudocode_short.tex}
\subsubsection{Implementation details}
\input{tables/oracle/db_all_oracle.tex}

\label{expe:domainbedimpledetails}%
We systematically apply Fishr only in the classifier $w_{\omega}$ in DomainBed.
Indeed, keeping individual gradients in memory for $\phi$ from a ResNet-50 was impossible for computational reasons.
Fishr$_{\theta}$ and Fishr$_{\omega}$ performed similarly in previous Section \ref{expe:irmcmnist}.
This is partly because the gradients in $\omega$ still depend on $\Phi_{\phi}$.
Additionally, as highlighted in Appendix \ref{appendix:iga}, this relaxation may improve results for real-world datasets.
Indeed, while Colored MNIST is a correlation shift challenge, the other datasets mostly demonstrate diversity shifts where \enquote{each domain represents a certain spectrum of diversity in data}  \citep{ye2021odbench}. Then, as the pixels distribution are quite different across domains, low-level layers may need to adapt to these domain-dependent peculiarities. Moreover, if we used all weights $\theta=(\phi, \omega)$ to compute gradient variances, the invariance in  $w_{\omega}$ may be overshadowed by $\Phi_{\phi}$ due to $|\omega| \ll |\phi|$. Finally, it's worth noting that this \textbf{last-layer approximation} is consistent with the IRM condition \citep{arjovsky2019invariant} and is common for unsupervised domain adaptation  \citep{ganin2016domain}.

Fishr relies on three \textbf{hyperparameters}. \textit{First}, the $\lambda$ coefficient controls the regularization strength: with $\lambda=0$ we recover ERM while a high $\lambda$ may cause underfitting. We show that Fishr is robust to the choice of the sampling distribution for hyperparameter $\lambda$ in Appendix \ref{appendix:v15}. \textit{Second} the warmup iteration defines the step at which we activate the regularization. This warmup strategy is taken from previous works such as IRM \citep{arjovsky2019invariant}, V-REx \citep{krueger2020utofdistribution} or Spectral Decoupling \citep{pezeshki2020radient}. Before that step, the DNN is trained with ERM to learn predictive features. After that step, the Fishr regularization encourages the DNN to have invariant gradient variances. \textit{Lastly}, the domain-level gradient variances are more accurate when estimated over more data points.
Rather than increasing the batch size, we follow \citet{le201115} and leverage an exponential moving average for computing stable gradient variances. Therefore our third hyperparameter is the coefficient $\gamma$ controlling the update speed: at step $t$, we match $\bar{\vv}_e^{t}=\gamma \bar{\vv}_e^{t-1}+(1-\gamma)\vv_e^{t}$ rather than of $\vv_e^{t}$ from Eq.\ \ref{eq:variance}.
The closer $\gamma$ is to 1, the smoother the variance is along training.
$\bar{\vv}_e^{t-1}$ from previous step $t-1$ is `detached' from the computational graph.
Similar strategies have already been used for OOD generalization \citep{NEURIPS2020_eddc3427,JMLR:v22:17-679}.
The memory overhead is $\left(|\mathcal{E}| \times |\omega|\right)$.
We study by \textbf{ablation} the importance of this warmup strategy and this $\gamma$ in Appendices
\ref{appendix:ablationema} and \ref{appendix:iga}.

Fishr is simple to implement (see the Algorithm \ref{pseudocode_short}) using the BackPACK \citep{dangel2020backpack} package. While PyTorch \citep{NEURIPS2019_9015} can compute efficiently batch gradients, BackPACK optimizes the computation of individual gradients, sample per sample, at almost no time overhead. Thus, Fishr is also at low computational costs. For example, on PACS (7 classes and $|\omega|=14,343$) with a ResNet-50 and batch size 32, Fishr induces an overhead in memory of +0.2\% and in training time of +2.7\% (with a Tesla V100) compared to ERM; on the larger-scale DomainNet (345 classes and $|\omega|=706,905$), the overhead is +7.0\% in memory and +6.5\% in training time. As a side note, keeping the full covariance of size $|\omega|^2\approx5\times10^8$ on DomainNet would not have been possible. In contrast, Fish \cite{shi2021gradient} leverages a meta-learning algorithm that is impractical as $|\mathcal{E}|$ times longer to train than ERM.

\subsubsection{Results}

Table \ref{table:db_all_oracle} summarizes the results on DomainBed using the `Test-domain' model selection: the validation set (to select the best hyperparameters) follows the same distribution as the test domain.
Appendix \ref{appendix:domainbedtrainingdomain} reports results with the `Training-domain' model selection while results are detailed per dataset in Appendix \ref{appendix:domainbedperdataset}.

ERM was carefully tuned in DomainBed and thus remains a strong baseline.
Moreover, all previous methods are far from the best score on at least one dataset.
Invariant predictors (IRM, V-REx) and gradient masking (AND-mask) approaches perform poorly on real datasets.
Additionally, CORAL not only performs worse than ERM on TerraIncognita, but most importantly fails to detect correlation shifts on Colored MNIST: this is because feature-based approaches do not take into account the label, as previously stated in Section \ref{model:fishrmatchrisks}.

Contrarily, Fishr is the \textbf{only method to efficiently tackle correlation and diversity shifts}, as defined in \citep{ye2021odbench}.
Indeed, not only Fishr outperforms ERM on Colored MNIST (68.8\% vs.\ 57.8\%), but Fishr also systematically performs better than ERM on all real datasets: the differences are over standard errors on VLCS (78.2\% vs.\ 77.6\%), OfficeHome (68.2\% vs.\ 66.4\%) and on the larger-scale DomainNet (41.8\% vs.\ 41.3\%). Appendix \ref{appendix:iga} shows that Fishr performs even better when combined with gradient-mean matching.
In summary, \textbf{Fishr consistently beats ERM} (despite the restricted hyperparameter search): this is the main point to validate the effectiveness of our method.

Additionally, Fishr performs best after averaging:
Firshr reaches 70.8\% vs.\ 69.2\% for the second best CORAL.
When ignoring the Colored MNIST task, averaging over the 6 other datasets leads to a similar ranking: 1.Fishr(avg=71.1), 2.CORAL(71.0), 3.Mixup(70.8) and 4.ERM(70.5).
This arguably partial metric is confirmed by the more robust ranking information; Fishr's median ranking of second reflects that \textbf{Fishr is consistently among the best methods}.
Overall, Fishr is the state-of-the-art approach, not only in average accuracy, but most importantly in average ranking.





%% file: pictures/pseudocode_short.tex
\begin{algorithm}[tb]%
    \caption{Training procedure for Fishr on DomainBed.}%
    \label{pseudocode_short}
    \begin{algorithmic}
        \STATE {\bfseries Input:} DNN $f_{\theta}$, observations $\mathcal{D}_{e}=\left\{\left(\vx_{e}^{i}, \vy_{e}^{i}\right)\right\}_{i=1}^{n_{e}}$ for domains $e\in\mathcal{E}$, regularization weight $\lambda$, warmup iteration $i_{\text{warmup}}$, exponential moving average $\gamma$ and batch size $b_s$
        \STATE {\bfseries Initialize:} moving averages: $\forall e\in\mathcal{E}, \vv_e^{\text{mean}}\gets 0$
        \FOR{$\text{iter}$ \textbf{from} $1$ \textbf{to} \#iters}
        \STATE \COMMENT{\# Step 1: standard ERM procedure}
        \FOR{$e\in\mathcal{E}$}
        \STATE Randomly select batch: $\{(\vx_{e}^{i}, \vy_{e}^{i})\}_{i\in \mathcal{B}}$ of size $b_s$
        \STATE Compute predictions: $\forall i\in \mathcal{B}, \hat{\vy}_e^{i} \gets f_{\theta}(\vx_{e}^i)$
        \STATE Compute empirical risks: $\mathcal{R}_{e}(\theta) \gets \sum_{i\in \mathcal{B}} \ell\left(\hat{\vy}_e^{i}, \vy_{e}^{i}\right)$
        \ENDFOR
        \STATE $ \mathcal{L}(\theta)=\frac{1}{|\mathcal{E}|} \sum_{e\in\mathcal{E}} \mathcal{R}_{e}(\theta)$
        \STATE  \COMMENT{\# Step 2: gradient variances in classifier}
        \FOR{$e\in\mathcal{E}$}
        \STATE Compute individual gradients in $w_{\omega}$ with BackPACK: $\forall i\in \mathcal{B}, \vg_{e}^{i} \gets \nabla_{\omega} \ell\left(\hat{\vy}_e^{i}, \vy_{e}^{i}\right)$
        \STATE Compute domain gradient variances $\vv_e$ (Eq.\ \ref{eq:variance})
        \STATE Update $\vv_e^{\text{mean}} = \vv_e \gets \gamma \vv_e^{\text{mean}}+(1-\gamma)\vv_e^{\text{iter}}$
        \ENDFOR
        \IF{$\text{iter} \geq i_{\text{warmup}}$}
        \STATE $ \mathcal{L}(\theta) \mathrel{{+}{=}} \lambda \mathcal{L}_{\text{Fishr}}(\theta)$ (Eq.\ \ref{eq:fishrreg})
        \ENDIF
        \STATE \COMMENT{\# Step 3: gradient descent in the whole network}
        \STATE Backpropagate gradients $\nabla_{\theta} \mathcal{L}(\theta)$ in the network $f_{\theta}$ with standard PyTorch
        \ENDFOR
    \end{algorithmic}
\end{algorithm}%


%% file: tables/oracle/db_all_oracle.tex
\begin{table*}[t]%
    \caption{\textbf{DomainBed benchmark}. We format \textbf{first}, \underline{second} and \textcolor{gray}{worse than ERM} results.}%
    \centering
    \adjustbox{width=1.0\textwidth}{
        \begin{tabular}{l|cccccccc|ccc}
            \toprule
            \multirow{3}{*}{\textbf{Algorithm}} & \multicolumn{8}{c|}{Accuracy ($\uparrow$)}    & \multicolumn{3}{c}{Ranking ($\downarrow$)}                                                                                                                                                                                                                                                                                                                                                            \\
                                                & \textbf{CMNIST}                               & \textbf{RMNIST}                               & \textbf{VLCS}                                 & \textbf{PACS}                                 & \textbf{OfficeHome}                           & \textbf{TerraInc}                             & \textbf{DomainNet}                            & \textbf{Avg}           & \begin{tabular}{@{}c@{}}\textbf{Arith.} \\ \textbf{mean}\end{tabular} & \begin{tabular}{@{}c@{}}\textbf{Geom.} \\ \textbf{mean}\end{tabular} & \textbf{Median}      \\
            \midrule
            ERM                                 & 57.8 \scriptsize{$\pm$ 0.2}                   & 97.8 \scriptsize{$\pm$ 0.1}                   & 77.6 \scriptsize{$\pm$ 0.3}                   & 86.7 \scriptsize{$\pm$ 0.3}                   & 66.4 \scriptsize{$\pm$ 0.5}                   & 53.0 \scriptsize{$\pm$ 0.3}                   & 41.3 \scriptsize{$\pm$ 0.1}                   & 68.7                   & 9.1                       & 8.1                       & 8                    \\
            IRM                                 & \underline{67.7} \scriptsize{$\pm$ 1.2}       & \textcolor{gray}{97.5} \scriptsize{$\pm$ 0.2} & \textcolor{gray}{76.9} \scriptsize{$\pm$ 0.6} & \textcolor{gray}{84.5} \scriptsize{$\pm$ 1.1} & \textcolor{gray}{63.0} \scriptsize{$\pm$ 2.7} & \textcolor{gray}{50.5} \scriptsize{$\pm$ 0.7} & \textcolor{gray}{28.0} \scriptsize{$\pm$ 5.1} & \textcolor{gray}{66.9} & \textcolor{gray}{14.7}    & \textcolor{gray}{12.4}    & \textcolor{gray}{16} \\
            GroupDRO                            & 61.1 \scriptsize{$\pm$ 0.9}                   & 97.9 \scriptsize{$\pm$ 0.1}                   & \textcolor{gray}{77.4} \scriptsize{$\pm$ 0.5} & 87.1 \scriptsize{$\pm$ 0.1}                   & \textcolor{gray}{66.2} \scriptsize{$\pm$ 0.6} & \textcolor{gray}{52.4} \scriptsize{$\pm$ 0.1} & \textcolor{gray}{33.4} \scriptsize{$\pm$ 0.3} & \textcolor{gray}{67.9} & 8.6                       & 7.5                       & 8                    \\
            Mixup                               & 58.4 \scriptsize{$\pm$ 0.2}                   & \underline{98.0} \scriptsize{$\pm$ 0.1}       & 78.1 \scriptsize{$\pm$ 0.3}                   & 86.8 \scriptsize{$\pm$ 0.3}                   & 68.0 \scriptsize{$\pm$ 0.2}                   & \textbf{54.4} \scriptsize{$\pm$ 0.3}          & \textcolor{gray}{39.6} \scriptsize{$\pm$ 0.1} & 69.0                   & 5.3                       & 3.9                       & 4                    \\
            MLDG                                & 58.2 \scriptsize{$\pm$ 0.4}                   & 97.8 \scriptsize{$\pm$ 0.1}                   & \textcolor{gray}{77.5} \scriptsize{$\pm$ 0.1} & 86.8 \scriptsize{$\pm$ 0.4}                   & 66.6 \scriptsize{$\pm$ 0.3}                   & \textcolor{gray}{52.0} \scriptsize{$\pm$ 0.1} & 41.6 \scriptsize{$\pm$ 0.1}                   & 68.7                   & 9.1                       & \textcolor{gray}{8.2}     & \textcolor{gray}{9}  \\
            CORAL                               & 58.6 \scriptsize{$\pm$ 0.5}                   & \underline{98.0} \scriptsize{$\pm$ 0.0}       & 77.7 \scriptsize{$\pm$ 0.2}                   & 87.1 \scriptsize{$\pm$ 0.5}                   & \textbf{68.4} \scriptsize{$\pm$ 0.2}          & \textcolor{gray}{52.8} \scriptsize{$\pm$ 0.2} & \underline{41.8} \scriptsize{$\pm$ 0.1}       & \underline{69.2}       & \underline{4.6}           & \underline{3.4}           & \underline{3}        \\
            MMD                                 & 63.3 \scriptsize{$\pm$ 1.3}                   & \underline{98.0} \scriptsize{$\pm$ 0.1}       & 77.9 \scriptsize{$\pm$ 0.1}                   & \textbf{87.2} \scriptsize{$\pm$ 0.1}          & \textcolor{gray}{66.2} \scriptsize{$\pm$ 0.3} & \textcolor{gray}{52.0} \scriptsize{$\pm$ 0.4} & \textcolor{gray}{23.5} \scriptsize{$\pm$ 9.4} & \textcolor{gray}{66.9} & 7.0                       & 4.9                       & 6                    \\
            DANN                                & \textcolor{gray}{57.0} \scriptsize{$\pm$ 1.0} & 97.9 \scriptsize{$\pm$ 0.1}                   & \underline{79.7} \scriptsize{$\pm$ 0.5}       & \textcolor{gray}{85.2} \scriptsize{$\pm$ 0.2} & \textcolor{gray}{65.3} \scriptsize{$\pm$ 0.8} & \textcolor{gray}{50.6} \scriptsize{$\pm$ 0.4} & \textcolor{gray}{38.3} \scriptsize{$\pm$ 0.1} & \textcolor{gray}{67.7} & \textcolor{gray}{11.9}    & \textcolor{gray}{9.6}     & \textcolor{gray}{15} \\
            CDANN                               & 59.5 \scriptsize{$\pm$ 2.0}                   & 97.9 \scriptsize{$\pm$ 0.0}                   & \textbf{79.9} \scriptsize{$\pm$ 0.2}          & \textcolor{gray}{85.8} \scriptsize{$\pm$ 0.8} & \textcolor{gray}{65.3} \scriptsize{$\pm$ 0.5} & \textcolor{gray}{50.8} \scriptsize{$\pm$ 0.6} & \textcolor{gray}{38.5} \scriptsize{$\pm$ 0.2} & \textcolor{gray}{68.2} & \textcolor{gray}{9.6}     & 7.4                       & \textcolor{gray}{10} \\
            MTL                                 & \textcolor{gray}{57.6} \scriptsize{$\pm$ 0.3} & 97.9 \scriptsize{$\pm$ 0.1}                   & 77.7 \scriptsize{$\pm$ 0.5}                   & 86.7 \scriptsize{$\pm$ 0.2}                   & 66.5 \scriptsize{$\pm$ 0.4}                   & \textcolor{gray}{52.2} \scriptsize{$\pm$ 0.4} & \textcolor{gray}{40.8} \scriptsize{$\pm$ 0.1} & \textcolor{gray}{68.5} & 8.4                       & 7.8                       & 7                    \\
            SagNet                              & 58.2 \scriptsize{$\pm$ 0.3}                   & 97.9 \scriptsize{$\pm$ 0.0}                   & 77.6 \scriptsize{$\pm$ 0.1}                   & \textcolor{gray}{86.4} \scriptsize{$\pm$ 0.4} & 67.5 \scriptsize{$\pm$ 0.2}                   & \textcolor{gray}{52.5} \scriptsize{$\pm$ 0.4} & \textcolor{gray}{40.8} \scriptsize{$\pm$ 0.2} & 68.7                   & 8.0                       & 7.2                       & 6                    \\
            ARM                                 & 63.2 \scriptsize{$\pm$ 0.7}                   & \textbf{98.1} \scriptsize{$\pm$ 0.1}          & 77.8 \scriptsize{$\pm$ 0.3}                   & \textcolor{gray}{85.8} \scriptsize{$\pm$ 0.2} & \textcolor{gray}{64.8} \scriptsize{$\pm$ 0.4} & \textcolor{gray}{51.2} \scriptsize{$\pm$ 0.5} & \textcolor{gray}{36.0} \scriptsize{$\pm$ 0.2} & \textcolor{gray}{68.1} & \textcolor{gray}{9.9}     & 7.5                       & \textcolor{gray}{12} \\
            V-REx                               & 67.0 \scriptsize{$\pm$ 1.3}                   & 97.9 \scriptsize{$\pm$ 0.1}                   & 78.1 \scriptsize{$\pm$ 0.2}                   & \textbf{87.2} \scriptsize{$\pm$ 0.6}          & \textcolor{gray}{65.7} \scriptsize{$\pm$ 0.3} & \textcolor{gray}{51.4} \scriptsize{$\pm$ 0.5} & \textcolor{gray}{30.1} \scriptsize{$\pm$ 3.7} & \textcolor{gray}{68.2} & 7.7                       & 5.5                       & 5                    \\
            RSC                                 & 58.5 \scriptsize{$\pm$ 0.5}                   & \textcolor{gray}{97.6} \scriptsize{$\pm$ 0.1} & 77.8 \scriptsize{$\pm$ 0.6}                   & \textcolor{gray}{86.2} \scriptsize{$\pm$ 0.5} & 66.5 \scriptsize{$\pm$ 0.6}                   & \textcolor{gray}{52.1} \scriptsize{$\pm$ 0.2} & \textcolor{gray}{38.9} \scriptsize{$\pm$ 0.6} & \textcolor{gray}{68.2} & \textcolor{gray}{9.9}     & \textcolor{gray}{9.4}     & \textcolor{gray}{9}  \\
            AND-mask                            & 58.6 \scriptsize{$\pm$ 0.4}                   & \textcolor{gray}{97.5} \scriptsize{$\pm$ 0.0} & \textcolor{gray}{76.4} \scriptsize{$\pm$ 0.4} & \textcolor{gray}{86.4} \scriptsize{$\pm$ 0.4} & \textcolor{gray}{66.1} \scriptsize{$\pm$ 0.2} & \textcolor{gray}{49.8} \scriptsize{$\pm$ 0.4} & \textcolor{gray}{37.9} \scriptsize{$\pm$ 0.6} & \textcolor{gray}{67.5} & \textcolor{gray}{13.4}    & \textcolor{gray}{13.1}    & \textcolor{gray}{12} \\
            SAND-mask                           & 62.3 \scriptsize{$\pm$ 1.0}                   & \textcolor{gray}{97.4} \scriptsize{$\pm$ 0.1} & \textcolor{gray}{76.2} \scriptsize{$\pm$ 0.5} & \textcolor{gray}{85.9} \scriptsize{$\pm$ 0.4} & \textcolor{gray}{65.9} \scriptsize{$\pm$ 0.5} & \textcolor{gray}{50.2} \scriptsize{$\pm$ 0.1} & \textcolor{gray}{32.2} \scriptsize{$\pm$ 0.6} & \textcolor{gray}{67.2} & \textcolor{gray}{14.3}    & \textcolor{gray}{13.5}    & \textcolor{gray}{15} \\
            Fish                                & 61.8 \scriptsize{$\pm$ 0.8}                   & 97.9 \scriptsize{$\pm$ 0.1}                   & 77.8 \scriptsize{$\pm$ 0.6}                   & \textcolor{gray}{85.8} \scriptsize{$\pm$ 0.6} & \textcolor{gray}{66.0} \scriptsize{$\pm$ 2.9} & \textcolor{gray}{50.8} \scriptsize{$\pm$ 0.4} & \textbf{43.4} \scriptsize{$\pm$ 0.3}          & 69.1                   & 8.4                       & 6.6                       & 7                    \\
            \midrule
            Fishr                               & \textbf{68.8} \scriptsize{$\pm$ 1.4}          & 97.8 \scriptsize{$\pm$ 0.1}                   & 78.2 \scriptsize{$\pm$ 0.2}                   & 86.9 \scriptsize{$\pm$ 0.2}                   & \underline{68.2} \scriptsize{$\pm$ 0.2}       & \underline{53.6} \scriptsize{$\pm$ 0.4}       & \underline{41.8} \scriptsize{$\pm$ 0.2}       & \textbf{70.8}          & \textbf{3.9}              & \textbf{2.8}              & \textbf{2}           \\
            \bottomrule
        \end{tabular}
    }
    \label{table:db_all_oracle}%
\end{table*}


%% file: sections/06_conclusion.tex
\section{Conclusion}%
In this paper, we addressed the task of out-of-distribution generalization for classification in computer vision. We derive a new and simple regularization --- Fishr --- that matches the gradient variances across domains as a proxy for matching domain-level risks and Hessians.
We prove that this reduces inconsistencies across domains.
Fishr reaches state-of-the-art performances on DomainBed when samples from the test domain are available for model selection.
Our experiments --- reproducible with our open-source implementation --- suggest that Fishr would consistently improve a deep classifier for real-world usages when dealing with data from multiple domains.
We hope to pave the way towards new gradient-based regularization to improve the generalization abilities of deep neural networks.

%% file: sections/07_appendix.tex
These Appendices complement the main paper.
\begin{enumerate}
\item We first detail some theoretical points. Appendix \ref{appendix:limitincon} demonstrates our Proposition \ref{prop:limitincon}. Appendix \ref{appendix:lineartheory} shows that Fishr acts as a feature-adaptive V-REx. Appendix \ref{appendix:ntk} motivates Fishr with intuitions from the Neural Tangent Kernel theory.
\item Appendix \ref{appendix:linearexpe} proves the effectiveness of our approach for a linear toy dataset.
\item Appendix \ref{appendix:cmnist} enriches the Colored MNIST experiment in the IRM setup. In detail, we first describe the experimental setup in Appendix \ref{appendix:cmnistdetails}. We then validate in Appendix \ref{appendix:cmnistexpeadd} some insights provided in the main paper; in particular, Appendix \ref{appendix:offdiag} motivates the diagonal approximation of the gradient covariance.
\item Appendix \ref{appendix:domainbed} enriches the DomainBed experiments.
After a description of the benchmark protocols in Appendix \ref{appendix:domainbeddetails}, Appendix \ref{appendix:domainbedtrainingdomain} discusses the model selection strategy. Then Appendix \ref{appendix:componentdomainbed} provides additional experiments to analyze key components of Fishr.
Specifically, \ref{appendix:ablationema} analyzes the exponential moving average; \ref{appendix:iga} compares gradient mean versus gradient variance matching and also motivates ignoring the gradients in the features extractor; \ref{appendix:v15} discusses the methodology to select hyperparameter distributions. Finally, Appendix \ref{appendix:domainbedperdataset} provides the per-dataset results.
\end{enumerate}

\section{Additional Theoretical Analysis}
\label{appendix:theory}
\input{sections/appendix/07_appendix_consistency.tex}

\input{sections/appendix/07_appendix_vrex.tex}
\input{sections/appendix/07_appendix_ntk.tex}

\input{sections/appendix/07_linear_justification}
\input{sections/appendix/07_appendix_cmnistdetails.tex}
\section{DomainBed}
\label{appendix:domainbed}
\input{sections/appendix/07_appendix_domainbeddetails.tex}
\subsection{Fishr component analysis on DomainBed}
\label{appendix:componentdomainbed}
\input{sections/appendix/07_appendix_ablationema.tex}

\input{sections/appendix/07_appendix_iga_full.tex}
\input{sections/appendix/07_appendix_domainbedv15.tex}
\input{sections/appendix/07_appendix_domainbed.tex}

%% file: sections/appendix/07_appendix_consistency.tex
\subsection{Demonstration of Proposition \ref{prop:limitincon} from Section \ref{model:inconsistency}}
\label{appendix:limitincon}

\begin{wrapfigure}[16]{h}{0.40\textwidth}
    \vspace{1em}
    \centering
    \includegraphics[width=1.0\linewidth]{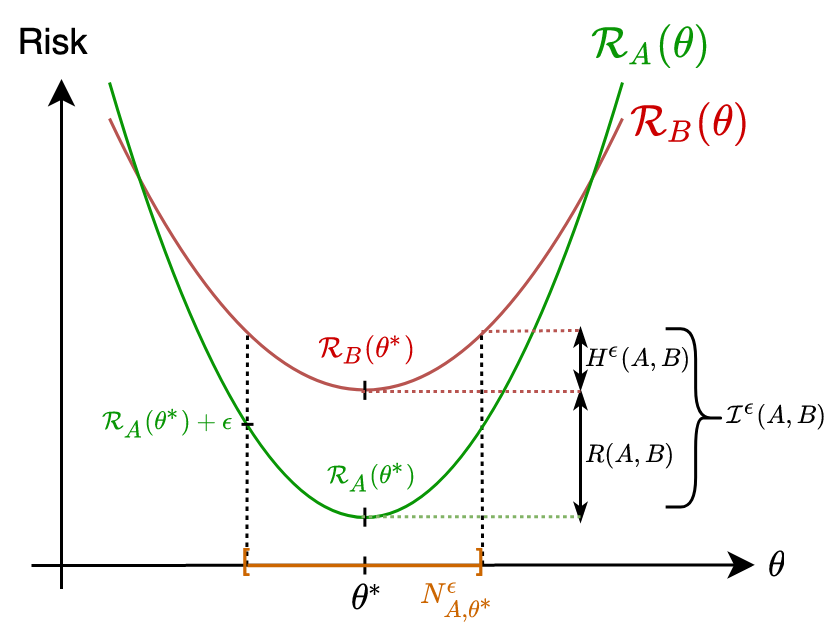}%
    \caption{\textbf{Inconsistency $\mathcal{I}^{\epsilon}(A, B)$ between domains $A$ and $B$}, decomposed into $R(A,B)$ depending on domain-level risks and $H^{\epsilon}(A,B)$ depending on domain-level curvatures at $\theta^{*}$.
    }%
    \label{fig:propconsistency}
\end{wrapfigure}

\begin{assumption}
    We make the quadratic bowl assumption around the local minima $\theta^{*}$ on all domains : $\forall e \in \mathcal{E}$,
    \begin{equation}
        \mathcal{R}_{e}(\theta)=\mathcal{R}_{e}(\theta^{*}) + \frac{1}{2} (\theta-\theta^{*})^{\top} H_{e} (\theta-\theta^{*}),
    \end{equation}
    where $H_{e}$ is positive definite of eigenvalues $\lambda_{1}^{e} \geq \cdots \geq \lambda_{h}^{e}>0$.
    \label{assum:quadratic}
\end{assumption}
\begin{remark}
    Assumption \ref{assum:quadratic} is milder on $ N_{e, \theta^{*}}^{\epsilon}$ for low $\epsilon$. Indeed, when $\epsilon\to0$, then
    $\max_{\theta\in N_{e, \theta^{*}}^{\epsilon}} \left\|\theta-\theta^{*}\right\|_{2}^{2} \to 0$
    and the quadratic approximation coincides with the second-order Taylor expansion around $\theta^{*}$.
    Moreover, this approximation is common in optimization \cite{pmlr-v28-schaul13,jastrzebski2018three}.
\end{remark}
\paragraph{}
\begin{proposition}
    (Reformulation of Proposition \ref{prop:limitincon}, illustrated in Fig.\ \ref{fig:propconsistency}).
    Let $\epsilon>0$, weights $\theta^{*}$. $\forall (A, B)\in\mathcal{E}^2$, with $N_{A, \theta^{*}}^{\epsilon}$ the largest path-connected region of weights space where the risk $\mathcal{R}_{A}$ remains in an ${\epsilon}$ interval around $\mathcal{R}_{A}(\theta^{*})$, we note:
    \begin{equation}
        \begin{split}
            \mathcal{I}^{\epsilon}(A, B) &= \max_{\theta \in N_{A, \theta^{*}}^{\epsilon}}\left|\mathcal{R}_{B}(\theta)-\mathcal{R}_{A}(\theta^{*})\right|,
            \\
            R(A,B) &= \mathcal{R}_{B}(\theta^{*}) - \mathcal{R}_{A}(\theta^{*}),\\
            H^{\epsilon}(A,B) &= \max _{\frac{1}{2} (\theta-\theta^{*})^{\top} H_{A} (\theta-\theta^{*}) \leq \epsilon} \frac{1}{2} (\theta-\theta^{*})^{\top} H_{B} (\theta-\theta^{*}).
        \end{split}
    \end{equation}
    If $\forall (A, B)\in\mathcal{E}^2$ such as $R(A,B)<0$, we have:
    \begin{equation}
        \epsilon \leq -R(A,B) \times \frac{\lambda_{h}^{A}}{\lambda_{1}^{B}},
        \label{eq:eps}
    \end{equation}
    then under previous Assumption \ref{assum:quadratic},
    \begin{equation}
        \max_{(A, B)\in\mathcal{E}^2} \mathcal{I}^{\epsilon}(A, B) = \max_{(A, B)\in\mathcal{E}^2} \left( R(A,B) + H^{\epsilon}(A,B) \right)
    \end{equation}
\end{proposition}

\paragraph{\textit{Proof}}
We first prove that, under quadratic Assumption \ref{assum:quadratic}, $\forall A\in\mathcal{E}, N_{A, \theta^{*}}^{\epsilon} = \left\{\theta | \left|\mathcal{R}_{A}(\theta) - \mathcal{R}_{A}(\theta^{*})\right| \leq \epsilon\right\}$.
Indeed, the former is always included in the latter by definition.
Reciprocally, be given $\theta$ in the latter, $\left\{\lambda\theta^{*} + (1-\lambda)\theta | \lambda \in [0,1] \right\}$
linearly connects $\theta^{*}$ to $\theta$ in parameter space
with the risk $\mathcal{R}_{A}$ remaining in an $\epsilon$ interval around $\mathcal{R}_{A}(\theta^{*})$
because
$\forall \mu \in [0,1]$ we have
$|\mathcal{R}_{A}(\mu\theta^{*} + (1-\mu)\theta) - \mathcal{R}_{A}(\theta^{*})| = (1-\mu)^2|\mathcal{R}_{A}(\theta) - \mathcal{R}_{A}(\theta^{*})| \leq (1-\mu)^2 \epsilon \leq \epsilon$.

Therefore $\forall (A, B) \in \mathcal{E}^2$:
\begin{equation}
    \mathcal{I}^{\epsilon}(A, B) =         \max _{\left|\mathcal{R}_{A}(\theta) - \mathcal{R}_{A}(\theta^{*})\right| \leq \epsilon} \left|\mathcal{R}_{B}(\theta)-\mathcal{R}_{A}(\theta^{*})\right| = \max _{\frac{1}{2} (\theta - \theta^{*})^{\top} H_{A} (\theta-\theta^{*}) \leq \epsilon} \left| R(A,B) + \frac{1}{2} (\theta-\theta^{*})^{\top} H_{B} (\theta-\theta^{*})\right|
    \label{eq:001}
\end{equation}
As the Hessians are positive, $H^{\epsilon}(A,B)>0$. We now need to split the analysis based on the sign of $R(A,B)$.

\paragraph{Case $R(A,B)\geq0$}
Both $R(A,B)$ and $H^{\epsilon}(A,B)$ are non-negative. Removing the absolute value from the RHS of Eq.\ \ref{eq:001} gives:
$
    \mathcal{I}^{\epsilon}(A, B) = R(A,B) + H^{\epsilon}(A,B)
$. Taking the maximum over $(A, B)\in\mathcal{E}^2$ where $R(A,B)\geq0$ gives:
\begin{equation}
\label{eq:00}
\begin{split}
    &\max_{(A, B)\in\mathcal{E}^2|R(A,B)\geq0} \mathcal{I}^{\epsilon}(A, B)\\
    & =\max_{(A, B)\in\mathcal{E}^2|R(A,B)\geq0} \left( R(A,B) + H^{\epsilon}(A,B) \right).
\end{split}
\end{equation}

\paragraph{Case $R(A,B)<0$}
Leveraging $\lambda_{1}^{B}$ the largest eigenvalue from $H_{B}$ and $\lambda_{h}^{A}$ the lowest eigenvalue from $H_{A}$, we upper bound:
\begin{equation}
    H^{\epsilon}(A,B) \leq \max _{\frac{\lambda_{h}^{A}}{2}\|\theta-\theta^{*}\|_{2}^{2} \leq \epsilon} \frac{\lambda_{1}^{B}}{2} \|\theta-\theta^{*}\|_{2}^{2}
    = \epsilon \times \frac{\lambda_{1}^{B}}{\lambda_{h}^{A}}.
\end{equation}
Then Eq.\ \ref{eq:eps} gives $H^{\epsilon}(A,B)<-R(A,B)$.
Thus the number inside the absolute value from the RHS of Eq.\ \ref{eq:001} is negative.
This leads to:
$
    \mathcal{I}^{\epsilon}(A, B) =
    - R(A,B) - H^{\epsilon}(A,B) < - R(A,B) = R(B,A) < \mathcal{I}^{\epsilon}(B, A).
$
Thus the max over $\mathcal{E}^2$ of function $(A, B) \rightarrow \mathcal{I}^{\epsilon}(A, B)$ can not be achieved for $(A, B)$ with $R(A,B)<0$. We obtain:
\begin{equation}
    \max_{(A, B)\in\mathcal{E}^2} \mathcal{I}^{\epsilon}(A, B) = \max_{(A, B)\in\mathcal{E}^2|R(A,B)\geq0} \mathcal{I}^{\epsilon}(A, B)
    \label{eq:01}
\end{equation}

Similarly, $R(A,B) + H^{\epsilon}(A,B)\leq 0 < \mathcal{R}(B,A) + H^{\epsilon}(B,A)$. Thus the max over $\mathcal{E}^2$ of function $(A, B) \rightarrow \left( R(A,B) + H^{\epsilon}(A,B) \right)$ can not be achieved for $(A, B)$ with $R(A,B)<0$. We obtain:
\begin{equation}
    \max_{(A, B)\in\mathcal{E}^2} \left( R(A,B) + H^{\epsilon}(A,B) \right) = \max_{(A, B)\in\mathcal{E}^2|R(A,B)\geq0} \left( R(A,B) + H^{\epsilon}(A,B) \right)
    \label{eq:02}
\end{equation}

\paragraph{Conclusion} Combining Eq.\ \ref{eq:00}, Eq.\ \ref{eq:01} and Eq.\ \ref{eq:02}, we conclude the proof.

%% file: sections/appendix/07_appendix_vrex.tex
\subsection{Fishr as a feature-adaptive version of V-REx}
\label{appendix:lineartheory}
We delve into the theoretical analysis of the Fishr regularization in the classifier $w_{\omega}$, that leverages $p$ features extracted from $\phi$. We note  $z_e^i \in \mathbb{R}^{p}$ the features for the $i$-th example from the domain $e$, $\hat{y}_e^i \in [0, 1]$ the predictions after sigmoid and $y_e^i \in \{0, 1\}$ the one-hot encoded target. The linear layer $W$ is parametrized by weights $\{w_{k}\}_{k=1}^{p}$ and bias $b$.

The gradient of the loss for this sample with respect to the \textbf{bias} $b$ is $\nabla_{b} \ell(y_e^i,\hat{y}_e^i) = (\hat{y}_e^i - y_e^i)$. Thus, the uncentered gradient variance in $b$ for domain $e$ is:
$\vv_e^{b} = \frac{1}{n_{e}} \sum_{i=1}^{n_{e}}(\hat{y}_e^i - y_e^i)^{2}$, which is exactly the mean squared error (MSE) between predictions and targets in domain $e$.
Thus, matching gradient variances in $b$ will match risks across domains. This is the objective from V-REx \citep{krueger2020utofdistribution}, where the squared error has replaced the negative log likelihood.

We can also look at the gradients with respect to the \textbf{weight} $w_{k}$: $\nabla_{w_{k}} \ell(y_e^i,\hat{y}_e^i)  = (\hat{y}_e^i - y_e^i) z_e^i[k]$.
Thus, the uncentered gradient variance in $w_k$ for domain $e$ is:
$\vv_e^{w_{k}} = \frac{1}{n_{e}} \sum_{i=1}^{n_{e}}\left((\hat{y}_e^i - y_e^i) z_e^i[k]\right)^{2}$.
This is the squared error, weighted for each sample ($z_e^i$, $y_e^i$) by the square of the $k$-th feature $z_e^i[k]$: matching gradient variances directly matches these weighted squared errors, with $k$ different weighting schemes, that depend on the features distribution. This describes Fishr as a \textbf{feature-adaptive version of V-REx} \cite{krueger2020utofdistribution}. An intuitive example is when features are binary ($z_e^i \in \{0,1\}$); in that case, Fishr matches domain-level risks on groups of samples having a shared feature.

More exactly in Fishr, we match centered gradient variances,
equivalent to the uncentered variance gradient matching at convergence under the assumption $\vg_{e} \approx 0$.
Experiments in Table \ref{tab:appendix-linear} and in Appendix \ref{appendix:centered} confirm that centering or not the variances perform similarly.

%% file: sections/appendix/07_appendix_ntk.tex
\subsection{Neural Tangent Kernel perspective}
\label{appendix:ntk}

In this Section we motivate the matching of gradient covariances with new arguments from the Neural Tangent Kernel (NTK) \citep{NEURIPS2018_5a4be1fa} theory. As a reminder, the NTK $\mK \in \mathbb{R}^{n\times n}$ is the gramian matrix with entries $\mK[i, j] = \nabla_{\theta} f_{\theta}(x^i)^{\top} \cdot \nabla_{\theta} f_{\theta}(x^j)$ that measure the gradients similarity at two different input points $x^i$ and $x^j$. This kernel dictates the training dynamics of the DNN and remains fixed in the infinite width limit. Most importantly, as stated in \citet{yang2019}, \enquote{the simplicity bias of a wide neural network can be read off quickly from the spectrum of $\mK$: if the largest eigenvalue [$\lambda^{\max }$] of $\mK$ accounts for most of $\operatorname{Tr}(\mK)$, then a typical random network looks like a function from the top eigenspace of $\mK$}: this holds for ReLu networks. In summary, gradient descent mostly happens in a tiny subspace \citep{gur2018gradient} whose directions are defined by the main eigenvectors from $\mK$. Moreover, the learning speed is dictated by $\lambda^{\max }$, which can be used to estimate a condition for a learning rate $\eta$ to converge: $\eta<2 / \lambda^{\max }$  \citep{karakida2019pathological}.

In a multi-domain framework, having similar spectral decompositions across $\{\mK_e\}_{e\in\mathcal{E}}$ during the optimization process would improve OOD generalization for two reasons:
\begin{enumerate}
\item Having similar top eigenvectors across $\{\mK_e\}_{e\in\mathcal{E}}$ would delete detrimental domain-dependent shortcuts and favor the learning of a common mechanism. Indeed, truly informative features should remain consistent across domains.
\item Having similar top eigenvalues across $\{\mK_e\}_{e\in\mathcal{E}}$ would improve the optimization schema for simultaneous training at the same speed. Indeed, it would facilitate the finding of a learning rate for simultaneous convergence on all domains. It's worth noting that if we quickly overfit on a first domain using spurious explanations, invariances will then be hard to learn due to the gradient starvation phenomena \citep{pezeshki2020radient}.
\end{enumerate}

Directly matching $\mK_e$ would require assuming that each domain coincides and contains the same samples; for example, with different pose angles \citep{ghifary2015domain}. To avoid such a strong assumption, we leverage the fact that the `true' Fisher Information Matrix $\mF$ and the NTK $\mK$ share the same non-zero eigenvalues since $\mF$ is dual to $\mK$ (see Appendix C.1 in \citet{maddox2019transfer}, notably for classification tasks). Moreover, their eigenvectors are strongly related (see Appendix C in \citet{kopitkov2019eural}).
Thus, having similar $\{\mF_e\}_{e\in\mathcal{E}}$ encourages $\{\mK_e\}_{e\in\mathcal{E}}$ to have similar spectral decomposition.
Based on the close relations between $\mC$ and $\mF$ (see Section \ref{model:fishrmatchhessians}), this further motivates the need to match gradient variances during the SGD trajectory --- and not only at convergence as in Section \ref{model:theo}.

%% file: sections/appendix/07_linear_justification.tex
\section{Experiments on a Linear Example}
\label{appendix:linearexpe}


We experimentally prove that Fishr is effective in the linear setting. To do so, we consider the binary classification dataset introduced in the Section 3.2 from Fish \citep{shi2021gradient}. Each example is composed of 4 static features ($f_1$, $f_2$, $f_3$, $f_4$). While $f_1$ is invariant across the two train domains and the test domain, the three other features are spurious: their correlations with the label vary in each domain.
The model is a linear logistic regression, with trainable weights $W$ and bias $b$.
As $f_2$ and $f_3$ have higher correlations with the label than $f_1$ in training, ERM relies mostly on $f_2$ and $f_3$. This is indicated in the first line of Table \ref{tab:appendix-linear} by the large values (3.3) for weights associated to $f_2$ and $f_3$; this induces low test accuracy (57\%). On the contrary, Fishr forces the linear model to rely mostly on the invariant feature $f_1$, as indicated by the lower values (1.2) for weights associated to $f_2$ and $f_3$; in accuracy, Fishr performs similarly in test and train (93\%).

\begin{table}[h!]
\centering
\adjustbox{width=0.65\textwidth}{
\begin{tabular}{cccccc}
    \toprule Method & Matched statistics & Train acc. & Test acc. & $W$ & $b$ \\
    \midrule
ERM & N/A & 97 \% & 57 \% & [2.8,3.3,3.3,0.0] & -2.7 \\
Fish & Gradient means & 93 \% & 93 \% & [0.4,0.2,0.2,0.0] & -0.4 \\
Fishr & Centered gradient variances & 93 \% & 93 \% & [2.0,1.2,1.2,0.0] & -0.6 \\
Fishr & Uncentered gradient variances & 93 \% & 93 \% & [1.9,0.9,0.9,0.0] & -0.6 \\
    \bottomrule
\end{tabular}
}
\caption{ \textbf{Performances comparison on the linear dataset} from \citep{shi2021gradient}}
\label{tab:appendix-linear}
\end{table}





%% file: sections/appendix/07_appendix_cmnistdetails.tex
\section{Colored MNIST in the IRM Setup}
\label{appendix:cmnist}

\subsection{Description of the Colored MNIST experiment}
\label{appendix:cmnistdetails}

Colored MNIST is a binary digit classification dataset introduced in IRM \citep{arjovsky2019invariant}. Compared to the traditional MNIST \citep{lecun2010mnist}, it has 2 main differences. \textit{First}, 0-4 and 5-9 digits are each collapsed into a single class, with a 25\% chance of label flipping. \textit{Second}, digits are either colored red or green, with a strong correlation between label and color in training. However, this correlation is reversed at test time. Specifically, in training, the model has access to two domains $\mathcal{E}=\{90\%,80\%\}$: in the first domain, green digits have a 90\% chance of being in 5-9; in the second, this chance goes down to 80\%. In test, green digits have a 10\% chance of being in 5-9.
Due to this modification in correlation, a model should ideally ignore the color information and only rely on the digits' shape: this would obtain a 75\% test accuracy.

In the experimental setup from IRM, the network is a 3 layers MLP with ReLu activation, optimized with Adam \citep{kingma2014adam}. IRM selected the following hyperparameters by random search over 50 trials: hidden dimension of 390, $l_2$ regularizer weight of 0.00110794568, learning rate of 0.0004898536566546834, penalty anneal iters (or warmup iter) of 190, penalty weight ($\lambda$) of 91257.18613115903, 501 epochs and batch size 25,000 (half of the dataset size). We strictly keep the same hyperparameters values in our proof of concept in Section \ref{expe:irmcmnist}.
The code is almost unchanged from \url{https://github.com/facebookresearch/InvariantRiskMinimization}.
\subsection{Empirical validation of some key insights}
\label{appendix:cmnistexpeadd}


\subsubsection{Hessian matching}

\label{appendix:hessianmatching}
Based on empirical works \citep{li2020hessian,NEURIPS2020_d1ff1ec8,interplayinfomatrix2020}, we argue in Section \ref{model:fishrmatchhessians} that gradient covariance $\mC$ can be used as a proxy to regularize the Hessian $\mH$ ---  even though the proper approximation bounds are out of scope of this paper.
This was empirically validated at convergence in Table \ref{table:hessianermfishr} and during training in Fig.\ \ref{fig:fishrhessianacc}.
We leveraged the \textit{DiagHessian} method from BackPACK to compute Hessian diagonals, in all network weights $\theta$.
Notably, Hessians are impractical in a training objective as computing \enquote{Hessian is an order of magnitude more computationally intensive} (see Fig.\ 9 in \citet{dangel2020backpack}).
This Appendix further analyzes the Hessian trajectory during training.

Fig.\ \ref{fig:hessianl2cos} illustrates the dynamics for Fishr$_{\theta}$: following the scheduling previously described in Appendix \ref{appendix:cmnistdetails}, $\lambda$ jumping to a high value at epoch 190 activates the regularization.
After this epoch, the domain-level Hessians are not only close in Frobenius distance, but also have similar norms and directions.
On the contrary, when using only ERM in Fig.\ \ref{fig:hessianl2coserm}, the distance between domain-level Hessians keeps increasing with the number of epochs.
As a side note, flatter loss landscapes in ERM --- as reflected by the Hessian norms in orange --- do not correlate with improved generalization \citep{pmlr-v70-dinh17b}.
\input{pictures/model/cmnist_dynamics_hessian_fishrerm.tex}
\clearpage
\input{pictures/model/cmnist_erm}

This is also visible in Fig.\ \ref{fig:ermhessianacc}, which is equivalent to Fig.\ \ref{fig:fishrhessianacc}, but for ERM (without the Fishr regularization). The distance between domain-level gradient variances (red) keeps increasing across domains $\mathcal{E}=\{90\%, 80\%\}$: so does the distance across Hessians (purple). The distance across risks (pink) decreases, but slower than with Fishr regularization. Overall, the network still predicts the digit's color while only slightly using the digit's shape. That's why the test accuracy (blue) remains low.
\input{tables/irm/cmnist_0.tex}
\vspace{2em}
\subsubsection{Colored MNIST without label flipping}
\label{appendix:cmnistclean}
To further validate that Fishr can tackle distribution shifts, we investigate Colored MNIST but without the 25\% label flipping.
In Table \ref{table:cmnist_0}, the label is then fully predictable from the digit shape.
Using hyperparameters defined previously in Appendix \ref{appendix:cmnistdetails}, we recover that IRM (82.2\%) fails when the invariant feature is fully predictive \citep{ahuja2021invariance}: indeed, it performs worse than ERM (91.8\%). In contrast, V-REx and Fishr$_{\omega}$ perform better (95.3\%): in conclusion, Fishr works even without label noise.

\vspace{2em}
\subsubsection{Gradient variance or covariance ?}
\label{appendix:offdiag}
We have justified ignoring the off-diagonal parts of the covariance to reduce the memory overhead.
For the sake of completeness, the second line in Table \ref{table:cmnistend} shows results with the full covariance matrix.
This experiment is possible only when considering gradient in the classifier $w_{\omega}$ for memory reasons.
Overall, results are similar (or slightly worse) as when using only the diagonal: the slight difference may be explained by the approaches' different suitability to the hyperparameters (that were optimized for IRM).
In conclusion, this preliminary experiment suggests that targeting the diagonal components is the most critical. We hope future works will further investigate this diagonal approximation or provide new methods to reduce the computational costs, such as K-FAC approximations \citep{heskes2000natural,10.5555/3045118.3045374}.

\input{tables/irm/cmnist_end.tex}

\subsubsection{Centered or uncentered variance ?}
\label{appendix:centered}
In Section \ref{model:fishrmatchhessians}, we argue that the gradient centered covariance $\mC$ and the empirical Fisher Information Matrix (or uncentered covariance) $\tilde{\mF}$ are highly related and equivalent when the DNN is at convergence and the gradient means are zero. So, we could have tackled the diagonals of the domain-level $\{\tilde{\mF}_e\}_{e\in\mathcal{E}}$ across domains, \textit{i.e.}, without centering the variances.
Empirically, comparing the first and third lines in Table \ref{table:cmnistend} shows that centering or not the variance are almost equivalent.
This holds true when applying Fishr on all weights $\theta$ (as lines fourth and six are also very similar).
This was empirically confirmed in DomainBed: for example, Fishr with either centered or uncentered variances reach 67.8.
Still, it’s worth noting that explicitly matching simultaneously the gradient centered variances along with the gradient means performs best in Appendix \ref{appendix:iga}.


%% file: pictures/model/cmnist_dynamics_hessian_fishrerm.tex
\begin{figure}[h]
    \centering
    \includegraphics[width=.85\linewidth]{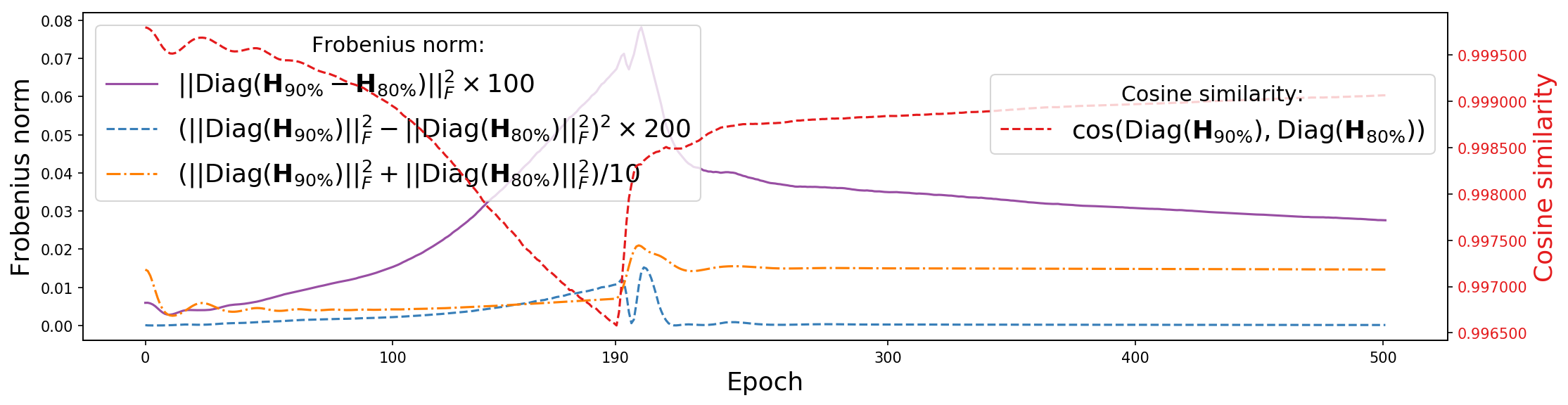}
    \caption{\textbf{Hessian dynamics on Colored MNIST with Fishr}: at epoch 190, $\lambda$ steps up. Then domain-level Hessians are matched across domains (purple). More precisely, they take similar directions --- high cosine similarity (red) --- and similar norms (blue). The Hessians' norms (orange) remain quite high thus the loss landscapes are rather sharp.}
    \label{fig:hessianl2cos}
\end{figure}

\begin{figure}[h]
    \centering
    \includegraphics[width=.85\linewidth]{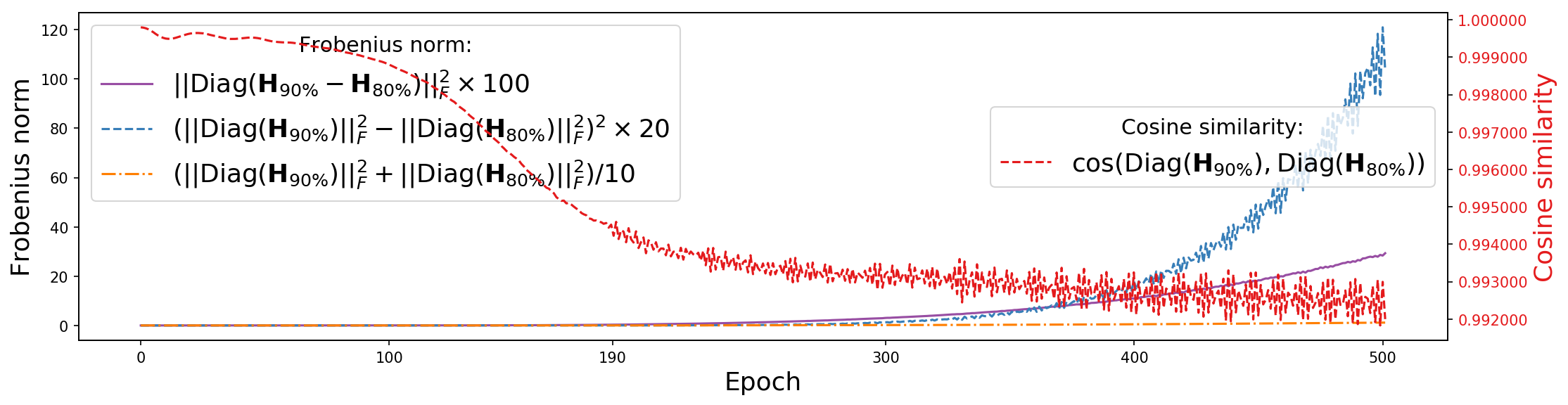}
    \caption{\textbf{Hessian dynamics on Colored MNIST with ERM}: $\lambda=0$ along training.
        The Frobenius distance between domain-level Hessians (purple) keeps increasing: so does the distance between their norms (blue).
        Their cosine similarity (red) steadily decreases.
        The loss landscapes are flat at convergence (low Hessian norms in orange).
    }
    \label{fig:hessianl2coserm}
\end{figure}

%% file: pictures/model/cmnist_erm.tex
\begin{wrapfigure}[9]{hR!}{0.40\textwidth}
    \centering
    \includegraphics[width=1.0\linewidth]{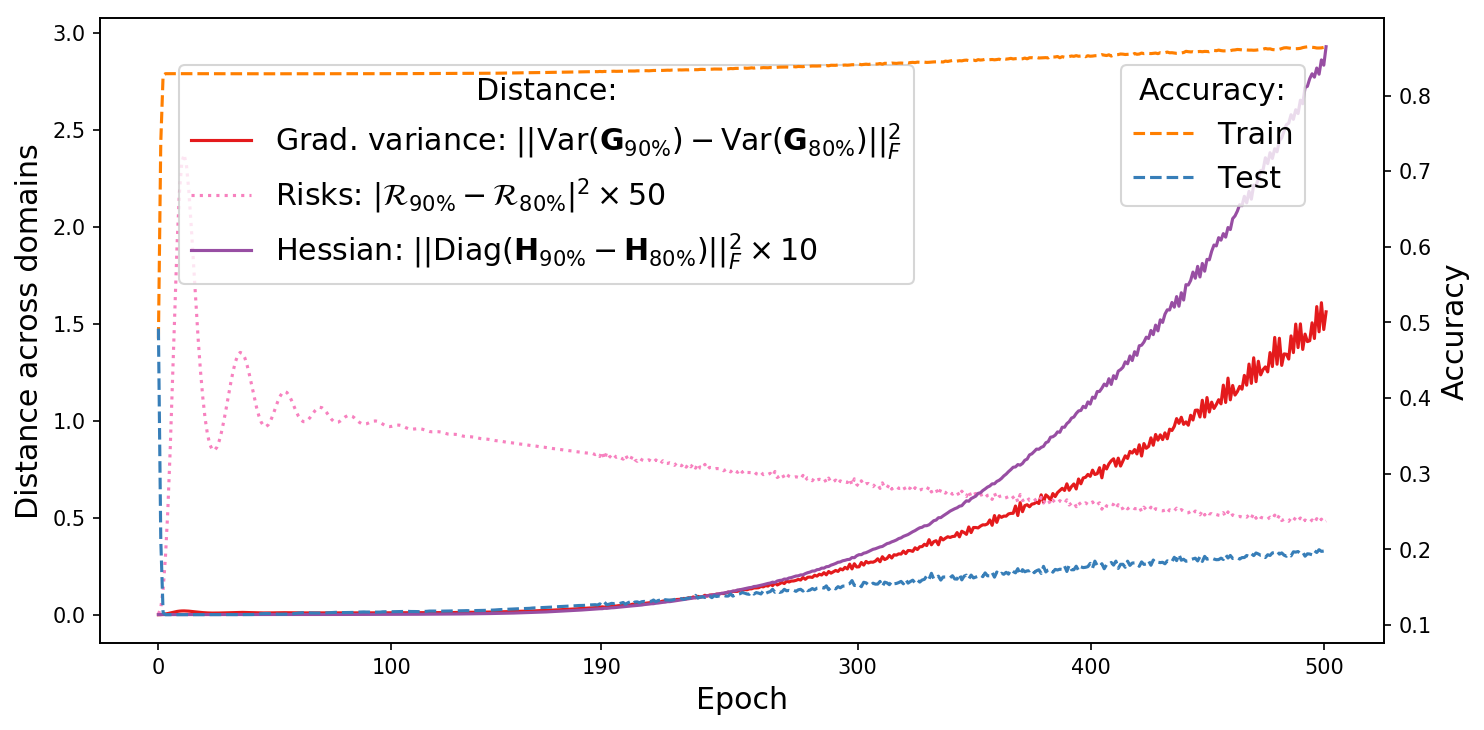}%
    \vspace{-0.5em}
    \caption{\textbf{Colored MNIST dynamics with ERM}.
    }
    \label{fig:ermhessianacc}%
\end{wrapfigure}

%% file: tables/irm/cmnist_0.tex
\begin{wraptable}[8]{hR!}{0.40\textwidth}
    \centering
    \vspace{-1.em}
    \centering
    \caption{\textbf{Colored MNIST experiments without label flipping}.}
    \resizebox{1.0\linewidth}{!}{%
        \begin{tabular}{c  c c c}
            \toprule
            Method           & Train acc.     & Test acc.      & Gray test acc. \\
            \midrule
            ERM              & 99.0 $\pm$ 0.0 & 91.8 $\pm$ 0.2 & 95.0 $\pm$ 0.4 \\
            IRM              & 96.4 $\pm$ 0.2 & 82.2 $\pm$ 0.1 & 92.6 $\pm$ 0.2 \\
            V-REx            & 97.1 $\pm$ 0.2 & 95.3 $\pm$ 0.4 & 94.1 $\pm$ 0.4 \\
            \midrule
            Fishr$_{\theta}$ & 97.9 $\pm$ 0.2 & 93.6 $\pm$ 0.4 & 94.8 $\pm$ 0.4 \\
            Fishr$_{\omega}$ & 97.0 $\pm$ 0.2 & 95.3 $\pm$ 0.4 & 94.1 $\pm$ 0.4 \\
            Fishr$_{\phi}$   & 97.9 $\pm$ 0.1 & 93.5 $\pm$ 0.3 & 94.8 $\pm$ 0.4 \\
            \bottomrule
        \end{tabular}%
    }
    \label{table:cmnist_0}%
\end{wraptable}



%% file: tables/irm/cmnist_end.tex
\begin{table}[!h]
    \caption{\textbf{Colored MNIST experiments} with different statistics matched. All hyperparameters were optimized for IRM.}
    \centering
    \resizebox{0.85\textwidth}{!}{%
        \begin{tabular}{c c c | c c c | c c c}
            \toprule
            \multicolumn{3}{c|}{Method} & \multicolumn{3}{c|}{25\% label flipping} & \multicolumn{3}{c}{No label flipping}                                                                                                                   \\
            \midrule
            Gradients in                & Name                                     & Matched statistics                                & Train acc.     & Test acc.      & Gray test acc. & Train acc.     & Test acc.      & Gray test acc. \\
            \midrule
            \multirow{3}{*}{$\omega$}   & Centered variance (= Fishr$_{\omega}$)   & $\Var(\mG_e)$                                     & 71.0 $\pm$ 0.9 & 69.5 $\pm$ 1.0 & 70.2 $\pm$ 1.1 & 97.0 $\pm$ 0.2 & 95.3 $\pm$ 0.4 & 94.1 $\pm$ 0.4 \\
                                        & Centered covariance                      & $\mC_e$                                           & 70.7 $\pm$ 1.0 & 69.1 $\pm$ 1.1 & 69.9 $\pm$ 1.1 & 97.0 $\pm$ 0.2 & 95.3 $\pm$ 0.4 & 94.0 $\pm$ 0.4 \\
                                        & Uncentered variance                      & $\operatorname{Diag}(\frac{1}{n_e}\tilde{\mF}_e)$ & 71.3 $\pm$ 0.9 & 69.5 $\pm$ 1.0 & 70.3 $\pm$ 1.0 & 97.0 $\pm$ 0.2 & 95.3 $\pm$ 0.4 & 94.1 $\pm$ 0.4 \\
            \midrule
            \multirow{3}{*}{$\theta$}   & Centered variance (= Fishr$_{\theta}$)   & $\Var(\mG_e)$                                     & 69.6 $\pm$ 0.9 & 71.2 $\pm$ 1.1 & 70.2 $\pm$ 0.7 & 97.9 $\pm$ 0.1 & 93.5 $\pm$ 0.3 & 94.7 $\pm$ 0.4 \\
                                        & Centered covariance                      & $\mC_e$                                           & Not            & possible       & for            & computational  & (memory)       & reasons        \\
                                        & Uncentered variance                      & $\operatorname{Diag}(\frac{1}{n_e}\tilde{\mF}_e)$ & 71.0 $\pm$ 0.8 & 70.0 $\pm$ 1.1 & 70.1 $\pm$ 0.9 & 97.9 $\pm$ 0.0 & 93.5 $\pm$ 0.3 & 94.8 $\pm$ 0.4 \\
            \midrule
            \multirow{3}{*}{$\phi$}     & Centered variance (= Fishr$_{\phi}$)     & $\Var(\mG_e)$                                     & 65.6 $\pm$ 1.3 & 73.8 $\pm$ 1.0 & 70.0 $\pm$ 0.9 & 97.9 $\pm$ 0.1 & 93.5 $\pm$ 0.3 & 94.8 $\pm$ 0.4 \\
                                        & Centered covariance                      & $\mC_e$                                           & Not            & possible       & for            & computational  & (memory)       & reasons        \\
                                        & Uncentered variance                      & $\operatorname{Diag}(\frac{1}{n_e}\tilde{\mF}_e)$ & 71.5 $\pm$ 0.8 & 69.1 $\pm$ 1.1 & 70.0 $\pm$ 1.0 & 97.9 $\pm$ 0.1 & 93.5 $\pm$ 0.3 & 94.8 $\pm$ 0.4 \\
            \bottomrule
        \end{tabular}%
    }
    \label{table:cmnistend}%
\end{table}

%% file: sections/appendix/07_appendix_domainbeddetails.tex
\subsection{Description of the DomainBed benchmark}
\label{appendix:domainbeddetails}

We now further detail our experiments on the DomainBed benchmark.
Scores from most baselines are taken from the DomainBed \citep{gulrajani2021in} paper.
Scores for AND-mask and SAND-mask are taken from the SAND-mask paper \citep{shahtalebi2021andmask}.
Scores for IGA \citep{koyama2020out} are not yet available: yet, for the sake of completeness, we analyze IGA in Appendix \ref{appendix:iga}.
Missing scores will be included when available.

The same procedure was applied for all methods: for each domain, a random hyperparameter search of 20 trials over a joint distribution, described in Table \ref{table:hyperparameters}, is performed.
We discuss the choice of these distributions in Appendix \ref{appendix:v15}.
The learning rate, the batch size (except for ARM), the weight decay and the dropout distributions are shared across all methods - all trained with Adam \citep{kingma2014adam}.
Specific hyperparameter distributions for concurrent methods can be found in the original work of \citet{gulrajani2021in}.
The data from each domain is split into 80\% (used as training and testing) and 20\% (used as validation for hyperparameter selection) splits. This random process is repeated with 3 different seeds:
the reported numbers are the means and the standard errors over these 3 seeds.

\input{tables/domainbed/hyperparameters_search.tex}

We clarify a subtle point (omitted in the Algorithm \ref{pseudocode_short}) concerning the hyperparameter $\gamma$ that controls: $\bar{\vv}_e^{t}=\gamma \bar{\vv}_e^{t-1}+(1-\gamma)\vv_e^{t}$ at step $t$.
We remind that $\bar{\vv}_e^{t-1}$ from previous step $t-1$ is `detached' from the computational graph.
Thus when $\mathcal{L}$ from Eq.\ \ref{eq:loss} is differentiated during SGD, the gradients going through $\vv_e^{t}$ are multiplied by $(1-\gamma)$.
To compensate this and decorrelate the impact of $\gamma$ and of $\lambda$ (that controls the regularization strength), we match $\frac{1}{1-\gamma}\bar{\vv}_e^{t}$.
Finally, with this $(1-\gamma)$ \textbf{correction}, the gradients' strength backpropagated in the network is independent of $\gamma$.%

\paragraph{}
Here we list all \textbf{concurrent approaches}.
\begin{itemize}
    \item ERM: Empirical Risk Minimization \citep{vapnik1999overview}
    \item IRM: Invariant Risk Minimization \citep{arjovsky2019invariant}
    \item GroupDRO: Group Distributionally Robust Optimization \citep{Sagawa2020Distributionally}
    \item Mixup: Interdomain Mixup \citep{yan2020improve}
    \item MLDG: Meta Learning Domain Generalization \citep{li2018learning}
    \item CORAL: Deep CORAL \citep{sun2016deep}
    \item MMD: Maximum Mean Discrepancy \citep{li2018domain}
    \item DANN: Domain Adversarial Neural Network \citep{ganin2016domain}
    \item CDANN: Conditional Domain Adversarial Neural Network \citep{li2018domaincond}
    \item MTL: Marginal Transfer Learning \citep{JMLR:v22:17-679}
    \item SagNet: Style Agnostic Networks \citep{nam2021reducing}
    \item ARM: Adaptive Risk Minimization \citep{zhang2020daptive}
    \item V-REx: Variance Risk Extrapolation \citep{krueger2020utofdistribution}
    \item RSC: Representation Self-Challenging \citep{huang2020self}
    \item AND-mask: Learning Explanations that are Hard to Vary \citep{parascandolo2021learning}
    \item SAND-mask: An Enhanced Gradient Masking Strategy for the Discovery of Invariances in Domain Generalization \citep{shahtalebi2021andmask}
    \item IGA: Out-of-distribution generalization with maximal invariant predictor \citep{koyama2020out}
    \item Fish: Gradient Matching for Domain Generalization \citep{shi2021gradient}
\end{itemize}
We omitted the recent weight averaging approaches \citep{cha2021wad,rame2022diwa} whose contribution is complementary to others, that uses a custom hyperparameter search and does not report scores with the `Test-domain' model selection.

\paragraph{}
DomainBed includes seven multi-domain computer vision classification \textbf{datasets}:
\begin{enumerate}
    \item Colored MNIST \citep{arjovsky2019invariant} is a variant of the MNIST handwritten digit classification dataset \citep{lecun2010mnist}. As described previously in Appendix \ref{appendix:cmnistdetails}, domain $d\in\{90\%, 80\%, 10\%\}$ contains a disjoint set of digits colored: the correlation strengths between color and label vary across domains. The dataset contains 70,000 examples of dimension $(2,28,28)$ and 2 classes. Most importantly, the network, the hyperparameters, the image shapes, etc. are \textbf{not} the same as in the IRM setup from Section \ref{expe:irmcmnist}.
    \item Rotated MNIST \citep{ghifary2015domain} is a variant of MNIST where domain $d\in\{0,15,30,45,60,75\}$ contains digits rotated by $d$ degrees, with 70,000 examples of dimension $(1,28,28)$ and 10 classes.
    \item VLCS \citep{fang2013unbiased} includes photographic domains $d\in\{$Caltech101, LabelMe, SUN09, VOC2007$\}$, with 10,729 examples of dimension $(3,224,224)$ and 5 classes.
    \item PACS \citep{li2017deeper} includes domains $d\in\{$art, cartoons, photos, sketches$\}$, with 9,991 examples of dimension $(3,224,224)$ and 7 classes.
    \item OfficeHome \citep{venkateswara2017deep} includes domains $d\in\{$art, clipart, product, real$\}$, with 15,588 examples of dimension $(3,224,224)$ and 65 classes.
    \item TerraIncognita \citep{beery2018recognition} contains photographs of wild animals taken by camera traps at locations $d\in\{$L100, L38, L43, L46$\}$, with 24,788 examples of dimension $(3,224,224)$ and 10 classes.
    \item DomainNet \citep{peng2019moment} has six domains $d\in\{$clipart, infograph, painting, quickdraw, real, sketch$\}$, with 586,575 examples of size $(3,224,224)$ and 345 classes.
\end{enumerate}
The convolutional neural network architecture used for the MNIST experiments is the one introduced in DomainBed: note that this is not the same MLP (described in Appendix \ref{appendix:cmnistdetails}) as in our proof of concept in Section \ref{expe:irmcmnist}. All real datasets leverage a `ResNet-50' pretrained on ImageNet, with a dropout layer before the newly added dense layer and fine-tuned with frozen batch normalization layers.




\subsection{`Training-domain' model selection}
\label{appendix:domainbedtrainingdomain}

\input{tables/training/db_all_training.tex}
In the main paper, we focus on the `Test-domain' model selection, where the validation set follows the same distribution as the test domain.
This is important to adapt the degree of model invariance according to the test domain. For Fishr, if the domain-dependant correlations are useful in test, the selected $\lambda$ would be small and Fishr would behave like ERM; in contrast, if the domain-dependant correlations are detrimental in test, the selected $\lambda$ would be large, and Fishr would improve over ERM by enforcing invariance.

In Table \ref{table:db_all_training}, we use the `Training-domain' model selection: the validation set is formed by randomly collecting 20\% of each training domain.
Fishr performs better than ERM on all real datasets (over standard errors for OfficeHome and DomainNet), except for PACS where the two reach 85.5\%.
In average, Fishr (67.1\%) finishes third and is above most methods such as V-REx (65.6\%). Fishr median ranking is fifth, with a mean ranking of 5.6.
These additional results were not included in the main paper due to space constraints and also because this `Training-domain' model selection has three clear limitations.

\textit{First}, learning causal mechanisms can be useless in this `Training-domain’ setup. Indeed, when the correlations are more predictive in training than the causal features, the variant model may be selected over the invariant one. This explains the poor results for all methods in `Training-domain’ Colored MNIST, where the color information is more predictive than the shape information in training.
The best model on this task is ARM \citep{zhang2020daptive} that uses test time adaptation - thus in a sense uses information from the test-domain - and whose contribution is mostly complementary to ours.

\textit{Second}, the `Training-domain' setup suffers from
underspecification: \enquote{predictors with equivalently strong held-out performance in the training domain [...] can behave very differently} in test \citep{d2020underspecification}.
This underspecification favors low regularization thus low values of $\lambda$.
To select the model with the best generalization properties, future benchmarks may consider the training calibration \citep{wald2021n}
rather than merely selecting the model with the best training accuracy.

\textit{Third}, the `Test-domain' model selection is more realistic for real applications.
Indeed, one user would easily label some samples to validate the efficiency of its algorithm. It’s not realistic to believe that the users would simply deploy their new algorithm without at least checking that the performances are correct.
We recall that the `Test-domain' setup in DomainBed benchmark is quite restricting, allowing only one evaluation per choice of hyperparameters, without early-stopping.

That’s why \citet{teneybias} even states that \enquote{OOD performance cannot, by definition, be performed with a validation set from the same distribution as the training data}.
Both opinions being reasonable and arguable, we included ‘Training-domain’ results for the sake of completeness, where Fishr remains stronger than ERM.
Yet, our state-of-the-art results on the ‘Test-domain’ setup from Table \ref{table:db_all_oracle} alone are sufficient to prove the usefulness of our approach for real-world applications.

%% file: tables/domainbed/hyperparameters_search.tex
\begin{table}[h]
    \caption{\textbf{Hyperparameters}, their default values and distributions for random search.}
    \centering
    \resizebox{.75\textwidth}{!}{%
        \begin{tabular}{llll}
            \toprule
            Condition        & Parameter                         & Default value & Random distribution                                                         \\
            \midrule
            VLCS / PACS /    & learning rate                     & $0.00005$     & $10^{\text{Uniform}(-5,-3.5)}$                                              \\
            OfficeHome /     & batch size                        & 32            & $2^{\text{Uniform}(3,5.5)}$ if not DomainNet else $2^{\text{Uniform}(3,5)}$ \\
            TerraIncognita / & weight decay                      & 0             & $10^{\text{Uniform}(-6,-2)}$                                                \\
            DomainNet        & dropout                           & 0             & RandomChoice $([0,0.1,0.5])$                                                \\
            \midrule
            Rotated MNIST /  & learning rate                     & $0.001$       & $10^{\text{Uniform}(-4.5,-3.5)}$                                            \\
            Colored MNIST    & batch size                        & 64            & $2^{\text{Uniform}(3,9)}$                                                   \\
                             & weight decay                      & 0             & 0                                                                           \\
            \midrule
            All              & steps                             & 5000          & 5000                                                                        \\
            \midrule
            \midrule
            \multirow{3}{*}{Fishr}            & regularization strength $\lambda$ & 1000          & $10^{\text{Uniform}(1,4)}$                                                  \\
                             & ema $\gamma$                      & 0.95          & Uniform$(0.9,0.99)$                                                         \\
                             & warmup iterations                 & 1500          & Uniform$(0,5000)$                                                           \\
            \bottomrule
        \end{tabular}}

    \label{table:hyperparameters}%
\end{table}

%% file: tables/training/db_all_training.tex
\begin{table}[!h]%
    \caption{\textbf{DomainBed with `Training-domain' model selection}. We format \textbf{first}, \underline{second} and \textcolor{gray}{worse than ERM} results.}
    \centering
    \adjustbox{width=0.8\textwidth}{
        \begin{tabular}{l|cccccccc|ccc}
            \toprule
            \multirow{3}{*}{\textbf{Algorithm}} & \multicolumn{8}{c|}{Accuracy ($\uparrow$)}    & \multicolumn{3}{c}{Ranking ($\downarrow$)}                                                                                                                                                                                                                                                                                                                                                                                                         \\
                                                & \textbf{CMNIST}                               & \textbf{RMNIST}                               & \textbf{VLCS}                                 & \textbf{PACS}                                 & \textbf{OfficeHome}                           & \textbf{TerraInc}                             & \textbf{DomainNet}                            & \textbf{Avg}           & \begin{tabular}{@{}c@{}}\textbf{Arith.} \\ \textbf{mean}\end{tabular} & \begin{tabular}{@{}c@{}}\textbf{Geom.} \\ \textbf{mean}\end{tabular} & \textbf{Median}                                                   \\

            \midrule
            ERM                                 & 51.5 \scriptsize{$\pm$ 0.1}                   & \underline{98.0} \scriptsize{$\pm$ 0.0}       & 77.5 \scriptsize{$\pm$ 0.4}                   & 85.5 \scriptsize{$\pm$ 0.2}                   & 66.5 \scriptsize{$\pm$ 0.3}                   & 46.1 \scriptsize{$\pm$ 1.8}                   & 40.9 \scriptsize{$\pm$ 0.1}                   & 66.6                   & 7.0                       & 5.9                       & 7                    \\
            IRM                                 & 52.0 \scriptsize{$\pm$ 0.1}                   & \textcolor{gray}{97.7} \scriptsize{$\pm$ 0.1} & \underline{78.5} \scriptsize{$\pm$ 0.5}       & \textcolor{gray}{83.5} \scriptsize{$\pm$ 0.8} & \textcolor{gray}{64.3} \scriptsize{$\pm$ 2.2} & 47.6 \scriptsize{$\pm$ 0.8}                   & \textcolor{gray}{33.9} \scriptsize{$\pm$ 2.8} & \textcolor{gray}{65.4} & \textcolor{gray}{10.7}    & \textcolor{gray}{8.5}     & \textcolor{gray}{14} \\
            GroupDRO                            & \underline{52.1} \scriptsize{$\pm$ 0.0}       & \underline{98.0} \scriptsize{$\pm$ 0.0}       & \textcolor{gray}{76.7} \scriptsize{$\pm$ 0.6} & \textcolor{gray}{84.4} \scriptsize{$\pm$ 0.8} & \textcolor{gray}{66.0} \scriptsize{$\pm$ 0.7} & \textcolor{gray}{43.2} \scriptsize{$\pm$ 1.1} & \textcolor{gray}{33.3} \scriptsize{$\pm$ 0.2} & \textcolor{gray}{64.8} & \textcolor{gray}{11.3}    & \textcolor{gray}{8.4}     & \textcolor{gray}{14} \\
            Mixup                               & \underline{52.1} \scriptsize{$\pm$ 0.2}       & \underline{98.0} \scriptsize{$\pm$ 0.1}       & \textcolor{gray}{77.4} \scriptsize{$\pm$ 0.6} & \textcolor{gray}{84.6} \scriptsize{$\pm$ 0.6} & 68.1 \scriptsize{$\pm$ 0.3}                   & \underline{47.9} \scriptsize{$\pm$ 0.8}       & \textcolor{gray}{39.2} \scriptsize{$\pm$ 0.1} & 66.7                   & 5.7                       & 4.2                       & \underline{3}        \\
            MLDG                                & 51.5 \scriptsize{$\pm$ 0.1}                   & \textcolor{gray}{97.9} \scriptsize{$\pm$ 0.0} & \textcolor{gray}{77.2} \scriptsize{$\pm$ 0.4} & \textcolor{gray}{84.9} \scriptsize{$\pm$ 1.0} & 66.8 \scriptsize{$\pm$ 0.6}                   & 47.7 \scriptsize{$\pm$ 0.9}                   & 41.2 \scriptsize{$\pm$ 0.1}                   & 66.7                   & \textcolor{gray}{8.0}     & \textcolor{gray}{7.0}     & \textcolor{gray}{8}  \\
            CORAL                               & 51.5 \scriptsize{$\pm$ 0.1}                   & \underline{98.0} \scriptsize{$\pm$ 0.1}       & \textbf{78.8} \scriptsize{$\pm$ 0.6}          & \underline{86.2} \scriptsize{$\pm$ 0.3}       & \textbf{68.7} \scriptsize{$\pm$ 0.3}          & 47.6 \scriptsize{$\pm$ 1.0}                   & 41.5 \scriptsize{$\pm$ 0.1}                   & \textbf{67.5}          & \textbf{3.6}              & \textbf{2.5}              & \textbf{2}           \\
            MMD                                 & 51.5 \scriptsize{$\pm$ 0.2}                   & \textcolor{gray}{97.9} \scriptsize{$\pm$ 0.0} & 77.5 \scriptsize{$\pm$ 0.9}                   & \textcolor{gray}{84.6} \scriptsize{$\pm$ 0.5} & \textcolor{gray}{66.3} \scriptsize{$\pm$ 0.1} & \textcolor{gray}{42.2} \scriptsize{$\pm$ 1.6} & \textcolor{gray}{23.4} \scriptsize{$\pm$ 9.5} & \textcolor{gray}{63.3} & \textcolor{gray}{12.3}    & \textcolor{gray}{11.8}    & \textcolor{gray}{10} \\
            DANN                                & 51.5 \scriptsize{$\pm$ 0.3}                   & \textcolor{gray}{97.8} \scriptsize{$\pm$ 0.1} & 78.6 \scriptsize{$\pm$ 0.4}                   & \textcolor{gray}{83.6} \scriptsize{$\pm$ 0.4} & \textcolor{gray}{65.9} \scriptsize{$\pm$ 0.6} & 46.7 \scriptsize{$\pm$ 0.5}                   & \textcolor{gray}{38.3} \scriptsize{$\pm$ 0.1} & \textcolor{gray}{66.1} & \textcolor{gray}{10.3}    & \textcolor{gray}{8.8}     & \textcolor{gray}{12} \\
            CDANN                               & 51.7 \scriptsize{$\pm$ 0.1}                   & \textcolor{gray}{97.9} \scriptsize{$\pm$ 0.1} & 77.5 \scriptsize{$\pm$ 0.1}                   & \textcolor{gray}{82.6} \scriptsize{$\pm$ 0.9} & \textcolor{gray}{65.8} \scriptsize{$\pm$ 1.3} & \textcolor{gray}{45.8} \scriptsize{$\pm$ 1.6} & \textcolor{gray}{38.3} \scriptsize{$\pm$ 0.3} & \textcolor{gray}{65.6} & \textcolor{gray}{11.1}    & \textcolor{gray}{10.7}    & \textcolor{gray}{10} \\
            MTL                                 & \textcolor{gray}{51.4} \scriptsize{$\pm$ 0.1} & \textcolor{gray}{97.9} \scriptsize{$\pm$ 0.0} & \textcolor{gray}{77.2} \scriptsize{$\pm$ 0.4} & \textcolor{gray}{84.6} \scriptsize{$\pm$ 0.5} & \textcolor{gray}{66.4} \scriptsize{$\pm$ 0.5} & \textcolor{gray}{45.6} \scriptsize{$\pm$ 1.2} & \textcolor{gray}{40.6} \scriptsize{$\pm$ 0.1} & \textcolor{gray}{66.2} & \textcolor{gray}{10.9}    & \textcolor{gray}{10.2}    & \textcolor{gray}{10} \\
            SagNet                              & 51.7 \scriptsize{$\pm$ 0.0}                   & \underline{98.0} \scriptsize{$\pm$ 0.0}       & 77.8 \scriptsize{$\pm$ 0.5}                   & \textbf{86.3} \scriptsize{$\pm$ 0.2}          & 68.1 \scriptsize{$\pm$ 0.1}                   & \textbf{48.6} \scriptsize{$\pm$ 1.0}          & \textcolor{gray}{40.3} \scriptsize{$\pm$ 0.1} & \underline{67.2}       & \underline{4.0}           & \underline{3.0}           & \underline{3}        \\
            ARM                                 & \textbf{56.2} \scriptsize{$\pm$ 0.2}          & \textbf{98.2} \scriptsize{$\pm$ 0.1}          & 77.6 \scriptsize{$\pm$ 0.3}                   & \textcolor{gray}{85.1} \scriptsize{$\pm$ 0.4} & \textcolor{gray}{64.8} \scriptsize{$\pm$ 0.3} & \textcolor{gray}{45.5} \scriptsize{$\pm$ 0.3} & \textcolor{gray}{35.5} \scriptsize{$\pm$ 0.2} & \textcolor{gray}{66.1} & \textcolor{gray}{8.7}     & 5.6                       & \textcolor{gray}{9}  \\
            V-REx                               & 51.8 \scriptsize{$\pm$ 0.1}                   & \textcolor{gray}{97.9} \scriptsize{$\pm$ 0.1} & 78.3 \scriptsize{$\pm$ 0.2}                   & \textcolor{gray}{84.9} \scriptsize{$\pm$ 0.6} & \textcolor{gray}{66.4} \scriptsize{$\pm$ 0.6} & 46.4 \scriptsize{$\pm$ 0.6}                   & \textcolor{gray}{33.6} \scriptsize{$\pm$ 2.9} & \textcolor{gray}{65.6} & \textcolor{gray}{8.3}     & \textcolor{gray}{7.7}     & \textcolor{gray}{8}  \\
            RSC                                 & 51.7 \scriptsize{$\pm$ 0.2}                   & \textcolor{gray}{97.6} \scriptsize{$\pm$ 0.1} & \textcolor{gray}{77.1} \scriptsize{$\pm$ 0.5} & \textcolor{gray}{85.2} \scriptsize{$\pm$ 0.9} & \textcolor{gray}{65.5} \scriptsize{$\pm$ 0.9} & 46.6 \scriptsize{$\pm$ 1.0}                   & \textcolor{gray}{38.9} \scriptsize{$\pm$ 0.5} & \textcolor{gray}{66.1} & \textcolor{gray}{11.4}    & \textcolor{gray}{10.6}    & \textcolor{gray}{9}  \\
            AND-mask                            & \textcolor{gray}{51.3} \scriptsize{$\pm$ 0.2} & \textcolor{gray}{97.6} \scriptsize{$\pm$ 0.1} & 78.1 \scriptsize{$\pm$ 0.9}                   & \textcolor{gray}{84.4} \scriptsize{$\pm$ 0.9} & \textcolor{gray}{65.6} \scriptsize{$\pm$ 0.4} & \textcolor{gray}{44.6} \scriptsize{$\pm$ 0.3} & \textcolor{gray}{37.2} \scriptsize{$\pm$ 0.6} & \textcolor{gray}{65.5} & \textcolor{gray}{13.6}    & \textcolor{gray}{12.7}    & \textcolor{gray}{15} \\
            SAND-mask                           & 51.8 \scriptsize{$\pm$ 0.2}                   & \textcolor{gray}{97.4} \scriptsize{$\pm$ 0.1} & \textcolor{gray}{77.4} \scriptsize{$\pm$ 0.2} & \textcolor{gray}{84.6} \scriptsize{$\pm$ 0.9} & \textcolor{gray}{65.8} \scriptsize{$\pm$ 0.4} & \textcolor{gray}{42.9} \scriptsize{$\pm$ 1.7} & \textcolor{gray}{32.1} \scriptsize{$\pm$ 0.6} & \textcolor{gray}{64.6} & \textcolor{gray}{13.4}    & \textcolor{gray}{12.7}    & \textcolor{gray}{13} \\
            Fish                                & 51.6 \scriptsize{$\pm$ 0.1}                   & \underline{98.0} \scriptsize{$\pm$ 0.0}       & 77.8 \scriptsize{$\pm$ 0.3}                   & 85.5 \scriptsize{$\pm$ 0.3}                   & \underline{68.6} \scriptsize{$\pm$ 0.4}       & \textcolor{gray}{45.1} \scriptsize{$\pm$ 1.3} & \textbf{42.7} \scriptsize{$\pm$ 0.2}          & 67.1                   & 5.6                       & 3.8                       & \underline{3}        \\
            \midrule
            Fishr                               & 52.0 \scriptsize{$\pm$ 0.2}                   & \textcolor{gray}{97.8} \scriptsize{$\pm$ 0.0} & 77.8 \scriptsize{$\pm$ 0.1}                   & 85.5 \scriptsize{$\pm$ 0.4}                   & 67.8 \scriptsize{$\pm$ 0.1}                   & 47.4 \scriptsize{$\pm$ 1.6}                   & \underline{41.7} \scriptsize{$\pm$ 0.0}       & 67.1                   & 5.6                       & 4.8                       & 5                    \\
            \bottomrule
        \end{tabular}
    }
    \label{table:db_all_training}%
\end{table}

%% file: sections/appendix/07_appendix_ablationema.tex
\subsubsection{Focus on the exponential moving average}
\label{appendix:ablationema}

Following \citet{le201115}, we use an exponential moving average (ema) parameterized by $\gamma$ for computing gradient variances in DomainBed:
the closer $\gamma$ is to 1, the longer a batch will impact the variance from later steps.
We now further analyze the impact of this strategy, which is not specific to Fishr and was used previously in other works \citep{NEURIPS2020_eddc3427,JMLR:v22:17-679,zhang2021deep} for OOD generalization.
Notably, this ema strategy could be applied to better estimate domain-level empirical risks in V-REx \citep{krueger2020utofdistribution}.
For a fair comparison, we introduce a new approach --- V-REx with ema --- that penalizes $|\bar{\mathcal{R}}_A^{t} - \bar{\mathcal{R}}_B^{t}|^2$ at step $t$ where $\bar{\mathcal{R}}_e^{t}=\gamma \bar{\mathcal{R}}_e^{t-1} + (1-\gamma)\mathcal{R}_e^{t}$ when $\mathcal{E}=\left\{A,B\right\}$.

Thus, we compare V-REx and Fishr, with $\gamma=0$ (\xmark) or with $\gamma\sim \text{Uniform}(0.9,0.99)$ (\cmark, as described in Table \ref{table:hyperparameters}).
On the synthetic Colored MNIST in Table \ref{table:ablationemacmnist}, the ema is critical for Fishr ---
notably when training on $\mathcal{E}=\{90\%,80\%\}$ and the dataset 10\% is in test (from \xmark 34.0\% to \cmark 58.9\% in `Test-domain'). V-REx also benefits from ema.
On the `real' dataset OfficeHome in Table \ref{table:ablationemahome}, the ema is less beneficial (from \xmark 67.5\% to \cmark 68.2\% in `Test-domain' for Fishr). Notably, it worsens V-REX. Overall, Fishr --- with and without ema --- outperforms V-REx on OfficeHome.

We speculate that ema mainly helps when the batch size is not sufficiently large to detect `slight' correlation shifts in the training datasets: \textit{e.g.}, when batch size $\sim2^{\text{Uniform}(3,9)}$ and training datasets $\mathcal{E}=\{90\%,80\%\}$ in Colored MNIST.
We remind that when the batch size was 25,000 in the Colored MNIST setup from IRM, Fishr reached 69.5\% (without ema) in Table \ref{table:cmnist_25} from Section \ref{expe:irmcmnist}.
On the contrary, when the shift is more prominent as in OfficeHome, the ema may be less necessary.
Most importantly, Fishr --- with and without ema --- improves over ERM on these datasets.

\input{tables/domainbed/table_ablationema.tex}

%% file: tables/domainbed/table_ablationema.tex
\begin{table}[t]%
    \caption{\textbf{Importance of the exponential moving average (ema)} on DomainBed's Colored MNIST.}%
    \centering
    \adjustbox{width=0.7\textwidth}{
        \begin{tabular}{llcccccccc}
            \toprule
            \textbf{Model selection}         & \textbf{Algorithm}     & \textbf{ema} & \textbf{+90\%}          & \textbf{+80\%}          & \textbf{10\%}           & \textbf{Avg}  \\
            \midrule
            \multirow{5}{*}{Test-domain}     & ERM                    & N/A          & 71.8 $\pm$ 0.4          & 72.9 $\pm$ 0.1          & 28.7 $\pm$ 0.5          & 57.8          \\
            \cmidrule{2-7}
                                             & \multirow{2}{*}{V-REx} & \xmark       & 72.8 $\pm$ 0.3          & 73.0 $\pm$ 0.3          & 55.2 $\pm$ 4.0          & 67.0          \\
                                             &                        & \cmark       & 73.0 $\pm$ 0.2          & 73.0 $\pm$ 0.3          & \textbf{59.9} $\pm$ 2.6 & 68.6          \\
            \cmidrule{2-7}
                                             & \multirow{2}{*}{Fishr} & \xmark       & 72.7 $\pm$ 0.3          & 72.8 $\pm$ 0.1          & 34.0 $\pm$ 4.5          & 59.8          \\
                                             &                        & \cmark       & \textbf{74.1} $\pm$ 0.6 & \textbf{73.3} $\pm$ 0.1 & 58.9 $\pm$ 3.7          & \textbf{68.8} \\
            \midrule
            \multirow{5}{*}{Training-domain} & ERM                    & N/A          & 71.7 $\pm$ 0.1          & 72.9 $\pm$ 0.2          & 10.0 $\pm$ 0.1          & 51.5          \\
            \cmidrule{2-7}
                                             & \multirow{2}{*}{V-REx} & \xmark       & 72.4 $\pm$ 0.3          & 72.9 $\pm$ 0.4          & \textbf{10.2} $\pm$ 0.0 & 51.8          \\
                                             &                        & \cmark       & \textbf{72.6} $\pm$ 0.5 & 73.3 $\pm$ 0.1          & 9.8 $\pm$ 0.1           & 51.9          \\
            \cmidrule{2-7}
                                             & \multirow{2}{*}{Fishr} & \xmark       & 71.1 $\pm$ 0.6          & \textbf{73.6} $\pm$ 0.1 & 10.1 $\pm$ 0.2          & 51.6          \\
                                             &                        & \cmark       & 72.3 $\pm$ 0.9          & 73.5 $\pm$ 0.2          & 10.1 $\pm$ 0.2          & \textbf{52.0} \\
            \bottomrule
        \end{tabular}
    }
    \label{table:ablationemacmnist}
\end{table}

\begin{table}[t]%
    \caption{\textbf{Importance of the exponential moving average (ema)} on DomainBed's OfficeHome.}%
    \centering
    \adjustbox{width=0.7\textwidth}{
        \begin{tabular}{llcccccccc}
            \toprule
            \textbf{Model selection}         & \textbf{Algorithm}     & \textbf{ema} & \textbf{A}              & \textbf{C}              & \textbf{P}              & \textbf{R}              & \textbf{Avg}  \\
            \midrule
            \multirow{5}{*}{Test-domain}     & ERM                    & N/A          & 61.7 $\pm$ 0.7          & 53.4 $\pm$ 0.3          & 74.1 $\pm$ 0.4          & 76.2 $\pm$ 0.6          & 66.4          \\
            \cmidrule{2-8}

                                             & \multirow{2}{*}{V-REx} & \xmark       & 59.6 $\pm$ 1.0          & 53.3 $\pm$ 0.3          & 73.2 $\pm$ 0.5          & 76.6 $\pm$ 0.4          & 65.7          \\
                                             &                        & \cmark       & 59.0 $\pm$ 0.7          & 52.8 $\pm$ 0.8          & 74.6 $\pm$ 0.4          & 75.5 $\pm$ 0.3          & 65.5          \\
            \cmidrule{2-8}
                                             & \multirow{2}{*}{Fishr} & \xmark       & \textbf{63.6} $\pm$ 0.4 & 53.2 $\pm$ 0.5          & 75.4 $\pm$ 0.5          & 77.8 $\pm$ 0.3          & 67.5          \\
                                             &                        & \cmark       & 63.4 $\pm$ 0.8          & \textbf{54.2} $\pm$ 0.3 & \textbf{76.4} $\pm$ 0.3 & \textbf{78.5} $\pm$ 0.2 & \textbf{68.2} \\
            \midrule
            \multirow{5}{*}{Training-domain} & ERM                    & N/A          & 61.3 $\pm$ 0.7          & 52.4 $\pm$ 0.3          & 75.8 $\pm$ 0.1          & 76.6 $\pm$ 0.3          & 66.5          \\
            \cmidrule{2-8}
                                             & \multirow{2}{*}{V-REx} & \xmark       & 60.7 $\pm$ 0.9          & 53.0 $\pm$ 0.9          & 75.3 $\pm$ 0.1          & 76.6 $\pm$ 0.5          & 66.4          \\
                                             &                        & \cmark       & 59.2 $\pm$ 1.0          & 51.7 $\pm$ 0.5          & 75.2 $\pm$ 0.2          & 76.6 $\pm$ 0.3          & 65.7          \\
            \cmidrule{2-8}
                                             & \multirow{2}{*}{Fishr} & \xmark       & 62.2 $\pm$ 1.0          & 53.5 $\pm$ 0.2          & \textbf{76.6} $\pm$ 0.2 & 77.8 $\pm$ 0.4          & 67.5          \\
                                             &                        & \cmark       & \textbf{62.4} $\pm$ 0.5 & \textbf{54.4} $\pm$ 0.4 & 76.2 $\pm$ 0.5          & \textbf{78.3} $\pm$ 0.1 & \textbf{67.8} \\
            \bottomrule
        \end{tabular}
    }
    \label{table:ablationemahome}
\end{table}

%% file: sections/appendix/07_appendix_iga_full.tex
\input{tables/domainbed/table_iga_full.tex}

\subsubsection{Component analysis by comparing gradient variance versus gradient mean matching}
\label{appendix:iga}

As a reminder from the Section \ref{relatedwork}, IGA  \citep{koyama2020out} is an unpublished gradient-based approach that matches gradient means across domains, \textit{i.e.},
minimizes $||\vg_{A} - \vg_{B}||_{2}^2$ when $\mathcal{E}=\left\{A,B\right\}$
and where $\vg_{e}=\frac{1}{n_{e}} \sum_{i=1}^{n_{e}} \nabla_{\theta} \ell\left(f_{\theta}(\vx_{e}), \vy_{e}\right)$. Scores for IGA are not available publicly and thus were not included in Section \ref{expe:domainbed}.
Moreover, IGA is very costly and impractical: IGA is approximately $\left(|\mathcal{E}| + 1\right)$ times longer to train than ERM. Yet, we ran the DomainBed implementation of IGA on one `synthetic' and one `real' dataset. Table \ref{table:igacmnist} shows that the IGA has little effect on Colored MNIST (58.0\% vs.\ 57.8\% for ERM in `Test-domain'). Moreover, on OfficeHome in Table \ref{table:igahome}, IGA hinders learning (56.9\% vs.\ 66.4\% for ERM in `Test-domain'). In brief, the seminal \enquote{IGA \textelp{} could completely fail when generalizing to unseen domains}, as stated in Fish \citep{shi2021gradient}.

In the rest of this Section, we include IGA in Fishr codebase so that both methods leverage the same implementation choices: this enables \textbf{fairer comparisons between gradient mean matching and gradient variance matching}.
These experiments provide further insights regarding Fishr main components: specifically, enforcing invariance (1) only in the classifier's weights $\omega$  (2) after a warmup period and (3) with an exponential moving average.

\textit{First}, Fishr only considers gradient variances in the classifier's weights $\omega$. Similarly, we try to apply IGA's gradient mean matching but only in $w_{\omega}$ rather than in $f_{\theta}$.
This new method works significantly better (67.2\% when $\vg_{e}=\frac{1}{n_{e}} \sum_{i=1}^{n_{e}} \nabla_{\omega} \ell\left(f_{\theta}(\vx_{e}), \vy_{e}\right)$ vs.\ 56.9\% when $\vg_{e}=\frac{1}{n_{e}} \sum_{i=1}^{n_{e}} \nabla_{\theta} \ell\left(f_{\theta}(\vx_{e}), \vy_{e}\right)$ for `Test-domain' OfficeHome in Table \ref{table:igahome}) while reducing the computational overhead.
This further motivates the \textbf{invariance in the classifier rather than in the low-level layers} (which need to adapt to shifts in pixels for instance).
We have done this analysis on IGA and not on Fishr because keeping all individual gradients for a ResNet-50 in the GPU memory was not possible on our hardware.

\textit{Second}, Fishr uses a double-stage scheduling inherited from IRM \citep{arjovsky2019invariant}: the DNN first learns predictive features with standard ERM ($\lambda=0$) until a given epoch, at which $\lambda$ takes its true (high) value to then force domain invariance. \textbf{This warmup strategy} slightly increases `Test-domain' results on Colored MNIST (from 58.6\% to 59.8\% for Fishr, from 58.3\% to 59.2\% for IGA) but does not seem critical: in particular, it reduces IGA `Test-domain' scores on OfficeHome.

\textit{Third}, the estimation of gradient variances was improved with an \textbf{exponential moving average} (see Section \ref{expe:domainbed} and Appendix \ref{appendix:ablationema}).
We now use this strategy with domain-level gradient means for IGA in $\omega$: $\bar{\vg}_e^{t}=\gamma \bar{\vg}_e^{t-1} + (1-\gamma)\vg_e^{t}$.
This improves IGA (from 67.0\% to 67.2\% in `Test-domain' on OfficeHome): yet, these scores remain consistently worse than Fishr's (from 67.5\% to 68.2\%).

In conclusion, this complements the experiments in Section \ref{expe:domainbed} which showed that tackling gradient variance does better than tackling gradient mean: indeed, Fishr performed better than Fish \citep{shi2021gradient}, AND-mask \citep{parascandolo2021learning} and SAND-mask \citep{shahtalebi2021andmask}.
As a final note, Fishr + IGA --- \textit{i.e.}, matching simultaneously gradient means (the first moment) and variances (the second moment) --- performs best.
Future works may further analyze the complementary of these gradient-based methods.

%% file: tables/domainbed/table_iga_full.tex
\begin{table}[ht]%
    \caption{\textbf{Fishr (gradient variance) vs.\ IGA (gradient mean)} on DomainBed's Colored MNIST.}%
    \centering
    \adjustbox{width=0.7\textwidth}{
        \begin{tabular}{llllcccccccc}
            \toprule
            \textbf{Model selection}         & \textbf{Algorithm}     & \textbf{Gradients in}       & \textbf{Warmup} & \textbf{ema} & \textbf{+90\%}          & \textbf{+80\%}          & \textbf{10\%}           & \textbf{Avg}  \\
            \midrule
            \multirow{9}{*}{Test-domain}     & ERM                    & N/A                         & N/A             & N/A          & 71.8 $\pm$ 0.4          & 72.9 $\pm$ 0.1          & 28.7 $\pm$ 0.5          & 57.8          \\
            \cmidrule{2-9}
                                             & \multirow{4}{*}{IGA}   & $\theta=\omega \oplus \phi$ & \xmark          & \xmark       & 71.8 $\pm$ 0.5          & 73.0 $\pm$ 0.3          & 29.2 $\pm$ 0.5          & 58.0          \\
                                             &                        & $\omega$                    & \xmark          & \xmark       & 72.4 $\pm$ 0.1          & \textbf{73.3} $\pm$ 0.2 & 29.3 $\pm$ 0.6          & 58.3          \\
                                             &                        & $\omega$                    & \cmark          & \xmark       & 72.5 $\pm$ 0.2          & \textbf{73.3} $\pm$ 0.1 & 31.8 $\pm$ 0.7          & 59.2          \\
                                             &                        & $\omega$                    & \cmark          & \cmark       & 72.6 $\pm$ 0.3          & 72.9 $\pm$ 0.2          & 50.0 $\pm$ 1.2          & 65.2          \\
            \cmidrule{2-9}
                                             & \multirow{3}{*}{Fishr} & \multirow{3}{*}{ $\omega$}  & \xmark          & \xmark       & 73.0 $\pm$ 0.3          & 73.2 $\pm$ 0.1          & 29.5 $\pm$ 1.1          & 58.6          \\
                                             &                        &                             & \cmark          & \xmark       & 72.7 $\pm$ 0.3          & 72.8 $\pm$ 0.1          & 34.0 $\pm$ 4.5          & 59.8          \\
                                             &                        &                             & \cmark          & \cmark       & \textbf{74.1} $\pm$ 0.6 & \textbf{73.3} $\pm$ 0.1 & 58.9 $\pm$ 3.7          & 68.8          \\
            \cmidrule{2-9}
                                             & Fishr + IGA            & $\omega$                    & \cmark          & \cmark       & 73.3 $\pm$ 0.0          & 72.6 $\pm$ 0.5          & \textbf{66.3} $\pm$ 2.9 & \textbf{70.7} \\
            \midrule
            \multirow{9}{*}{Training-domain} & ERM                    & N/A                         & N/A             & N/A          & 71.7 $\pm$ 0.1          & 72.9 $\pm$ 0.2          & 10.0 $\pm$ 0.1          & 51.5          \\
            \cmidrule{2-9}
                                             & \multirow{4}{*}{IGA}   & $\theta=\omega \oplus \phi$ & \xmark          & \xmark       & 71.8 $\pm$ 0.3          & 73.2 $\pm$ 0.2          & 9.8 $\pm$ 0.0           & 51.6          \\
                                             &                        & $\omega$                    & \xmark          & \xmark       & 71.8 $\pm$ 0.1          & 73.2 $\pm$ 0.2          & \textbf{10.1} $\pm$ 0.0 & 51.7          \\
                                             &                        & $\omega$                    & \cmark          & \xmark       & 71.8 $\pm$ 0.2          & 73.1 $\pm$ 0.2          & \textbf{10.1} $\pm$ 0.0 & 51.7          \\
                                             &                        & $\omega$                    & \cmark          & \cmark       & \textbf{72.5} $\pm$ 0.4 & 73.3 $\pm$ 0.2          & \textbf{10.1} $\pm$ 0.1 & \textbf{52.0} \\
            \cmidrule{2-9}
                                             & \multirow{3}{*}{Fishr} & \multirow{3}{*}{$\omega$}   & \xmark          & \xmark       & 71.6 $\pm$ 0.1          & 73.2 $\pm$ 0.1          & 9.9 $\pm$ 0.0           & 51.6          \\
                                             &                        &                             & \cmark          & \xmark       & 71.1 $\pm$ 0.6          & \textbf{73.6} $\pm$ 0.1 & \textbf{10.1} $\pm$ 0.2 & 51.6          \\
                                             &                        &                             & \cmark          & \cmark       & 72.3 $\pm$ 0.9          & 73.5 $\pm$ 0.2          & \textbf{10.1} $\pm$ 0.2 & \textbf{52.0} \\
            \cmidrule{2-9}
                                             & Fishr + IGA            & $\omega$                    & \cmark          & \cmark       & 72.4 $\pm$ 0.4          & 73.1 $\pm$ 0.1          & \textbf{10.1} $\pm$ 0.1 & 51.8          \\
            \bottomrule
        \end{tabular}
    }
    \label{table:igacmnist}
\end{table}
\begin{table}[h]%
    \caption{\textbf{Fishr (gradient variance) vs.\ IGA (gradient mean)} on DomainBed's OfficeHome.}%
    \centering
    \adjustbox{width=0.7\textwidth}{
        \begin{tabular}{llllcccccccc}
            \toprule
            \textbf{Model selection}         & \textbf{Algorithm}     & \textbf{Gradients in}       & \textbf{Warmup} & \textbf{ema} & \textbf{A}              & \textbf{C}              & \textbf{P}              & \textbf{R}              & \textbf{Avg}  \\
            \midrule
            \multirow{9}{*}{Test-domain}     & ERM                    & N/A                         & N/A             & N/A          & 61.7 $\pm$ 0.7          & 53.4 $\pm$ 0.3          & 74.1 $\pm$ 0.4          & 76.2 $\pm$ 0.6          & 66.4          \\
            \cmidrule{2-10}
                                             & \multirow{4}{*}{IGA}   & $\theta=\omega \oplus \phi$ & \xmark          & \xmark       & 50.1 $\pm$ 2.5          & 49.6 $\pm$ 1.6          & 59.5 $\pm$ 6.7          & 68.5 $\pm$ 1.2          & 56.9          \\
                                             &                        & $\omega$                    & \xmark          & \xmark       & 62.3 $\pm$ 0.3          & 53.9 $\pm$ 0.2          & 75.2 $\pm$ 0.4          & 77.4 $\pm$ 0.1          & 67.2          \\
                                             &                        & $\omega$                    & \cmark          & \xmark       & 61.9 $\pm$ 0.4          & 52.6 $\pm$ 0.6          & 76.0 $\pm$ 0.8          & 77.5 $\pm$ 0.3          & 67.0          \\
                                             &                        & $\omega$                    & \cmark          & \cmark       & 62.3 $\pm$ 1.0          & 53.4 $\pm$ 0.3          & 76.0 $\pm$ 0.7          & 77.0 $\pm$ 0.1          & 67.2          \\
            \cmidrule{2-10}
                                             & \multirow{3}{*}{Fishr} & \multirow{3}{*}{$\omega$}   & \xmark          & \xmark       & 61.8 $\pm$ 0.9          & 53.8 $\pm$ 0.4          & \textbf{76.6} $\pm$ 0.6 & 77.7 $\pm$ 0.2          & 67.5          \\
                                             &                        &                             & \cmark          & \xmark       & \textbf{63.6} $\pm$ 0.4 & 53.2 $\pm$ 0.5          & 75.4 $\pm$ 0.5          & 77.8 $\pm$ 0.3          & 67.5          \\
                                             &                        &                             & \cmark          & \cmark       & 63.4 $\pm$ 0.8          & 54.2 $\pm$ 0.3          & 76.4 $\pm$ 0.3          & \textbf{78.5} $\pm$ 0.2 & 68.2          \\
            \cmidrule{2-10}
                                             & Fishr + IGA            & $\omega$                    & \cmark          & \cmark       & \textbf{63.6} $\pm$ 1.0 & \textbf{54.6} $\pm$ 0.5 & \textbf{76.6} $\pm$ 0.2 & 78.4 $\pm$ 0.4          & \textbf{68.3} \\
            \midrule
            \multirow{9}{*}{Training-domain} & ERM                    & N/A                         & N/A             & N/A          & 61.3 $\pm$ 0.7          & 52.4 $\pm$ 0.3          & 75.8 $\pm$ 0.1          & 76.6 $\pm$ 0.3          & 66.5          \\
            \cmidrule{2-10}
                                             & \multirow{4}{*}{IGA}   & $\theta=\omega \oplus \phi$ & \xmark          & \xmark       & 51.7 $\pm$ 1.3          & 49.3 $\pm$ 1.5          & 58.6 $\pm$ 7.1          & 69.0 $\pm$ 1.1          & 57.1          \\
                                             &                        & $\omega$                    & \xmark          & \xmark       & 61.9 $\pm$ 0.0          & 53.6 $\pm$ 0.9          & 75.7 $\pm$ 0.5          & 76.0 $\pm$ 0.1          & 66.8          \\
                                             &                        & $\omega$                    & \cmark          & \xmark       & 61.2 $\pm$ 0.1          & 52.2 $\pm$ 0.5          & 76.1 $\pm$ 0.2          & 77.2 $\pm$ 0.3          & 66.7          \\
                                             &                        & $\omega$                    & \cmark          & \cmark       & 61.7 $\pm$ 0.5          & 52.4 $\pm$ 0.7          & 75.9 $\pm$ 0.4          & 77.1 $\pm$ 0.2          & 66.8          \\
            \cmidrule{2-10}
                                             & \multirow{3}{*}{Fishr} & \multirow{3}{*}{$\omega$}   & \xmark          & \xmark       & \textbf{63.8} $\pm$ 0.6 & 52.5 $\pm$ 0.5          & \textbf{76.7} $\pm$ 0.6 & 77.1 $\pm$ 1.0          & 67.5          \\
                                             &                        &                             & \cmark          & \xmark       & 62.2 $\pm$ 1.0          & 53.5 $\pm$ 0.2          & 76.6 $\pm$ 0.2          & 77.8 $\pm$ 0.4          & 67.5          \\
                                             &                        &                             & \cmark          & \cmark       & 62.4 $\pm$ 0.5          & \textbf{54.4} $\pm$ 0.4 & 76.2 $\pm$ 0.5          & \textbf{78.3} $\pm$ 0.1 & 67.8          \\
            \cmidrule{2-10}
                                             & Fishr + IGA            & $\omega$                    & \cmark          & \cmark       & 63.3 $\pm$ 1.0          & 54.1 $\pm$ 0.3          & 76.5 $\pm$ 0.4          & 78.2 $\pm$ 0.6          & \textbf{68.0} \\
            \bottomrule
        \end{tabular}
    }
    \label{table:igahome}
\end{table}

%% file: sections/appendix/07_appendix_domainbedv15.tex
\subsubsection{Hyperparameter distributions}
\input{tables/domainbed/db_v15.tex}
\label{appendix:v15}



This Section is a preliminary introduction to a meta-discussion, not about the methodology to select the best hyperparameters, but about the methodology to select the hyperparameter distributions in DomainBed. This question has not been discussed in previous works (as far as we know).

After few initial iterations on the main idea of the paper, we had to select the distributions to sample our three hyperparameters from, as described in Table \ref{table:hyperparameters}.
\textit{First}, to select the ema $\gamma$ distribution, we knew that the authors from \citet{le201115} have not noticed \enquote{any significant difference in validation errors} for different values higher than 0.9. Moreover $\gamma$ should remain strictly lower than 1. Thus, sampling from Uniform$(0.9, 0.99)$ seemed appropriate.
\textit{Second}, sampling the number of warmup iterations uniformly along training from Uniform$(0, 5000)$ seemed the most natural and neutral choice.
\textit{Lastly}, the choice of the $\lambda$ distribution was more complex. As a reminder, a low $\lambda$ inactivates the regularization while an extremely high $\lambda$ may destabilize the training.


In Table \ref{table:v15}, we investigate two distributions: $\lambda \sim 10^{\text{Uniform}(1,4)}$ (eventually chosen for Fishr) and $\lambda \sim 10^{\text{Uniform}(1,5)}$.
\textit{First}, we observe that results are mostly similar: it confirms that Fishr is consistently better than ERM (where $\lambda=0$), and in average is the best approach with the `Test-domain' model selection and among the best approaches with the `Training-domain' model selection.
\textit{Second}, the existence of consistent differences in results suggests that the best hyperparameter distribution depends on the dataset at hand and that the performance gap depends on the selection method.

While out of the scope of this paper, we believe these results were important for transparency (along with publishing our code), and may motivate the need for new protocols --- for example with bayesian hyperparameter search \citep{pmlr-v133-turner21a} --- that future benchmarks may introduce.

%% file: tables/domainbed/db_v15.tex
\begin{table}[h]%
    \caption{\textbf{Impact of the $\lambda$ distribution} from Table \ref{table:hyperparameters}.}%
    \centering
    \adjustbox{width=0.9\textwidth}{
        \begin{tabular}{llcccccccc}
            \toprule
            \textbf{Model selection}         & \textbf{$\lambda$ distribution} & \textbf{CMNIST}         & \textbf{RMNIST}         & \textbf{VLCS}           & \textbf{PACS}           & \textbf{OfficeHome}     & \textbf{TerraInc}       & \textbf{DomainNet}      & \textbf{Avg}  \\
            \midrule
            \multirow{3}{*}{Test-domain}     & Constant($0$) (= ERM)           & 57.8 $\pm$ 0.2          & 97.8 $\pm$ 0.1          & 77.6 $\pm$ 0.3          & 86.7 $\pm$ 0.3          & 66.4 $\pm$ 0.5          & 53.0 $\pm$ 0.3          & 41.3 $\pm$ 0.1          & 68.7          \\
                                             & $10^{\text {Uniform }(1, 4)}$   & \textbf{68.8} $\pm$ 1.4 & 97.8 $\pm$ 0.1          & 78.2 $\pm$ 0.2          & 86.9 $\pm$ 0.2          & \textbf{68.2} $\pm$ 0.2 & \textbf{53.6} $\pm$ 0.4 & 41.8 $\pm$ 0.1          & \textbf{70.8} \\
                                             & $10^{\text {Uniform }(1, 5)}$   & 68.7 $\pm$ 1.3          & 97.8 $\pm$ 0.0          & \textbf{78.7} $\pm$ 0.3 & \textbf{87.5} $\pm$ 0.1 & 68.0 $\pm$ 0.4          & 52.2 $\pm$ 0.5          & \textbf{42.0} $\pm$ 0.1 & 70.7          \\
            \midrule
            \multirow{3}{*}{Training-domain} & Constant($0$) (= ERM)           & 51.5 $\pm$ 0.1          & \textbf{98.0} $\pm$ 0.0 & 77.5 $\pm$ 0.4          & 85.5 $\pm$ 0.2          & 66.5 $\pm$ 0.3          & 46.1 $\pm$ 1.8          & 40.9 $\pm$ 0.1          & 66.6          \\
                                             & $10^{\text {Uniform }(1, 4)}$   & \textbf{52.0} $\pm$ 0.2 & 97.8 $\pm$ 0.0          & 77.8 $\pm$ 0.1          & 85.5 $\pm$ 0.4          & \textbf{67.8} $\pm$ 0.1 & \textbf{47.4} $\pm$ 1.6 & 41.7 $\pm$ 0.0          & \textbf{67.1} \\
                                             & $10^{\text {Uniform }(1, 5)}$   & 51.8 $\pm$ 0.3          & 97.9 $\pm$ 0.0          & \textbf{77.9} $\pm$ 0.1 & 85.5 $\pm$ 0.6          & 67.4 $\pm$ 0.3          & 47.2 $\pm$ 1.0          & \textbf{41.8} $\pm$ 0.1 & \textbf{67.1} \\
            \bottomrule
        \end{tabular}
    }
    \label{table:v15}
\end{table}


%% file: sections/appendix/07_appendix_domainbed.tex
\subsection{Full DomainBed results}
\label{appendix:domainbedperdataset}




Tables below detail results for each dataset with 'Test-domain' and 'Training-domain' model selection methods.
We format \textbf{first} and \underline{second} best accuracies.
Note that the per-dataset results for Fish \citep{shi2021gradient} are not available.

\clearpage
\subsubsection{Colored MNIST}
\label{appendix:db_cmnist}%

\input{tables/oracle/db_cmnist_oracle.tex}
\input{tables/training/db_cmnist_training.tex}

\clearpage
\subsubsection{Rotated MNIST}

\input{tables/oracle/db_rmnist_oracle.tex}

\input{tables/training/db_rmnist_training.tex}
\clearpage
\subsubsection{VLCS}
\input{tables/oracle/db_vlcs_oracle}
\input{tables/training/db_vlcs_training.tex}
\clearpage
\subsubsection{PACS}
\input{tables/oracle/db_pacs_oracle}
\input{tables/training/db_pacs_training.tex}
\clearpage
\subsubsection{OfficeHome}

\input{tables/oracle/db_officehome_oracle.tex}
\input{tables/training/db_officehome_training.tex}
\clearpage
\subsubsection{TerraIncognita}
\input{tables/oracle/db_terra_oracle.tex}
\input{tables/training/db_terra_training.tex}
\clearpage
\subsubsection{DomainNet}
\input{tables/oracle/db_net_oracle.tex}
\input{tables/training/db_net_training.tex}

\newpage

%% file: tables/oracle/db_cmnist_oracle.tex
\begin{center}
    \adjustbox{max width=\textwidth}{%
\begin{tabular}{lccccc}
            \toprule
            \multicolumn{6}{l}{\textbf{Colored MNIST. Model selection: `Test-domain' validation set}}                                                \\
            \midrule
            \textbf{Algorithm} & \textbf{+90\%}             & \textbf{+80\%}             & \textbf{10\%}              & \textbf{Avg}     & \textbf{Ranking} \\
            \midrule
            ERM                & 71.8 $\pm$ 0.4             & 72.9 $\pm$ 0.1             & 28.7 $\pm$ 0.5             & 57.8             & 16               \\
            IRM                & 72.0 $\pm$ 0.1             & 72.5 $\pm$ 0.3             & \underline{58.5} $\pm$ 3.3 & \underline{67.7} & 2                \\
            GroupDRO           & 73.5 $\pm$ 0.3             & 73.0 $\pm$ 0.3             & 36.8 $\pm$ 2.8             & 61.1             & 8                \\
            Mixup              & 72.5 $\pm$ 0.2             & 73.9 $\pm$ 0.4             & 28.6 $\pm$ 0.2             & 58.4             & 13               \\
            MLDG               & 71.9 $\pm$ 0.3             & 73.5 $\pm$ 0.2             & 29.1 $\pm$ 0.9             & 58.2             & 14               \\
            CORAL              & 71.1 $\pm$ 0.2             & 73.4 $\pm$ 0.2             & 31.1 $\pm$ 1.6             & 58.6             & 10               \\
\bottomrule
        \end{tabular}
    }
\end{center}

%% file: tables/training/db_cmnist_training.tex
\begin{center}
    \adjustbox{max width=\textwidth}{%
        \begin{tabular}{lccccc}
            \toprule
            \multicolumn{6}{l}{\textbf{Colored MNIST. Model selection: `Training-domain' validation set}}                                                      \\
            \midrule
            \textbf{Algorithm} & \textbf{+90\%}             & \textbf{+80\%}             & \textbf{10\%}              & \textbf{Avg}     & \textbf{Ranking} \\
            \midrule
            ERM                & 71.7 $\pm$ 0.1             & 72.9 $\pm$ 0.2             & 10.0 $\pm$ 0.1             & 51.5             & 12               \\
            IRM                & 72.5 $\pm$ 0.1             & 73.3 $\pm$ 0.5             & 10.2 $\pm$ 0.3             & 52.0             & 4                \\
            GroupDRO           & \underline{73.1} $\pm$ 0.3 & 73.2 $\pm$ 0.2             & 10.0 $\pm$ 0.2             & \underline{52.1} & 2                \\
            Mixup              & 72.7 $\pm$ 0.4             & 73.4 $\pm$ 0.1             & 10.1 $\pm$ 0.1             & \underline{52.1} & 2                \\
            MLDG               & 71.5 $\pm$ 0.2             & 73.1 $\pm$ 0.2             & 9.8 $\pm$ 0.1              & 51.5             & 12               \\
            CORAL              & 71.6 $\pm$ 0.3             & 73.1 $\pm$ 0.1             & 9.9 $\pm$ 0.1              & 51.5             & 12               \\
\bottomrule
        \end{tabular}
    }
\end{center}

%% file: tables/oracle/db_rmnist_oracle.tex
\begin{center}
    \adjustbox{max width=\textwidth}{%
        \begin{tabular}{lcccccccc}
            \toprule
            \multicolumn{9}{l}{\textbf{Rotated MNIST. Model selection: `Test-domain' validation set}}                                                                                                                               \\
            \midrule
            \textbf{Algorithm} & \textbf{0}                 & \textbf{15}             & \textbf{30}             & \textbf{45}             & \textbf{60}                & \textbf{75}                & \textbf{Avg}     & \textbf{Ranking} \\
            \midrule
            ERM                & 95.3 $\pm$ 0.2             & 98.7 $\pm$ 0.1          & 98.9 $\pm$ 0.1          & 98.7 $\pm$ 0.2          & 98.9 $\pm$ 0.0             & 96.2 $\pm$ 0.2             & 97.8             & 12               \\
            IRM                & 94.9 $\pm$ 0.6             & 98.7 $\pm$ 0.2          & 98.6 $\pm$ 0.1          & 98.6 $\pm$ 0.2          & 98.7 $\pm$ 0.1             & 95.2 $\pm$ 0.3             & 97.5             & 16               \\
            GroupDRO           & 95.9 $\pm$ 0.1             & \textbf{99.0} $\pm$ 0.1 & 98.9 $\pm$ 0.1          & 98.8 $\pm$ 0.1          & 98.6 $\pm$ 0.1             & 96.3 $\pm$ 0.4             & 97.9             & 5                \\
            Mixup              & 95.8 $\pm$ 0.3             & 98.7 $\pm$ 0.0          & \textbf{99.0} $\pm$ 0.1 & 98.8 $\pm$ 0.1          & 98.8 $\pm$ 0.1             & \underline{96.6} $\pm$ 0.2 & \underline{98.0} & 2                \\
            MLDG               & 95.7 $\pm$ 0.2             & 98.9 $\pm$ 0.1          & 98.8 $\pm$ 0.1          & \textbf{98.9} $\pm$ 0.1 & 98.6 $\pm$ 0.1             & 95.8 $\pm$ 0.4             & 97.8             & 12               \\
            CORAL              & \textbf{96.2} $\pm$ 0.2    & 98.8 $\pm$ 0.1          & 98.8 $\pm$ 0.1          & 98.8 $\pm$ 0.1          & 98.9 $\pm$ 0.1             & 96.4 $\pm$ 0.2             & \underline{98.0} & 2                \\
\bottomrule
        \end{tabular}
    }
    \label{table:db_rmnist_oracle}%
\end{center}

%% file: tables/training/db_rmnist_training.tex
\begin{center}
    \adjustbox{max width=\textwidth}{%
        \begin{tabular}{lcccccccc}
            \toprule
            \multicolumn{9}{l}{\textbf{Rotated MNIST. Model selection: `Training-domain' validation set}}                                                                                                                                           \\
            \midrule
            \textbf{Algorithm} & \textbf{0}                 & \textbf{15}                & \textbf{30}                & \textbf{45}                & \textbf{60}                & \textbf{75}                & \textbf{Avg}     & \textbf{Ranking} \\
            \midrule
            ERM                & \underline{95.9} $\pm$ 0.1 & 98.9 $\pm$ 0.0             & 98.8 $\pm$ 0.0             & 98.9 $\pm$ 0.0             & 98.9 $\pm$ 0.0             & 96.4 $\pm$ 0.0             & \underline{98.0} & 2                \\
            IRM                & 95.5 $\pm$ 0.1             & 98.8 $\pm$ 0.2             & 98.7 $\pm$ 0.1             & 98.6 $\pm$ 0.1             & 98.7 $\pm$ 0.0             & 95.9 $\pm$ 0.2             & 97.7             & 15               \\
            GroupDRO           & 95.6 $\pm$ 0.1             & 98.9 $\pm$ 0.1             & 98.9 $\pm$ 0.1             & \underline{99.0} $\pm$ 0.0 & 98.9 $\pm$ 0.0             & \textbf{96.5} $\pm$ 0.2    & \underline{98.0} & 2                \\
            Mixup              & 95.8 $\pm$ 0.3             & 98.9 $\pm$ 0.0             & 98.9 $\pm$ 0.0             & 98.9 $\pm$ 0.0             & 98.8 $\pm$ 0.1             & \textbf{96.5} $\pm$ 0.3    & \underline{98.0} & 2                \\
            MLDG               & 95.8 $\pm$ 0.1             & 98.9 $\pm$ 0.1             & \underline{99.0} $\pm$ 0.0 & 98.9 $\pm$ 0.1             & \underline{99.0} $\pm$ 0.0 & 95.8 $\pm$ 0.3             & 97.9             & 8                \\
\bottomrule
        \end{tabular}}
\end{center}

%% file: tables/oracle/db_vlcs_oracle.tex
\begin{center}
    \adjustbox{max width=\textwidth}{%
        \begin{tabular}{l|cccccc}
            \toprule
            \multicolumn{7}{l}{\textbf{VLCS. Model selection: `Test-domain' validation set}}                                                                                             \\
            \midrule
            \textbf{Algorithm} & \textbf{C}                 & \textbf{L}                 & \textbf{S}                 & \textbf{V}                 & \textbf{Avg}     & \textbf{Ranking} \\
            \midrule
            ERM                & 97.6 $\pm$ 0.3             & \underline{67.9} $\pm$ 0.7 & 70.9 $\pm$ 0.2             & 74.0 $\pm$ 0.6             & 77.6             & 12               \\
            IRM                & 97.3 $\pm$ 0.2             & 66.7 $\pm$ 0.1             & 71.0 $\pm$ 2.3             & 72.8 $\pm$ 0.4             & 76.9             & 16               \\
            GroupDRO           & 97.7 $\pm$ 0.2             & 65.9 $\pm$ 0.2             & 72.8 $\pm$ 0.8             & 73.4 $\pm$ 1.3             & 77.4             & 15               \\
            Mixup              & 97.8 $\pm$ 0.4             & 67.2 $\pm$ 0.4             & 71.5 $\pm$ 0.2             & 75.7 $\pm$ 0.6             & 78.1             & 4                \\
            MLDG               & 97.1 $\pm$ 0.5             & 66.6 $\pm$ 0.5             & 71.5 $\pm$ 0.1             & 75.0 $\pm$ 0.9             & 77.5             & 14               \\
            CORAL              & 97.3 $\pm$ 0.2             & 67.5 $\pm$ 0.6             & 71.6 $\pm$ 0.6             & 74.5 $\pm$ 0.0             & 77.7             & 10               \\
\bottomrule
        \end{tabular}}
    \label{table:db_vlcs_oracle}%
\end{center}

%% file: tables/training/db_vlcs_training.tex
\begin{center}
    \adjustbox{max width=\textwidth}{%
        \begin{tabular}{lcccccc}
            \toprule
            \multicolumn{7}{l}{\textbf{VLCS. Model selection: `Training-domain' validation set}}                                                                                       \\
            \midrule
            \textbf{Algorithm} & \textbf{C}                 & \textbf{L}                 & \textbf{S}                 & \textbf{V}              & \textbf{Avg}     & \textbf{Ranking} \\
            \midrule
            ERM                & 97.7 $\pm$ 0.4             & 64.3 $\pm$ 0.9             & 73.4 $\pm$ 0.5             & 74.6 $\pm$ 1.3          & 77.5             & 10               \\
            IRM                & 98.6 $\pm$ 0.1             & 64.9 $\pm$ 0.9             & 73.4 $\pm$ 0.6             & 77.3 $\pm$ 0.9          & \underline{78.5} & 3                \\
            GroupDRO           & 97.3 $\pm$ 0.3             & 63.4 $\pm$ 0.9             & 69.5 $\pm$ 0.8             & 76.7 $\pm$ 0.7          & 76.7             & 18               \\
            Mixup              & 98.3 $\pm$ 0.6             & 64.8 $\pm$ 1.0             & 72.1 $\pm$ 0.5             & 74.3 $\pm$ 0.8          & 77.4             & 13               \\
            MLDG               & 97.4 $\pm$ 0.2             & \textbf{65.2} $\pm$ 0.7    & 71.0 $\pm$ 1.4             & 75.3 $\pm$ 1.0          & 77.2             & 15               \\
            CORAL              & 98.3 $\pm$ 0.1             & 66.1 $\pm$ 1.2             & 73.4 $\pm$ 0.3             & \textbf{77.5} $\pm$ 1.2 & \textbf{78.8}    & 1                \\
\bottomrule
        \end{tabular}}
\end{center}

%% file: tables/oracle/db_pacs_oracle.tex
\begin{center}
    \adjustbox{max width=\textwidth}{%
        \begin{tabular}{l|cccccc}
            \toprule
            \multicolumn{7}{l}{\textbf{PACS. Model selection: `Test-domain' validation set}}                                                                                 \\
            \midrule
            \textbf{Algorithm} & \textbf{A}                 & \textbf{C}                 & \textbf{P}                 & \textbf{S}              & \textbf{Avg}  & \textbf{Ranking} \\
            \midrule
            ERM                & 86.5 $\pm$ 1.0             & 81.3 $\pm$ 0.6             & 96.2 $\pm$ 0.3             & \textbf{82.7} $\pm$ 1.1 & 86.7          & 8                \\
            IRM                & 84.2 $\pm$ 0.9             & 79.7 $\pm$ 1.5             & 95.9 $\pm$ 0.4             & 78.3 $\pm$ 2.1          & 84.5          & 18               \\
            GroupDRO           & 87.5 $\pm$ 0.5             & \textbf{82.9} $\pm$ 0.6    & 97.1 $\pm$ 0.3             & 81.1 $\pm$ 1.2          & 87.1          & 3                \\
            Mixup              & 87.5 $\pm$ 0.4             & 81.6 $\pm$ 0.7             & \underline{97.4} $\pm$ 0.2 & 80.8 $\pm$ 0.9          & 86.8          & 6                \\
            MLDG               & 87.0 $\pm$ 1.2             & 82.5 $\pm$ 0.9             & 96.7 $\pm$ 0.3             & 81.2 $\pm$ 0.6          & 86.8          & 6                \\
            CORAL              & 86.6 $\pm$ 0.8             & 81.8 $\pm$ 0.9             & 97.1 $\pm$ 0.5             & \textbf{82.7} $\pm$ 0.6 & 87.1          & 3                \\
\bottomrule
        \end{tabular}}
    \label{table:db_pacs_oracle}%
\end{center}

%% file: tables/training/db_pacs_training.tex
\begin{center}
    \adjustbox{max width=\textwidth}{%
        \begin{tabular}{lcccccc}
            \toprule
            \multicolumn{7}{l}{\textbf{PACS. Model selection: `Training-domain' validation set}}                                                                                        \\
            \midrule
            \textbf{Algorithm} & \textbf{A}                 & \textbf{C}                 & \textbf{P}                 & \textbf{S}                 & \textbf{Avg}     & \textbf{Ranking} \\
            \midrule
            ERM                & 84.7 $\pm$ 0.4             & \textbf{80.8} $\pm$ 0.6    & 97.2 $\pm$ 0.3             & \underline{79.3} $\pm$ 1.0 & 85.5             & 3                \\
            IRM                & 84.8 $\pm$ 1.3             & 76.4 $\pm$ 1.1             & 96.7 $\pm$ 0.6             & 76.1 $\pm$ 1.0             & 83.5             & 17               \\
            GroupDRO           & 83.5 $\pm$ 0.9             & 79.1 $\pm$ 0.6             & 96.7 $\pm$ 0.3             & 78.3 $\pm$ 2.0             & 84.4             & 14               \\
            Mixup              & 86.1 $\pm$ 0.5             & 78.9 $\pm$ 0.8             & 97.6 $\pm$ 0.1             & 75.8 $\pm$ 1.8             & 84.6             & 10               \\
            MLDG               & 85.5 $\pm$ 1.4             & 80.1 $\pm$ 1.7             & 97.4 $\pm$ 0.3             & 76.6 $\pm$ 1.1             & 84.9             & 8                \\
            CORAL              & 88.3 $\pm$ 0.2             & 80.0 $\pm$ 0.5             & \underline{97.5} $\pm$ 0.3 & 78.8 $\pm$ 1.3             & \underline{86.2} & 2                \\
\bottomrule
        \end{tabular}}
\end{center}

%% file: tables/oracle/db_officehome_oracle.tex
\begin{center}
    \adjustbox{max width=\textwidth}{%
        \begin{tabular}{l|cccccc}
            \toprule
            \multicolumn{7}{l}{\textbf{OfficeHome. Model selection: `Test-domain' validation set}}                                                                                       \\
            \midrule
            \textbf{Algorithm} & \textbf{A}                 & \textbf{C}                 & \textbf{P}                 & \textbf{R}                 & \textbf{Avg}     & \textbf{Ranking} \\
            \midrule
            ERM                & 61.7 $\pm$ 0.7             & 53.4 $\pm$ 0.3             & 74.1 $\pm$ 0.4             & 76.2 $\pm$ 0.6             & 66.4             & 8                \\
            IRM                & 56.4 $\pm$ 3.2             & 51.2 $\pm$ 2.3             & 71.7 $\pm$ 2.7             & 72.7 $\pm$ 2.7             & 63.0             & 18               \\
            GroupDRO           & 60.5 $\pm$ 1.6             & 53.1 $\pm$ 0.3             & 75.5 $\pm$ 0.3             & 75.9 $\pm$ 0.7             & 66.2             & 3                \\
            Mixup              & \underline{63.5} $\pm$ 0.2 & \textbf{54.6} $\pm$ 0.4    & 76.0 $\pm$ 0.3             & 78.0 $\pm$ 0.7             & 68.0             & 6                \\
            MLDG               & 60.5 $\pm$ 0.7             & \underline{54.2} $\pm$ 0.5 & 75.0 $\pm$ 0.2             & 76.7 $\pm$ 0.5             & 66.6             & 6                \\
            CORAL              & \textbf{64.8} $\pm$ 0.8    & 54.1 $\pm$ 0.9             & \textbf{76.5} $\pm$ 0.4    & \underline{78.2} $\pm$ 0.4 & \textbf{68.4}    & 3                \\
\bottomrule
        \end{tabular}}
    \label{table:db_officehome_oracle}%
\end{center}

%% file: tables/training/db_officehome_training.tex
\begin{center}
    \adjustbox{max width=\textwidth}{%
        \begin{tabular}{lcccccc}
            \toprule
            \multicolumn{7}{l}{\textbf{OfficeHome. Model selection: `Training-domain' validation set}}                                                                                \\
            \midrule
            \textbf{Algorithm} & \textbf{A}                 & \textbf{C}              & \textbf{P}                 & \textbf{R}                 & \textbf{Avg}     & \textbf{Ranking} \\
            \midrule
            ERM                & 61.3 $\pm$ 0.7             & 52.4 $\pm$ 0.3          & 75.8 $\pm$ 0.1             & 76.6 $\pm$ 0.3             & 66.5             & 7                \\
            IRM                & 58.9 $\pm$ 2.3             & 52.2 $\pm$ 1.6          & 72.1 $\pm$ 2.9             & 74.0 $\pm$ 2.5             & 64.3             & 18               \\
            GroupDRO           & 60.4 $\pm$ 0.7             & 52.7 $\pm$ 1.0          & 75.0 $\pm$ 0.7             & 76.0 $\pm$ 0.7             & 66.0             & 11               \\
            Mixup              & 62.4 $\pm$ 0.8             & \textbf{54.8} $\pm$ 0.6 & \textbf{76.9} $\pm$ 0.3    & \underline{78.3} $\pm$ 0.2 & 68.1             & 3                \\
            MLDG               & 61.5 $\pm$ 0.9             & 53.2 $\pm$ 0.6          & 75.0 $\pm$ 1.2             & 77.5 $\pm$ 0.4             & 66.8             & 6                \\
            CORAL              & \textbf{65.3} $\pm$ 0.4    & 54.4 $\pm$ 0.5          & \underline{76.5} $\pm$ 0.1 & \textbf{78.4} $\pm$ 0.5    & \textbf{68.7}    & 1                \\
\bottomrule
        \end{tabular}}
\end{center}

%% file: tables/oracle/db_terra_oracle.tex
\begin{center}
    \adjustbox{max width=\textwidth}{%
        \begin{tabular}{lcccccc}
            \toprule
            \multicolumn{7}{l}{\textbf{TerraIncognita. Model selection: `Test-domain' validation set}}                                                                          \\
            \midrule
            Algorithm & \textbf{L100}              & \textbf{L38}               & \textbf{L43}               & \textbf{L46}               & \textbf{Avg}     & \textbf{Ranking} \\
            \midrule
            ERM       & 59.4 $\pm$ 0.9             & 49.3 $\pm$ 0.6             & \textbf{60.1} $\pm$ 1.1    & 43.2 $\pm$ 0.5             & 53.0             & 3                \\
            IRM       & 56.5 $\pm$ 2.5             & 49.8 $\pm$ 1.5             & 57.1 $\pm$ 2.2             & 38.6 $\pm$ 1.0             & 50.5             & 16               \\
            GroupDRO  & \underline{60.4} $\pm$ 1.5 & 48.3 $\pm$ 0.4             & 58.6 $\pm$ 0.8             & 42.2 $\pm$ 0.8             & 52.4             & 6                \\
            Mixup     & 67.6 $\pm$ 1.8             & \textbf{51.0} $\pm$ 1.3    & 59.0 $\pm$ 0.0             & 40.0 $\pm$ 1.1             & \textbf{54.4}    & 1                \\
            MLDG      & 59.2 $\pm$ 0.1             & 49.0 $\pm$ 0.9             & 58.4 $\pm$ 0.9             & 41.4 $\pm$ 1.0             & 52.0             & 9                \\
            CORAL     & \underline{60.4} $\pm$ 0.9 & 47.2 $\pm$ 0.5             & 59.3 $\pm$ 0.4             & \underline{44.4} $\pm$ 0.4 & 52.8             & 4                \\
\bottomrule
        \end{tabular}
    }
    \label{table:db_terra_oracle}%
\end{center}

%% file: tables/training/db_terra_training.tex
\begin{center}
    \adjustbox{max width=\textwidth}{%
        \begin{tabular}{lcccccc}
            \toprule
            \multicolumn{7}{l}{\textbf{TerraIncognita. Model selection: `Training-domain' validation set}}                                                                     \\
            \midrule
            Algorithm & \textbf{L100}              & \textbf{L38}               & \textbf{L43}               & \textbf{L46}               & \textbf{Avg}     & \textbf{Ranking} \\
            \midrule
            ERM       & 49.8 $\pm$ 4.4             & 42.1 $\pm$ 1.4             & 56.9 $\pm$ 1.8             & 35.7 $\pm$ 3.9             & 46.1             & 10               \\
            IRM       & \underline{54.6} $\pm$ 1.3 & 39.8 $\pm$ 1.9             & 56.2 $\pm$ 1.8             & 39.6 $\pm$ 0.8             & 47.6             & 4                \\
            GroupDRO  & 41.2 $\pm$ 0.7             & 38.6 $\pm$ 2.1             & 56.7 $\pm$ 0.9             & 36.4 $\pm$ 2.1             & 43.2             & 16               \\
            Mixup     & \textbf{59.6} $\pm$ 2.0    & 42.2 $\pm$ 1.4             & 55.9 $\pm$ 0.8             & 33.9 $\pm$ 1.4             & \underline{47.9} & 2                \\
            MLDG      & 54.2 $\pm$ 3.0             & \textbf{44.3} $\pm$ 1.1    & 55.6 $\pm$ 0.3             & 36.9 $\pm$ 2.2             & 47.7             & 3                \\
            CORAL     & 51.6 $\pm$ 2.4             & 42.2 $\pm$ 1.0             & 57.0 $\pm$ 1.0             & \underline{39.8} $\pm$ 2.9 & 47.6             & 4                \\
\bottomrule
        \end{tabular}
    }
    \label{table:db_terra_training}%
\end{center}

%% file: tables/oracle/db_net_oracle.tex
\begin{center}
    \adjustbox{max width=\textwidth}{%
        \begin{tabular}{lcccccccc}
            \toprule
            \multicolumn{9}{l}{\textbf{DomainNet. Model selection: `Test-domain' validation set}}                                                                                                                                                  \\
            \midrule
            \textbf{Algorithm} & \textbf{clip}              & \textbf{info}              & \textbf{paint}             & \textbf{quick}             & \textbf{real}              & \textbf{sketch}            & \textbf{Avg}     & \textbf{Ranking} \\
            \midrule
            ERM                & 58.6 $\pm$ 0.3             & 19.2 $\pm$ 0.2             & 47.0 $\pm$ 0.3             & 13.2 $\pm$ 0.2             & 59.9 $\pm$ 0.3             & 49.8 $\pm$ 0.4             & 41.3             & 5                \\
            IRM                & 40.4 $\pm$ 6.6             & 12.1 $\pm$ 2.7             & 31.4 $\pm$ 5.7             & 9.8 $\pm$ 1.2              & 37.7 $\pm$ 9.0             & 36.7 $\pm$ 5.3             & 28.0             & 17               \\
            GroupDRO           & 47.2 $\pm$ 0.5             & 17.5 $\pm$ 0.4             & 34.2 $\pm$ 0.3             & 9.2 $\pm$ 0.4              & 51.9 $\pm$ 0.5             & 40.1 $\pm$ 0.6             & 33.4             & 14               \\
            Mixup              & 55.6 $\pm$ 0.1             & 18.7 $\pm$ 0.4             & 45.1 $\pm$ 0.5             & 12.8 $\pm$ 0.3             & 57.6 $\pm$ 0.5             & 48.2 $\pm$ 0.4             & 39.6             & 8                \\
            MLDG               & \textbf{59.3} $\pm$ 0.1    & 19.6 $\pm$ 0.2             & 46.8 $\pm$ 0.2             & 13.4 $\pm$ 0.2             & \underline{60.1} $\pm$ 0.4 & \underline{50.4} $\pm$ 0.3 & 41.6             & 4                \\
\bottomrule
        \end{tabular}}
\end{center}

%% file: tables/training/db_net_training.tex
\begin{center}
    \adjustbox{max width=\textwidth}{%
        \begin{tabular}{lcccccccc}
            \toprule
            \multicolumn{9}{l}{\textbf{DomainNet. Model selection: `Training-domain' validation set}}                                                                                                                                           \\
            \midrule
            \textbf{Algorithm} & \textbf{clip}              & \textbf{info}              & \textbf{paint}             & \textbf{quick}          & \textbf{real}              & \textbf{sketch}            & \textbf{Avg}     & \textbf{Ranking} \\
            \midrule
            ERM                & 58.1 $\pm$ 0.3             & 18.8 $\pm$ 0.3             & \underline{46.7} $\pm$ 0.3 & 12.2 $\pm$ 0.4          & 59.6 $\pm$ 0.1             & 49.8 $\pm$ 0.4             & 40.9             & 5                \\
            IRM                & 48.5 $\pm$ 2.8             & 15.0 $\pm$ 1.5             & 38.3 $\pm$ 4.3             & 10.9 $\pm$ 0.5          & 48.2 $\pm$ 5.2             & 42.3 $\pm$ 3.1             & 33.9             & 14               \\
            GroupDRO           & 47.2 $\pm$ 0.5             & 17.5 $\pm$ 0.4             & 33.8 $\pm$ 0.5             & 9.3 $\pm$ 0.3           & 51.6 $\pm$ 0.4             & 40.1 $\pm$ 0.6             & 33.3             & 16               \\
            Mixup              & 55.7 $\pm$ 0.3             & 18.5 $\pm$ 0.5             & 44.3 $\pm$ 0.5             & 12.5 $\pm$ 0.4          & 55.8 $\pm$ 0.3             & 48.2 $\pm$ 0.5             & 39.2             & 8                \\
            MLDG               & \underline{59.1} $\pm$ 0.2 & 19.1 $\pm$ 0.3             & 45.8 $\pm$ 0.7             & \textbf{13.4} $\pm$ 0.3 & 59.6 $\pm$ 0.2             & \underline{50.2} $\pm$ 0.4 & 41.2             & 4                \\
\bottomrule
        \end{tabular}}
\end{center}

%% file: main.bbl
\begin{thebibliography}{118}
\providecommand{\natexlab}[1]{#1}
\providecommand{\url}[1]{\texttt{#1}}
\expandafter\ifx\csname urlstyle\endcsname\relax
  \providecommand{\doi}[1]{doi: #1}\else
  \providecommand{\doi}{doi: \begingroup \urlstyle{rm}\Url}\fi

\bibitem[Ahmed et~al.(2021)Ahmed, Bengio, van Seijen, and
  Courville]{ahmed2021systematic}
Ahmed, F., Bengio, Y., van Seijen, H., and Courville, A.
\newblock Systematic generalisation with group invariant predictions.
\newblock In \emph{ICLR}, 2021.

\bibitem[Ahuja et~al.(2019)Ahuja, Caballero, Zhang, Bengio, Mitliagkas, and
  Rish]{ahuja2021invariance}
Ahuja, K., Caballero, E., Zhang, D., Bengio, Y., Mitliagkas, I., and Rish, I.
\newblock Invariance principle meets information bottleneck for
  out-of-distribution generalization.
\newblock In \emph{NeurIPS}, 2019.

\bibitem[Arjovsky et~al.(2019)Arjovsky, Bottou, Gulrajani, and
  Lopez-Paz]{arjovsky2019invariant}
Arjovsky, M., Bottou, L., Gulrajani, I., and Lopez-Paz, D.
\newblock Invariant risk minimization.
\newblock \emph{arXiv preprint}, 2019.

\bibitem[Becker \& Le~Cun(1988)Becker and Le~Cun]{becker1988improving}
Becker, S. and Le~Cun, Y.
\newblock Improving the convergence of back-propagation learning with second
  order methods.
\newblock In \emph{Connectionist models summer school}, 1988.

\bibitem[Beery et~al.(2018)Beery, Van~Horn, and Perona]{beery2018recognition}
Beery, S., Van~Horn, G., and Perona, P.
\newblock Recognition in terra incognita.
\newblock In \emph{ECCV}, 2018.

\bibitem[Blanchard et~al.(2011)Blanchard, Lee, and
  Scott]{blanchard2011generalizing}
Blanchard, G., Lee, G., and Scott, C.
\newblock Generalizing from several related classification tasks to a new
  unlabeled sample.
\newblock In \emph{NeurIPS}, 2011.

\bibitem[Blanchard et~al.(2021)Blanchard, Deshmukh, Dogan, Lee, and
  Scott]{JMLR:v22:17-679}
Blanchard, G., Deshmukh, A.~A., Dogan, U., Lee, G., and Scott, C.
\newblock Domain generalization by marginal transfer learning.
\newblock \emph{JMLR}, 2021.

\bibitem[Cha et~al.(2021)Cha, Chun, Lee, Cho, Park, Lee, and Park]{cha2021wad}
Cha, J., Chun, S., Lee, K., Cho, H.-C., Park, S., Lee, Y., and Park, S.
\newblock {SWAD}: Domain generalization by seeking flat minima.
\newblock In \emph{NeurIPS}, 2021.

\bibitem[Chang et~al.(2020)Chang, Zhang, Yu, and Jaakkola]{chang2020invariant}
Chang, S., Zhang, Y., Yu, M., and Jaakkola, T.
\newblock Invariant rationalization.
\newblock In \emph{ICML}, 2020.

\bibitem[Charpiat et~al.(2019)Charpiat, Girard, Felardos, and
  Tarabalka]{charpiat2019input}
Charpiat, G., Girard, N., Felardos, L., and Tarabalka, Y.
\newblock Input similarity from the neural network perspective.
\newblock In \emph{NeurIPS}, 2019.

\bibitem[C.R.(1945)]{raofim1945}
C.R., R.
\newblock Information and accuracy attainable in the estimation of statistical
  parameters.
\newblock In \emph{Bulletin of the Calcutta Mathematical Society}, 1945.

\bibitem[Creager et~al.(2021)Creager, Jacobsen, and
  Zemel]{creager2020nvironment}
Creager, E., Jacobsen, J.-H., and Zemel, R.
\newblock Environment inference for invariant learning.
\newblock In \emph{ICML}, 2021.

\bibitem[D'Amour et~al.(2020)D'Amour, Heller, Moldovan, Adlam, Alipanahi,
  Beutel, Chen, Deaton, Eisenstein, Hoffman, et~al.]{d2020underspecification}
D'Amour, A., Heller, K., Moldovan, D., Adlam, B., Alipanahi, B., Beutel, A.,
  Chen, C., Deaton, J., Eisenstein, J., Hoffman, M.~D., et~al.
\newblock Underspecification presents challenges for credibility in modern
  machine learning.
\newblock \emph{JMLR}, 2020.

\bibitem[Dangel et~al.(2020)Dangel, Kunstner, and Hennig]{dangel2020backpack}
Dangel, F., Kunstner, F., and Hennig, P.
\newblock Back{PACK}: Packing more into backprop.
\newblock In \emph{ICLR}, 2020.

\bibitem[Dangel et~al.(2021)Dangel, Tatzel, and Hennig]{dangel2021vivit}
Dangel, F., Tatzel, L., and Hennig, P.
\newblock Vivit: Curvature access through the generalized gauss-newton's
  low-rank structure.
\newblock \emph{arXiv preprint}, 2021.

\bibitem[DeGrave et~al.(2021)DeGrave, Janizek, and Lee]{degrave2021ai}
DeGrave, A.~J., Janizek, J.~D., and Lee, S.-I.
\newblock Ai for radiographic covid-19 detection selects shortcuts over signal.
\newblock \emph{Nature Machine Intelligence}, 2021.

\bibitem[Deutsch(2011)]{hardtovary}
Deutsch, D.
\newblock The beginning of infinity: Explanations that transform the world.
\newblock \emph{Penguin UK}, 2011.

\bibitem[Ding \& Fu(2017)Ding and Fu]{ding2017deep}
Ding, Z. and Fu, Y.
\newblock Deep domain generalization with structured low-rank constraint.
\newblock In \emph{TIP}, 2017.

\bibitem[Dinh et~al.(2017)Dinh, Pascanu, Bengio, and Bengio]{pmlr-v70-dinh17b}
Dinh, L., Pascanu, R., Bengio, S., and Bengio, Y.
\newblock Sharp minima can generalize for deep nets.
\newblock In \emph{ICML}, 2017.

\bibitem[Du et~al.(2018)Du, Czarnecki, Jayakumar, Farajtabar, Pascanu, and
  Lakshminarayanan]{du2018adapting}
Du, Y., Czarnecki, W.~M., Jayakumar, S.~M., Farajtabar, M., Pascanu, R., and
  Lakshminarayanan, B.
\newblock Adapting auxiliary losses using gradient similarity.
\newblock \emph{arXiv preprint}, 2018.

\bibitem[Faghri et~al.(2020)Faghri, Duvenaud, Fleet, and Ba]{faghri2020study}
Faghri, F., Duvenaud, D., Fleet, D.~J., and Ba, J.
\newblock A study of gradient variance in deep learning.
\newblock \emph{arXiv preprint}, 2020.

\bibitem[Fang et~al.(2013)Fang, Xu, and Rockmore]{fang2013unbiased}
Fang, C., Xu, Y., and Rockmore, D.~N.
\newblock Unbiased metric learning: On the utilization of multiple datasets and
  web images for softening bias.
\newblock In \emph{ICCV}, 2013.

\bibitem[Finn et~al.(2017)Finn, Abbeel, and Levine]{finn2017odelagnostic}
Finn, C., Abbeel, P., and Levine, S.
\newblock Model-agnostic meta-learning for fast adaptation of deep networks.
\newblock In \emph{ICML}, 2017.

\bibitem[Fisher(1922)]{fisher1922mathematical}
Fisher, R.~A.
\newblock On the mathematical foundations of theoretical statistics.
\newblock \emph{Philosophical Transactions of the Royal Society of London.},
  1922.

\bibitem[Foret et~al.(2021)Foret, Kleiner, Mobahi, and
  Neyshabur]{foret2021sharpnessaware}
Foret, P., Kleiner, A., Mobahi, H., and Neyshabur, B.
\newblock Sharpness-aware minimization for efficiently improving
  generalization.
\newblock In \emph{ICLR}, 2021.

\bibitem[Fort et~al.(2019)Fort, Nowak, Jastrzebski, and
  Narayanan]{fort2019stiffness}
Fort, S., Nowak, P.~K., Jastrzebski, S., and Narayanan, S.
\newblock Stiffness: A new perspective on generalization in neural networks.
\newblock \emph{arXiv preprint}, 2019.

\bibitem[Frantar et~al.(2021)Frantar, Kurtic, and
  Alistarh]{frantar2021fficient}
Frantar, E., Kurtic, E., and Alistarh, D.
\newblock Efficient matrix-free approximations of second-order information,
  with applications to pruning and optimization.
\newblock \emph{arXiv preprint}, 2021.

\bibitem[Ganin et~al.(2016)Ganin, Ustinova, Ajakan, Germain, Larochelle,
  Laviolette, Marchand, and Lempitsky]{ganin2016domain}
Ganin, Y., Ustinova, E., Ajakan, H., Germain, P., Larochelle, H., Laviolette,
  F., Marchand, M., and Lempitsky, V.
\newblock Domain-adversarial training of neural networks.
\newblock \emph{JMLR}, 2016.

\bibitem[Geirhos et~al.(2020)Geirhos, Jacobsen, Michaelis, Zemel, Brendel,
  Bethge, and Wichmann]{geirhos2020shortcut}
Geirhos, R., Jacobsen, J.-H., Michaelis, C., Zemel, R., Brendel, W., Bethge,
  M., and Wichmann, F.~A.
\newblock Shortcut learning in deep neural networks.
\newblock \emph{Nature Machine Intelligence}, 2020.

\bibitem[Ghifary et~al.(2015)Ghifary, Kleijn, Zhang, and
  Balduzzi]{ghifary2015domain}
Ghifary, M., Kleijn, W.~B., Zhang, M., and Balduzzi, D.
\newblock Domain generalization for object recognition with multi-task
  autoencoders.
\newblock In \emph{ICCV}, 2015.

\bibitem[Ghorbani et~al.(2019)Ghorbani, Krishnan, and
  Xiao]{pmlr-v97-ghorbani19b}
Ghorbani, B., Krishnan, S., and Xiao, Y.
\newblock An investigation into neural net optimization via hessian eigenvalue
  density.
\newblock In \emph{ICML}, 2019.

\bibitem[Gong et~al.(2016)Gong, Zhang, Liu, Tao, Glymour, and
  Schölkopf]{pmlr-v48-gong16}
Gong, M., Zhang, K., Liu, T., Tao, D., Glymour, C., and Schölkopf, B.
\newblock Domain adaptation with conditional transferable components.
\newblock In \emph{ICML}, 2016.

\bibitem[Gulrajani \& Lopez-Paz(2021)Gulrajani and Lopez-Paz]{gulrajani2021in}
Gulrajani, I. and Lopez-Paz, D.
\newblock In search of lost domain generalization.
\newblock In \emph{ICLR}, 2021.

\bibitem[Guo et~al.(2021)Guo, Zhang, Liu, and Kiciman]{guo2021utofdistribution}
Guo, R., Zhang, P., Liu, H., and Kiciman, E.
\newblock Out-of-distribution prediction with invariant risk minimization: The
  limitation and an effective fix.
\newblock \emph{arXiv preprint}, 2021.

\bibitem[Gur-Ari et~al.(2018)Gur-Ari, Roberts, and Dyer]{gur2018gradient}
Gur-Ari, G., Roberts, D.~A., and Dyer, E.
\newblock Gradient descent happens in a tiny subspace.
\newblock \emph{arXiv preprint}, 2018.

\bibitem[Heskes(2000)]{heskes2000natural}
Heskes, T.
\newblock On “natural” learning and pruning in multilayered perceptrons.
\newblock \emph{Neural Computation}, 2000.

\bibitem[Huang et~al.(2020)Huang, Wang, Xing, and Huang]{huang2020self}
Huang, Z., Wang, H., Xing, E.~P., and Huang, D.
\newblock Self-challenging improves cross-domain generalization.
\newblock In \emph{ECCV}, 2020.

\bibitem[Idnani \& Kao(2020)Idnani and Kao]{idnanilearning}
Idnani, D. and Kao, J.~C.
\newblock Learning robust representations with score invariant learning.
\newblock In \emph{ICML UDL Workshop}, 2020.

\bibitem[Izmailov et~al.(2018)Izmailov, Podoprikhin, Garipov, Vetrov, and
  Wilson]{izmailov2018averaging}
Izmailov, P., Podoprikhin, D., Garipov, T., Vetrov, D., and Wilson, A.
\newblock Averaging weights leads to wider optima and better generalization.
\newblock In \emph{UAI}, 2018.

\bibitem[Jacot et~al.(2018)Jacot, Gabriel, and Hongler]{NEURIPS2018_5a4be1fa}
Jacot, A., Gabriel, F., and Hongler, C.
\newblock Neural tangent kernel: Convergence and generalization in neural
  networks.
\newblock In \emph{NeurIPS}, 2018.

\bibitem[Jastrzebski et~al.(2018)Jastrzebski, Kenton, Arpit, Ballas, Fischer,
  Storkey, and Bengio]{jastrzebski2018three}
Jastrzebski, S., Kenton, Z., Arpit, D., Ballas, N., Fischer, A., Storkey, A.,
  and Bengio, Y.
\newblock Three factors influencing minima in {SGD}.
\newblock In \emph{ICANN}, 2018.

\bibitem[Jastrzebski et~al.(2021)Jastrzebski, Arpit, Astrand, Kerg, Wang,
  Xiong, Socher, Cho, and Geras]{jastrzebski2021catastrophic}
Jastrzebski, S., Arpit, D., Astrand, O., Kerg, G.~B., Wang, H., Xiong, C.,
  Socher, R., Cho, K., and Geras, K.~J.
\newblock Catastrophic fisher explosion: Early phase fisher matrix impacts
  generalization.
\newblock In \emph{ICML}, 2021.

\bibitem[Johansson et~al.(2019)Johansson, Sontag, and
  Ranganath]{johansson2019support}
Johansson, F.~D., Sontag, D., and Ranganath, R.
\newblock Support and invertibility in domain-invariant representations.
\newblock In \emph{AISTATS}, 2019.

\bibitem[Kamath et~al.(2021)Kamath, Tangella, Sutherland, and
  Srebro]{pmlr-v130-kamath21a}
Kamath, P., Tangella, A., Sutherland, D., and Srebro, N.
\newblock Does invariant risk minimization capture invariance?
\newblock In \emph{AISTATS}, 2021.

\bibitem[Karakida et~al.(2019)Karakida, Akaho, and
  Amari]{karakida2019pathological}
Karakida, R., Akaho, S., and Amari, S.-i.
\newblock Pathological spectra of the fisher information metric and its
  variants in deep neural networks.
\newblock \emph{arXiv preprint}, 2019.

\bibitem[Kingma \& Ba(2014)Kingma and Ba]{kingma2014adam}
Kingma, D.~P. and Ba, J.
\newblock Adam: A method for stochastic optimization.
\newblock \emph{arXiv preprint}, 2014.

\bibitem[Kirkpatrick et~al.(2017)Kirkpatrick, Pascanu, Rabinowitz, Veness,
  Desjardins, Rusu, Milan, Quan, Ramalho, Grabska-Barwinska,
  et~al.]{kirkpatrick2017overcoming}
Kirkpatrick, J., Pascanu, R., Rabinowitz, N., Veness, J., Desjardins, G., Rusu,
  A.~A., Milan, K., Quan, J., Ramalho, T., Grabska-Barwinska, A., et~al.
\newblock Overcoming catastrophic forgetting in neural networks.
\newblock In \emph{PNAS}, 2017.

\bibitem[Koh et~al.(2020)Koh, Sagawa, Marklund, Xie, Zhang, Balsubramani, Hu,
  Yasunaga, Phillips, Gao, et~al.]{koh2020wilds}
Koh, P.~W., Sagawa, S., Marklund, H., Xie, S.~M., Zhang, M., Balsubramani, A.,
  Hu, W., Yasunaga, M., Phillips, R.~L., Gao, I., et~al.
\newblock Wilds: A benchmark of in-the-wild distribution shifts.
\newblock \emph{arXiv preprint}, 2020.

\bibitem[Kopitkov \& Indelman(2019)Kopitkov and Indelman]{kopitkov2019eural}
Kopitkov, D. and Indelman, V.
\newblock Neural spectrum alignment: Empirical study.
\newblock \emph{arXiv preprint}, 2019.

\bibitem[Koyama \& Yamaguchi(2020)Koyama and Yamaguchi]{koyama2020out}
Koyama, M. and Yamaguchi, S.
\newblock Out-of-distribution generalization with maximal invariant predictor.
\newblock \emph{arXiv preprint}, 2020.

\bibitem[Krizhevsky et~al.(2012)Krizhevsky, Sutskever, and
  Hinton]{krizhevsky2012imagenet}
Krizhevsky, A., Sutskever, I., and Hinton, G.~E.
\newblock Imagenet classification with deep convolutional neural networks.
\newblock In \emph{NeurIPS}, 2012.

\bibitem[Krueger et~al.(2021)Krueger, Caballero, Jacobsen, Zhang, Binas, Zhang,
  Priol, and Courville]{krueger2020utofdistribution}
Krueger, D., Caballero, E., Jacobsen, J.-H., Zhang, A., Binas, J., Zhang, D.,
  Priol, R.~L., and Courville, A.
\newblock Out-of-distribution generalization via risk extrapolation (rex).
\newblock In \emph{ICML}, 2021.

\bibitem[Kunstner et~al.(2019)Kunstner, Hennig, and
  Balles]{NEURIPS2019_46a558d9}
Kunstner, F., Hennig, P., and Balles, L.
\newblock Limitations of the empirical fisher approximation for natural
  gradient descent.
\newblock In \emph{NeurIPS}, 2019.

\bibitem[Le~Roux et~al.(2011)Le~Roux, Bengio, and Fitzgibbon]{le201115}
Le~Roux, N., Bengio, Y., and Fitzgibbon, A.
\newblock Improving first and second-order methods by modeling uncertainty.
\newblock \emph{Optimization for Machine Learning}, 2011.

\bibitem[LeCun et~al.(1990)LeCun, Denker, Solla, Howard, and
  Jackel]{e9ee2143e19d49cf9cbe8861950b6b2a}
LeCun, Y., Denker, J., Solla, S., Howard, R., and Jackel, L.
\newblock Optimal brain damage.
\newblock In \emph{NeurIPS}, 1990.

\bibitem[LeCun et~al.(2010)LeCun, Cortes, and Burges]{lecun2010mnist}
LeCun, Y., Cortes, C., and Burges, C.
\newblock Mnist handwritten digit database, 2010.

\bibitem[LeCun et~al.(2012)LeCun, Bottou, Orr, and
  Müller]{series/lncs/LeCunBOM12}
LeCun, Y., Bottou, L., Orr, G.~B., and Müller, K.-R.
\newblock Efficient backprop.
\newblock In \emph{Neural Networks}. 2012.

\bibitem[Li et~al.(2017)Li, Yang, Song, and Hospedales]{li2017deeper}
Li, D., Yang, Y., Song, Y.-Z., and Hospedales, T.~M.
\newblock Deeper, broader and artier domain generalization.
\newblock In \emph{ICCV}, 2017.

\bibitem[Li et~al.(2018{\natexlab{a}})Li, Yang, Song, and
  Hospedales]{li2018learning}
Li, D., Yang, Y., Song, Y.-Z., and Hospedales, T.
\newblock Learning to generalize: Meta-learning for domain generalization.
\newblock In \emph{AAAI}, 2018{\natexlab{a}}.

\bibitem[Li et~al.(2018{\natexlab{b}})Li, Pan, Wang, and Kot]{li2018domain}
Li, H., Pan, S.~J., Wang, S., and Kot, A.~C.
\newblock Domain generalization with adversarial feature learning.
\newblock In \emph{CVPR}, 2018{\natexlab{b}}.

\bibitem[Li et~al.(2020)Li, Gu, Zhou, Chen, and Banerjee]{li2020hessian}
Li, X., Gu, Q., Zhou, Y., Chen, T., and Banerjee, A.
\newblock Hessian based analysis of sgd for deep nets: Dynamics and
  generalization.
\newblock In \emph{SIAM}, 2020.

\bibitem[Li et~al.(2018{\natexlab{c}})Li, Gong, Tian, Liu, and
  Tao]{li2018domaincond}
Li, Y., Gong, M., Tian, X., Liu, T., and Tao, D.
\newblock Domain generalization via conditional invariant representations.
\newblock In \emph{AAAI}, 2018{\natexlab{c}}.

\bibitem[Liu et~al.(2021)Liu, Zhang, Kuang, Zhou, Xue, Wang, Chen, Yang, Liao,
  and Zhang]{pmlr-v139-liu21ab}
Liu, L., Zhang, S., Kuang, Z., Zhou, A., Xue, J.-H., Wang, X., Chen, Y., Yang,
  W., Liao, Q., and Zhang, W.
\newblock Group fisher pruning for practical network compression.
\newblock In \emph{ICML}, 2021.

\bibitem[Long et~al.(2014)Long, Wang, Ding, Sun, and Yu]{long2014transfer}
Long, M., Wang, J., Ding, G., Sun, J., and Yu, P.~S.
\newblock Transfer joint matching for unsupervised domain adaptation.
\newblock In \emph{CVPR}, 2014.

\bibitem[Lopez-Paz \& Ranzato(2017)Lopez-Paz and Ranzato]{NIPS2017_f8752278}
Lopez-Paz, D. and Ranzato, M.~A.
\newblock Gradient episodic memory for continual learning.
\newblock In \emph{NeurIPS}, 2017.

\bibitem[Maddox et~al.(2019)Maddox, Tang, Moreno, Wilson, and
  Damianou]{maddox2019transfer}
Maddox, W.~J., Tang, S., Moreno, P.~G., Wilson, A.~G., and Damianou, A.
\newblock On transfer learning via linearized neural networks.
\newblock In \emph{NeurIPS workshop}, 2019.

\bibitem[Mancini et~al.(2018)Mancini, Bulo, Caputo, and Ricci]{mancini2018best}
Mancini, M., Bulo, S.~R., Caputo, B., and Ricci, E.
\newblock Best sources forward: domain generalization through source-specific
  nets.
\newblock In \emph{ICIP}, 2018.

\bibitem[Mansilla et~al.(2021)Mansilla, Echeveste, Milone, and
  Ferrante]{mansilla2021omain}
Mansilla, L., Echeveste, R., Milone, D.~H., and Ferrante, E.
\newblock Domain generalization via gradient surgery.
\newblock In \emph{ICCV}, 2021.

\bibitem[Martens(2014)]{martens2014new}
Martens, J.
\newblock New insights and perspectives on the natural gradient method.
\newblock \emph{arXiv preprint}, 2014.

\bibitem[Martens \& Grosse(2015)Martens and Grosse]{10.5555/3045118.3045374}
Martens, J. and Grosse, R.
\newblock Optimizing neural networks with kronecker-factored approximate
  curvature.
\newblock In \emph{ICML}, 2015.

\bibitem[Muandet et~al.(2013)Muandet, Balduzzi, and
  Sch{\"o}lkopf]{muandet2013domain}
Muandet, K., Balduzzi, D., and Sch{\"o}lkopf, B.
\newblock Domain generalization via invariant feature representation.
\newblock In \emph{ICML}, 2013.

\bibitem[Nam et~al.(2021)Nam, Lee, Park, Yoon, and Yoo]{nam2021reducing}
Nam, H., Lee, H., Park, J., Yoon, W., and Yoo, D.
\newblock Reducing domain gap by reducing style bias.
\newblock In \emph{CVPR}, 2021.

\bibitem[Nam et~al.(2020)Nam, Cha, Ahn, Lee, and Shin]{NEURIPS2020_eddc3427}
Nam, J., Cha, H., Ahn, S., Lee, J., and Shin, J.
\newblock Learning from failure: De-biasing classifier from biased classifier.
\newblock In \emph{NeurIPS}, 2020.

\bibitem[Parascandolo et~al.(2021)Parascandolo, Neitz, Orvieto, Gresele, and
  Sch{\"o}lkopf]{parascandolo2021learning}
Parascandolo, G., Neitz, A., Orvieto, A., Gresele, L., and Sch{\"o}lkopf, B.
\newblock Learning explanations that are hard to vary.
\newblock In \emph{ICLR}, 2021.

\bibitem[Paszke et~al.(2019)Paszke, Gross, Massa, Lerer, Bradbury, Chanan,
  Killeen, Lin, Gimelshein, Antiga, Desmaison, Kopf, Yang, DeVito, Raison,
  Tejani, Chilamkurthy, Steiner, Fang, Bai, and Chintala]{NEURIPS2019_9015}
Paszke, A., Gross, S., Massa, F., Lerer, A., Bradbury, J., Chanan, G., Killeen,
  T., Lin, Z., Gimelshein, N., Antiga, L., Desmaison, A., Kopf, A., Yang, E.,
  DeVito, Z., Raison, M., Tejani, A., Chilamkurthy, S., Steiner, B., Fang, L.,
  Bai, J., and Chintala, S.
\newblock Pytorch: An imperative style, high-performance deep learning library.
\newblock In \emph{NeurIPS}, 2019.

\bibitem[Pearl(2009)]{pearl2009causality}
Pearl, J.
\newblock \emph{Causality}.
\newblock Cambridge university press, 2009.

\bibitem[Peng et~al.(2019)Peng, Bai, Xia, Huang, Saenko, and
  Wang]{peng2019moment}
Peng, X., Bai, Q., Xia, X., Huang, Z., Saenko, K., and Wang, B.
\newblock Moment matching for multi-source domain adaptation.
\newblock In \emph{ICCV}, 2019.

\bibitem[Peters et~al.(2016)Peters, B{\"u}hlmann, and
  Meinshausen]{peters2016causal}
Peters, J., B{\"u}hlmann, P., and Meinshausen, N.
\newblock Causal inference by using invariant prediction: identification and
  confidence intervals.
\newblock \emph{Journal of the Royal Statistical Society}, 2016.

\bibitem[Pezeshki et~al.(2021)Pezeshki, Kaba, Bengio, Courville, Precup, and
  Lajoie]{pezeshki2020radient}
Pezeshki, M., Kaba, S.-O., Bengio, Y., Courville, A., Precup, D., and Lajoie,
  G.
\newblock Gradient starvation: A learning proclivity in neural networks.
\newblock In \emph{NeurIPS}, 2021.

\bibitem[Rame et~al.(2022)Rame, Kirchmeyer, Rahier, Rakotomamonjy, Gallinari,
  and Cord]{rame2022diwa}
Rame, A., Kirchmeyer, M., Rahier, T., Rakotomamonjy, A., Gallinari, P., and
  Cord, M.
\newblock Diverse weight averaging for out-of-distribution generalization.
\newblock \emph{arXiv preprint}, 2022.

\bibitem[Roberts et~al.(2021)Roberts, Driggs, Thorpe, Gilbey, Yeung, Ursprung,
  Aviles-Rivero, Etmann, McCague, Beer, et~al.]{roberts2021common}
Roberts, M., Driggs, D., Thorpe, M., Gilbey, J., Yeung, M., Ursprung, S.,
  Aviles-Rivero, A.~I., Etmann, C., McCague, C., Beer, L., et~al.
\newblock Common pitfalls and recommendations for using machine learning to
  detect and prognosticate for covid-19 using chest radiographs and ct scans.
\newblock \emph{Nature Machine Intelligence}, 2021.

\bibitem[Rojas-Carulla et~al.(2018)Rojas-Carulla, Sch{\"o}lkopf, Turner, and
  Peters]{rojas2018invariant}
Rojas-Carulla, M., Sch{\"o}lkopf, B., Turner, R., and Peters, J.
\newblock Invariant models for causal transfer learning.
\newblock \emph{JMLR}, 2018.

\bibitem[Rosenfeld et~al.(2021)Rosenfeld, Ravikumar, and
  Risteski]{rosenfeld2021the}
Rosenfeld, E., Ravikumar, P.~K., and Risteski, A.
\newblock The risks of invariant risk minimization.
\newblock In \emph{ICLR}, 2021.

\bibitem[Sagawa et~al.(2020{\natexlab{a}})Sagawa, Koh, Hashimoto, and
  Liang]{Sagawa2020Distributionally}
Sagawa, S., Koh, P.~W., Hashimoto, T.~B., and Liang, P.
\newblock Distributionally robust neural networks.
\newblock In \emph{ICLR}, 2020{\natexlab{a}}.

\bibitem[Sagawa et~al.(2020{\natexlab{b}})Sagawa, Raghunathan, Koh, and
  Liang]{pmlr-v119-sagawa20a}
Sagawa, S., Raghunathan, A., Koh, P.~W., and Liang, P.
\newblock An investigation of why overparameterization exacerbates spurious
  correlations.
\newblock In \emph{ICML}, 2020{\natexlab{b}}.

\bibitem[Sagun et~al.(2018)Sagun, Evci, Guney, Dauphin, and
  Bottou]{sagun2018empirical}
Sagun, L., Evci, U., Guney, V.~U., Dauphin, Y., and Bottou, L.
\newblock Empirical analysis of the hessian of over-parametrized neural
  networks, 2018.

\bibitem[Sankararaman et~al.(2020)Sankararaman, De, Xu, Huang, and
  Goldstein]{sankararaman2020impact}
Sankararaman, K.~A., De, S., Xu, Z., Huang, W.~R., and Goldstein, T.
\newblock The impact of neural network overparameterization on gradient
  confusion and stochastic gradient descent.
\newblock In \emph{ICML}, 2020.

\bibitem[Schaul et~al.(2013)Schaul, Zhang, and LeCun]{pmlr-v28-schaul13}
Schaul, T., Zhang, S., and LeCun, Y.
\newblock No more pesky learning rates.
\newblock In \emph{ICML}, 2013.

\bibitem[Schraudolph(2002)]{schraudolph2002fast}
Schraudolph, N.~N.
\newblock Fast curvature matrix-vector products for second-order gradient
  descent.
\newblock In \emph{Neural computation}, 2002.

\bibitem[Shah et~al.(2020)Shah, Tamuly, Raghunathan, Jain, and
  Netrapalli]{NEURIPS2020_6cfe0e61}
Shah, H., Tamuly, K., Raghunathan, A., Jain, P., and Netrapalli, P.
\newblock The pitfalls of simplicity bias in neural networks.
\newblock In \emph{NeurIPS}, 2020.

\bibitem[Shahtalebi et~al.(2021)Shahtalebi, Gagnon-Audet, Laleh, Faramarzi,
  Ahuja, and Rish]{shahtalebi2021andmask}
Shahtalebi, S., Gagnon-Audet, J.-C., Laleh, T., Faramarzi, M., Ahuja, K., and
  Rish, I.
\newblock Sand-mask: An enhanced gradient masking strategy for the discovery of
  invariances in domain generalization.
\newblock In \emph{ICML UDL Workshop}, 2021.

\bibitem[Shi et~al.(2021)Shi, Seely, Torr, Siddharth, Hannun, Usunier, and
  Synnaeve]{shi2021gradient}
Shi, Y., Seely, J., Torr, P.~H., Siddharth, N., Hannun, A., Usunier, N., and
  Synnaeve, G.
\newblock Gradient matching for domain generalization.
\newblock \emph{arXiv preprint}, 2021.

\bibitem[Singh \& Alistarh(2020)Singh and Alistarh]{NEURIPS2020_d1ff1ec8}
Singh, S.~P. and Alistarh, D.
\newblock Woodfisher: Efficient second-order approximation for neural network
  compression.
\newblock In \emph{NeurIPS}, 2020.

\bibitem[Sun \& Saenko(2016)Sun and Saenko]{sun2016deep}
Sun, B. and Saenko, K.
\newblock Deep coral: Correlation alignment for deep domain adaptation.
\newblock In \emph{ECCV}, 2016.

\bibitem[Sun et~al.(2016)Sun, Feng, and Saenko]{coral216aaai}
Sun, B., Feng, J., and Saenko, K.
\newblock Return of frustratingly easy domain adaptation.
\newblock In \emph{AAAI}, 2016.

\bibitem[Tenenbaum(2018)]{tenenbaum2018building}
Tenenbaum, J.
\newblock Building machines that learn and think like people.
\newblock In \emph{AAMAS}, 2018.

\bibitem[Teney et~al.(2020)Teney, Abbasnejad, and van~den
  Hengel]{teney2020nshuffling}
Teney, D., Abbasnejad, E., and van~den Hengel, A.
\newblock Unshuffling data for improved generalization.
\newblock \emph{arXiv preprint}, 2020.

\bibitem[Teney et~al.(2021)Teney, Abbasnejad, Lucey, and van~den
  Hengel]{teneybias}
Teney, D., Abbasnejad, E., Lucey, S., and van~den Hengel, A.
\newblock Evading the simplicity bias: Training a diverse set of models
  discovers solutions with superior {OOD} generalization.
\newblock \emph{arXiv preprint}, 2021.

\bibitem[Theis et~al.(2018)Theis, Korshunova, Tejani, and Huszár]{Theis2018a}
Theis, L., Korshunova, I., Tejani, A., and Huszár, F.
\newblock Faster gaze prediction with dense networks and fisher pruning.
\newblock \emph{arXiv preprint}, 2018.

\bibitem[Thomas et~al.(2020)Thomas, Pedregosa, van Merriënboer, Manzagol,
  Bengio, and Roux]{interplayinfomatrix2020}
Thomas, V., Pedregosa, F., van Merriënboer, B., Manzagol, P.-A., Bengio, Y.,
  and Roux, N.~L.
\newblock On the interplay between noise and curvature and its effect on
  optimization and generalization.
\newblock In \emph{AISTATS}, 2020.

\bibitem[Turner et~al.(2021)Turner, Eriksson, McCourt, Kiili, Laaksonen, Xu,
  and Guyon]{pmlr-v133-turner21a}
Turner, R., Eriksson, D., McCourt, M., Kiili, J., Laaksonen, E., Xu, Z., and
  Guyon, I.
\newblock Bayesian optimization is superior to random search for machine
  learning hyperparameter tuning: Analysis of the black-box optimization
  challenge 2020.
\newblock In \emph{NeurIPS}, 2021.

\bibitem[Tzeng et~al.(2014)Tzeng, Hoffman, Zhang, Saenko, and
  Darrell]{tzeng2014deep}
Tzeng, E., Hoffman, J., Zhang, N., Saenko, K., and Darrell, T.
\newblock Deep domain confusion: Maximizing for domain invariance.
\newblock In \emph{CoRR}, 2014.

\bibitem[Valle-Perez et~al.(2019)Valle-Perez, Camargo, and
  Louis]{valle-perez2018deep}
Valle-Perez, G., Camargo, C.~Q., and Louis, A.~A.
\newblock Deep learning generalizes because the parameter-function map is
  biased towards simple functions.
\newblock In \emph{ICLR}, 2019.

\bibitem[Vapnik(1999)]{vapnik1999overview}
Vapnik, V.~N.
\newblock An overview of statistical learning theory.
\newblock In \emph{TNN}, 1999.

\bibitem[Venkateswara et~al.(2017)Venkateswara, Eusebio, Chakraborty, and
  Panchanathan]{venkateswara2017deep}
Venkateswara, H., Eusebio, J., Chakraborty, S., and Panchanathan, S.
\newblock Deep hashing network for unsupervised domain adaptation.
\newblock In \emph{CVPR}, 2017.

\bibitem[Wald et~al.(2021)Wald, Feder, Greenfeld, and Shalit]{wald2021n}
Wald, Y., Feder, A., Greenfeld, D., and Shalit, U.
\newblock On calibration and out-of-domain generalization.
\newblock In \emph{NeurIPS}, 2021.

\bibitem[Wang et~al.(2020)Wang, Li, and Kot]{wang2020heterogeneous}
Wang, Y., Li, H., and Kot, A.~C.
\newblock Heterogeneous domain generalization via domain mixup.
\newblock In \emph{ICASSP}, 2020.

\bibitem[Wu et~al.(2020)Wu, Inkpen, and El-Roby]{wu2020dual}
Wu, Y., Inkpen, D., and El-Roby, A.
\newblock Dual mixup regularized learning for adversarial domain adaptation.
\newblock In \emph{ECCV}, 2020.

\bibitem[Xie et~al.(2021)Xie, Kumar, Jones, Khani, Ma, and
  Liang]{xie2021innout}
Xie, S.~M., Kumar, A., Jones, R., Khani, F., Ma, T., and Liang, P.
\newblock In-n-out: Pre-training and self-training using auxiliary information
  for out-of-distribution robustness.
\newblock In \emph{ICLR}, 2021.

\bibitem[Yan et~al.(2020)Yan, Song, Li, Zou, and Ren]{yan2020improve}
Yan, S., Song, H., Li, N., Zou, L., and Ren, L.
\newblock Improve unsupervised domain adaptation with mixup training.
\newblock \emph{arXiv preprint}, 2020.

\bibitem[Yang \& Salman(2019)Yang and Salman]{yang2019}
Yang, G. and Salman, H.
\newblock A fine-grained spectral perspective on neural networks.
\newblock \emph{arXiv preprint}, 2019.

\bibitem[Ye et~al.(2021)Ye, Li, Hong, Bai, Chen, Zhou, and Li]{ye2021odbench}
Ye, N., Li, K., Hong, L., Bai, H., Chen, Y., Zhou, F., and Li, Z.
\newblock Ood-bench: Benchmarking and understanding out-of-distribution
  generalization datasets and algorithms.
\newblock \emph{arXiv preprint}, 2021.

\bibitem[Yin et~al.(2018)Yin, Pananjady, Lam, Papailiopoulos, Ramchandran, and
  Bartlett]{yin2018gradient}
Yin, D., Pananjady, A., Lam, M., Papailiopoulos, D., Ramchandran, K., and
  Bartlett, P.
\newblock Gradient diversity: a key ingredient for scalable distributed
  learning.
\newblock In \emph{AISTATS}, 2018.

\bibitem[Yu et~al.(2020)Yu, Kumar, Gupta, Levine, Hausman, and
  Finn]{yu2020gradient}
Yu, T., Kumar, S., Gupta, A., Levine, S., Hausman, K., and Finn, C.
\newblock Gradient surgery for multi-task learning.
\newblock In \emph{NeurIPS}, 2020.

\bibitem[Zhang et~al.(2020)Zhang, Marklund, Dhawan, Gupta, Levine, and
  Finn]{zhang2020daptive}
Zhang, M., Marklund, H., Dhawan, N., Gupta, A., Levine, S., and Finn, C.
\newblock Adaptive risk minimization: A meta-learning approach for tackling
  group distribution shift.
\newblock \emph{arXiv preprint}, 2020.

\bibitem[Zhang et~al.(2021)Zhang, Cui, Xu, Zhou, He, and Shen]{zhang2021deep}
Zhang, X., Cui, P., Xu, R., Zhou, L., He, Y., and Shen, Z.
\newblock Deep stable learning for out-of-distribution generalization.
\newblock In \emph{CVPR}, 2021.

\bibitem[Zhang et~al.(2019)Zhang, Yu, and Turk]{zhang2019learning}
Zhang, Y., Yu, W., and Turk, G.
\newblock Learning novel policies for tasks.
\newblock In \emph{ICML}, 2019.

\bibitem[Zhao et~al.(2019)Zhao, Des~Combes, Zhang, and
  Gordon]{zhao2019learning}
Zhao, H., Des~Combes, R.~T., Zhang, K., and Gordon, G.
\newblock On learning invariant representations for domain adaptation.
\newblock In \emph{ICML}, 2019.

\end{thebibliography}
